\documentclass{article}
\usepackage{arxiv,times}

\usepackage{amsmath,amsfonts,bm,dsfont,amsthm,nicefrac}

\newcommand{\softmax}{\mathrm{softmax}}

\newcommand{\result}[2]{#1$_{\pm #2}$}

\renewcommand{\inf}[1]{\underset{#1}{\text{inf}}\,}

\newcommand{\argmin}[1]{\underset{#1}{\text{arg min}}\,}
\newcommand{\argmax}[1]{\underset{#1}{\text{arg max}}\,}
\newcommand{\arginf}[1]{\underset{#1}{\text{arg inf}}\,}
\newcommand{\argsup}[1]{\underset{#1}{\text{arg sup}}\,}

\newcommand{\sg}{\operatorname{sg}}

\usepackage{pifont}
\usepackage{booktabs}
\usepackage{hyperref}
\usepackage{url}
\usepackage[acronym]{glossaries}
\usepackage{nicefrac}
\usepackage{graphicx, subcaption, wrapfig}
\usepackage{amsmath, amsfonts, amsthm, dsfont, nicefrac}
\usepackage[table, dvipsnames]{xcolor}
\usepackage{wrapfig}
\usepackage{algorithm}
\usepackage{algpseudocode}
\usepackage{float}
\usepackage{multirow}
\usepackage{rotating}

\definecolor{lightblue}{rgb}{0.21,0.49,0.74}
\definecolor{myblue}{rgb}{0.00,0.45,0.70}
\definecolor{myorange}{rgb}{0.90,0.62,0.00}
\definecolor{myteal}{rgb}{0.00,0.62,0.45}

\glsdisablehyper



\newcommand{\cmark}{\ding{51}}%
\newcommand{\xmark}{\ding{55}}%

\usepackage{tikz}

\newacronym{ot}{OT}{Optimal Transport}
\newacronym{otdd}{OTDD}{OT Dataset Distance}
\newacronym{icnn}{ICNN}{Input Convex Neural Network}
\newacronym{jko}{JKO}{Jordan, Kinderlehrer, Otto}
\newacronym{pl}{PL}{Polyak Łojasiewicz}
\newacronym{wgf}{WGF}{Wasserstein Gradient Flow}
\newacronym{ml}{ML}{Machine Learning}
\newacronym{gmm}{GMM}{Gaussian Mixture Model}
\newacronym{eeg}{EEG}{electroencephalogram}
\newacronym{erm}{ERM}{Empirical Risk Minimization}

\newacronym{ode}{ODE}{Ordinary Differential Equation}
\newacronym{sde}{SDE}{Stochastic Differential Equation}
\newacronym{cfm}{CFM}{Conditional Flow Matching}
\newacronym{bcd}{BCD}{Block Coordinate Descent}
\newacronym{sbm}{SBM}{Schrondinger Bridge Matching}
\newacronym{mmot}{MMOT}{Multi-Marginal OT}

\newacronym{baryfm}{BaryFM}{Barycentric Flow Matching}

\newacronym{bw2-uvp}{BW2-UVP}{Bures Wasserstein Unexplained Variance}
\newacronym{mom}{MoM}{Measure over Measure}

\newacronym{lodo}{LODO}{Leave-One-Domain-Out}
\newacronym{wbt}{WBT}{Wasserstein Barycenter Transport}
\newacronym{dp}{DP}{Demographic Parity}
\newacronym{di}{DI}{Disparate Impact}

\newacronym{pca}{PCA}{Principal Component Analysis}
\newacronym{vae}{VAE}{Variational Autoencoder}

\newacronym{mcmc}{MCMC}{Markov Chain Monte Carlo}
\newacronym{cfg}{CFG}{Classifier Free Guidance}

\newacronym{mlp}{MLP}{Multilayer Perceptron}

\usepackage{hyperref}
\usepackage{url}

\title{Towards Universal Wasserstein Barycenters through Flow Matching}

\author{Eduardo Fernandes Montesuma \\
Sigma Nova \\
Paris, France \\
\texttt{eduardo.montesuma@sigmanova.ai}
}

\begin{document}

\maketitle

\begin{abstract}
Defining a weighted mean over probability measures under probability metrics is a central tool in probabilistic machine learning. Under the Wasserstein metric, these are called \emph{Wasserstein barycenters}. While most approaches compute barycenters for a fixed weight vector, approximating the whole family of barycenters over the simplex, which we call the \emph{Wasserstein simplex}, remains underexplored. We refer to this problem as \emph{Universal Barycenter Approximation}, and propose \texttt{BaryFM}, a flow matching model transporting the marginal measures into any barycenter in the Wasserstein simplex. Once trained, the network can draw samples from measures in the Wasserstein simplex through an ordinary differential equation. We validate our method on 4 downstream tasks: domain adaptation, generalization, Bayesian posterior aggregation and algorithmic fairness. \texttt{BaryFM} achieves the best average rank among 15 competing methods across 10 domain adaptation benchmarks, matching or surpassing non-universal solvers.
\end{abstract}

\section{Introduction}

\begin{wrapfigure}[19]{r}{0.35\textwidth}
  \centering
  \includegraphics[width=\linewidth]{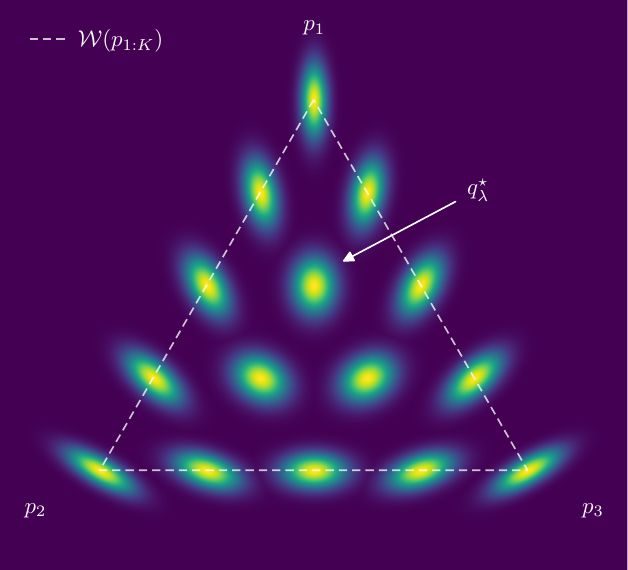}
  \caption{Wasserstein simplex over the set of Gaussian measures. In this paper, we study how to approximate every barycenter in the family $\mathcal{W}(p_{1:K})$.}
  \label{fig:wasserstein-simplex}
\end{wrapfigure}
In probabilistic machine learning, one is often faced with the question of how to compute an average or centroid over a family of probability measures~\citep{srivastava2018scalable,montesuma2021wasserstein} under a suitable metric. Previous studies~\citep{cuturi2014fast,korotin2022wasserstein,howard2025schrdinger,montesuma2025computing} consider this question through the lens of the Wasserstein distance, in which case the centroid is known as Wasserstein barycenter~\citep{agueh2011barycenters}, but despite specific cases~\citep[Appendix B]{fan2020scalable}, the question of how to approximate the family of \emph{interpolating measures} remains open. Figure~\ref{fig:wasserstein-simplex} shows an example of this family over $K=3$ Gaussian marginals $p_{1},p_{2},p_{3}$ ($p_{1:3}$, in short). Therefore, we study the \emph{Wasserstein simplex},
\begin{align}
    \mathcal{W}(p_{1:K}) = \{\text{Bar}(\lambda,p_{1:K}): \lambda \in \Delta_{K}\},\label{eq:wasserstein-simplex}
\end{align}
where $\Delta_{K} = \{ \lambda \in \mathbb{R}^{K}_{\geq 0}: \sum_k \lambda_k = 1 \}$ is the simplex, and $q_{\lambda}^{\star} = \text{Bar}(\lambda,p_{1:K})$ denotes the Wasserstein barycenter (c.f. equation~\ref{eq:barycenter}) of marginals $p_{1:K}$ with weights $\lambda \in \Delta_{K}$.

In this context, barycenter methods can be classified in 2 axes of \emph{amortization}. Most methods are designed to amortize sampling, meaning, drawing new examples from a barycentric measure $q_{\lambda}^{\star}$, usually with uniform weights $\lambda = \bar{\lambda} = (K^{-1},\cdots,K^{-1}) \in \Delta_{K}$. The second axis of amortization concerns the measures $q_{\lambda}^{\star} \in \mathcal{W}(p_{1:K})$ themselves, namely, the possibility of drawing samples from $q_{\lambda}^{\star}$ on any query $\lambda \in \Delta_{K}$ without solving a new optimization problem. The second problem is under-explored, and, to the best of our knowledge, only~\cite[Appendix B]{fan2020scalable} has considered it in restricted scenarios (e.g., Gaussian measures). We call this problem \textbf{Universal Barycenter Approximation}, and propose a universal barycenter approximation algorithm, i.e., a method that amortizes sampling along the 2 prescribed axes. In the following, we discuss prior art on each axis.

On the first axis, existing approaches are mainly based on generative models and neural networks. For instance,~\cite{fan2020scalable} and~\cite{korotin2021continuous} propose approaches based on \glspl{icnn}~\citep{amos2017input} to approximate transport maps. On the generative modeling side,~\cite{korotin2022neural} parametrize the barycentric measure as a map from a latent space to the data space,~\cite{visentin2025computing} computes barycenters via normalizing flows~\citep{papamakarios2021normalizing}, and~\cite{noble2023tree,howard2025schrdinger} use Schrodinger bridges~\citep{de2021diffusion}. Non-amortized methods~\citep{cuturi2014fast,montesuma2025computing}, also called empirical, directly optimize the support of the barycentric measure, which limits its scalability since each new drawn requires a new problem solve.

On the second axis, the aforementioned families share a common limitation, that is, they need to solve equation~\ref{eq:barycenter} from scratch for each $\lambda \in \Delta_{K}$. A notable exception is~\citep[Appendix B]{fan2020scalable}, who adds $\lambda$ as an input to their $(2K+1)$-\glspl{icnn}, and validates their approach over Gaussian measures. Beyond these amortization axes, downstream tasks such as domain adaptation and generalization require barycenters to computed over joint measures~\citep{montesuma2021wasserstein,montesuma2025computing}, i.e., barycenters over $\Omega = \mathbb{R}^{d} \times \Delta_{C}$, for $C$-classes. To the best of our knowledge, no existing neural or universal method handles joint measures (c.f. Table~\ref{tab:sota-comparison}).

We propose \texttt{BaryFM}, a \gls{cfm}~\citep{lipman2022flow} model that transports the marginal measures $p_{1:K}$ into any measure in their Wasserstein simplex $\mathcal{W}(p_{1:K})$. Our architecture (c.f. Figure~\ref{fig:arch}) is composed of a shared encoding layer that embeds the variables associated with the flow, namely $z = (x, y)$, $t \in [0, 1]$, $k \in \{1,\cdots,K\}$ and $\lambda \in \Delta_{K}$. This architecture has 2 heads. One \gls{cfm} head that predicts the velocity field of the associated flow, and one dual \gls{ot} head, that serves to match samples from the involved measures, as well as to transport them. The distinguishing feature of our \gls{cfm} method is that it estimates its targets \emph{on-the-fly}. Concretely, our contributions are as follows,
\begin{enumerate}
    \item We propose \texttt{BaryFM} and its architecture, to the best of our knowledge, the first neural, universal barycenter solver that handles joint measures,
    \item We provide a comprehensive evaluation of our method against prior art on 4 downstream applications of Wasserstein barycenters: domain adaptation (Section~\ref{sec:exp-da}), generalization (Section~\ref{sec:exp-dg}), fairness (Section~\ref{sec:additional-fairness}) and Bayesian Posterior Aggregation (Section~\ref{sec:additional-bpa}).
\end{enumerate}

\section{Background}\label{sec:background}

\subsection{Optimal Transport and Barycenters}\label{sec:ot}

Let $\Omega \subseteq \mathbb{R}^{\text{dim}}$ be a convex sub-set of the Euclidean space. \gls{ot} is a field of mathematics studying the transport of probability measures at least effort~\citep{villani2009optimal,peyre2019computational}. It was originally founded by Gaspar Monge~\cite{monge1781memoire}, where one seeks for the \gls{ot} map from $p \in \mathcal{P}_{2}(\Omega)$ to $q \in \mathcal{P}_{2}(\Omega)$, where $\mathcal{P}_{2}(\Omega)$ is the space of probability measures with finite 2nd moments:
\begin{align}
    T_{p\to q}^{\star} = \arginf{T \in \mathcal{T}(p,q)}\mathbb{E}_{z\sim p}[c(z,T(z))]\quad\mathcal{T}(p,q)=\{T:\Omega\to\Omega:T_{\sharp}p=q\},\label{eq:monge}
\end{align}
where $T_{\sharp}p$ denotes the push-forward of $p$ by the map $T$, and $c:\Omega\times\Omega\to\mathbb{R}$ is a ground-cost. In the 20th century,~\cite{kantorovich1942translocation} introduced a relaxation of the Monge problem,
\begin{align}
    \gamma^{\star} = \arginf{\gamma \in \Gamma(p, q)}\mathbb{E}_{(z,z') \sim \gamma}[c(z,z')]\quad\Gamma(p,q) = \{\gamma \in \mathcal{P}_{2}(\Omega\times\Omega):\pi_{1,\sharp}\gamma=p,\pi_{2,\sharp}\gamma=q\},\label{eq:kantorovich}
\end{align}
where $\mathcal{P}_{2}(\Omega)$ denotes the set of probability measures with finite second moment and $\pi_{i}(z_{1},\cdots,z_{n}) = z_{i}$ denotes the $i-$th coordinate projection. Following~\cite{cuturi2013sinkhorn}, one can regularize the Kantorovich problem with the entropy of $\gamma$, leading to,
\begin{align}
    \gamma_{\epsilon}^{\star} = \text{OT}_{\epsilon}(p,q) = \arginf{\gamma \in \Gamma(p, q)}\mathbb{E}_{(z,z') \sim \gamma}[c(z,z')] - \epsilon H(\gamma)\quad H(\gamma) =-\int \log \dfrac{d\gamma}{d(p\otimes q)}d\gamma,\label{eq:sinkhorn}
\end{align}
where $p\otimes q$ denotes the product measure, and the regularization strength $\epsilon \geq 0$. Here, we use the convention $H(\gamma) = -\infty$ when $\gamma \not\ll p\otimes q$. Through equations~\ref{eq:kantorovich} and~\ref{eq:sinkhorn} one can define notions of dissimilarity on $\mathcal{P}_{2}$, known as Wasserstein distance ($\epsilon = 0$) or its entropic approximation ($\epsilon > 0$). Let $c(z, z') = \lVert z - z' \rVert^2_{2}$ be the Euclidean metric on $\Omega$,
\begin{align}
    W_{2,\epsilon}(p,q)^{2} = \inf{\gamma \in \Gamma(p,q)}\mathbb{E}_{(z,z')\sim\gamma}[d(z,z')^{2}]- \epsilon H(\gamma).\label{eq:wasserstein}
\end{align}
In general \gls{ot} lifts $(\Omega, \lVert \cdot \rVert_{2})$ into $(\mathcal{P}_{2}(\Omega),W_{2,\epsilon})$, and more precisely, when $\epsilon = 0$ $(\mathcal{P}_{2}(\Omega), W_{2,0})$ defines a metric space. The Kantorovich problem (equation~\ref{eq:kantorovich}) defines a linear program. As such, via Fenchel-Rockafellar duality~\citep{rockafellar1997convex,villani2021topics,genevay2016stochastic}, the primal forms in equations~\ref{eq:kantorovich} and~\ref{eq:sinkhorn} admit a dual,
\begin{align}
    (f^{\star}_{\epsilon}, g^{\star}_{\epsilon}) = \text{DualOT}_{\epsilon}(p,q) = \argsup{f,g\in\mathcal{C}(\Omega)}\mathbb{E}_{z\sim p}[f(z)]+\mathbb{E}_{z'\sim q}[g(z')]-I_{\epsilon}(
    f,g,c),\label{eq:dual_ot}
\end{align}
where $I_{\epsilon}(f,g,c) = \epsilon \mathbb{E}_{z\sim p,z'\sim q}[\exp(\frac{f(z)+g(z')-c(z,z')}{\epsilon})]$ for $\epsilon > 0$ and $I_{0}(f,g,c) = \iota_{\Phi(c)}(f, g)$, where $\iota_{A}(x)$ is the indicator function of set $A$ (i.e., $0$ if $x \in A$ and $+\infty$ otherwise), $\Phi(c) = \{ (f,g ): f(z) + g(z') \leq c(z, z') \}$, and $\mathcal{C}(\Omega)$ is the set of continuous functions over $\Omega$. Following~\cite{genevay2016stochastic}, the dual problem can be cast as a single function maximization using the c-transform,
\begin{align}
    f^{(c)}_{\epsilon}(z') = \text{c-transform}_{\epsilon}(f_{\epsilon}^{\star})(z') = \begin{cases}
       \text{min}_{z \in \Omega}\{c(z,z')-f_{\epsilon}^{\star}(z)\} & \epsilon = 0,\\
       -\epsilon \log \mathbb{E}_{z \sim p}[\exp(\frac{f_{\epsilon}^{\star}(z)-c(z,z')}{\epsilon})] & \epsilon > 0.
    \end{cases}\label{eq:c-transform}
\end{align}
We refer readers to~\cite[Remark 4.25]{peyre2019computational} for further computational insight. Equations~\ref{eq:dual_ot} and~\ref{eq:c-transform} will be useful in Section~\ref{sec:methodology}, as we approximate $f_{\epsilon}^{\star}$ with a neural network, similarly to~\cite{seguy2018large}. $g_{\epsilon}^{\star}$ is recovered analytically through~\ref{eq:c-transform}. Furthermore, from common entropic \gls{ot} theory, we recover the \gls{ot} plan, up to a normalization constant, through,
\begin{align}
    \dfrac{d\gamma_{\epsilon}^{\star}}{d (p\otimes q)}(z,z') \propto \exp\biggr( \frac{f_{\epsilon}^{\star}(z)+f_{\epsilon}^{(c)}(z')-c(z,z')}{\epsilon}\biggr).\label{eq:transport-plan}
\end{align}
From the entropic transport plan $\gamma_{\epsilon}^{\star}$ and the dual potentials $(f_{\epsilon}^{\star},f_{\epsilon}^{(c)})$, we can retrieve the notion of \emph{barycentric mapping} as in~\cite[Equations 12 and 13]{pooladian2021entropic},
\begin{align}
    T_{\epsilon}^{\star}(z) = \mathbb{E}_{z' \sim \gamma_{\epsilon}^{\star}(\cdot | z)}[z'] = \dfrac{1}{Z(z)}\int z'\exp\biggl(\frac{f_{\epsilon}^{(c)}(z')-c(z,z')}{\epsilon}\biggr)dq(z'),\label{eq:continuous-barymap}
\end{align}
$Z = \int \exp(\nicefrac{(f_{\epsilon}^{(c)}(z')-c(z,z'))}{\epsilon})dq(z')$ and $\gamma(\cdot \lvert z)$ denote the conditional law of $z'$ given $z$ under $\gamma$.

The central question of this paper is the computation of Wasserstein barycenters. Given measures $p_{1:K} \subset \mathcal{P}_{2}(\Omega)$, \emph{barycentric coordinates} $\lambda \in \Delta_{K}$, the barycenter problem is,
\begin{align}
    q_{\lambda}^{\star} = \text{Bar}(\lambda,p_{1:K}) = \arginf{q \in \mathcal{P}_{2}(\Omega)} \sum_{k=1}^{K}\lambda_{k}W_{2}(p_k,q)^{2},\label{eq:barycenter}
\end{align}
Introduced by~\cite{agueh2011barycenters}, this problem admits minimizers for any $p_{1:K} \subset \mathcal{P}_{2}(\Omega)$, and a unique solution when some $p_{k}$ with $\lambda_{k} > 0$ vanishes on small sets. To ensure uniqueness for every $\lambda \in \Delta_{K}$, we further assume that all $p_{k}$ vanish on small sets, and refer to $q_{\lambda}^{\star}$ as \emph{the barycenter}.~\cite{alvarez2016fixed} introduced a fixed-point algorithm for Wasserstein barycenters, through the operator,
\begin{align}
    G:\mathcal{P}_{2}(\Omega)\to\mathcal{P}_{2}(\Omega)\text{, such that }G(q) = \biggr( \sum_{k=1}^{K}\lambda_{k}T_{k}^{\star} \biggr)_{\sharp}q,\label{eq:fixed-point-ae}
\end{align}
where $T_{k}^{\star}$ is the Monge map pushing $q$ to $p_{k}$, well-defined for $q$ vanishing on small sets.

\subsection{Conditional Flow Matching}\label{sec:cfm}

\gls{cfm}~\citep{lipman2022flow} is a generative modeling technique that learns a flow from a source measure $p$ to a target measure $q$. In brief, it models a flow of particles via $\phi(z, t)$, driven by a vector field $v(z, t)$ through $\partial_t \phi(z, t) = v(\phi(z, t), t)$, where $\phi(z, 0) = z$. This flow defines a curve in $\mathcal{P}_{2}(\Omega)$, via $\rho_t = \phi(\cdot, t)_{\sharp}p$, with $\rho_0 = p$ and $\rho_1 = q$. In this setting, $\rho_t$ suffices the continuity equation, $\partial_t\rho_t = -\text{div}(\rho_t v_t)$ with velocity field $v_t = v(\cdot, t)$, and $\text{div}$ denotes the divergence operator.

At this point, the main question is how to determine $v_t$. Formally, $v_t(z) = \mathbb{E}[v_t(z_1|z_0)|z_t = z]$, where $v_t(z_1 | z_0) = z_1 - z_0$ denotes the conditional velocity of a pair $(z_0, z_1)$, and $z_t = (1-t)z_0 + tz_1$ is the linear interpolant. This conditional expectation is intractable because it requires knowing the posterior over $(z_0, z_1)$ given $z_t$. Based on this limitation,~\cite{lipman2022flow} introduced \gls{cfm}, which learns a neural net $v_{\theta}(z_t, t)$ with parameters $\theta$ by the following objective,
\begin{align}
    \mathcal{L}_{\text{CFM}}(\theta) = \mathbb{E}_{t,z_{0},z_{1}}[\lVert v_{\theta}(z_{t},t) - (z_1 - z_0) \rVert_{2}^{2}]\text{, where }t\sim\mathcal{U}[0,1],z_{0}\sim p_{0}, z_{1}\sim p_{1}.\label{eq:cfm}
\end{align}
Here,~\cite{lipman2022flow} imposes $p_0 = \mathcal{N}(0,I)$. \gls{cfm} flows pure noise to a structured, complex measure. In a later development,~\cite{tong2023improving} relaxed the assumption that $z_{0} \sim p_{0} = \mathcal{N}(0, I)$ by using a coupling $\gamma \in \Gamma(p, q)$,
\begin{align}
    \mathcal{L}_{\text{CFM}}(\theta;\gamma) = \mathbb{E}_{t,z_0,z_1}[\lVert v_{\theta}(z_{t},t) - (z_1-z_0) \rVert_{2}^{2}]\text{, where }t\sim\mathcal{U}[0, 1]\text{, and }(z_0,z_1)\sim \gamma.\label{eq:ot-cfm}
\end{align}
For $\gamma = p \otimes q$, equation~\ref{eq:ot-cfm} is equivalent to~\ref{eq:cfm}. More interestingly, using an \gls{ot} coupling, $(z_0, z_1) \sim \gamma^{\star} = \text{OT}(p_{0}, p_{1})$ leads to the so-called \gls{ot}-\gls{cfm}, $ \mathcal{L}_{\text{OT-CFM}}(\theta)= \mathcal{L}_{\text{CFM}}(\theta;\gamma^{\star})$. This construction allows building a flow between an arbitrary $p_0$ and $p_1$.

\section{\texttt{BaryFM:} Amortized, Universal Wasserstein Barycenters}\label{sec:methodology}

\begin{figure}[ht]
    \centering
    \includegraphics[width=\linewidth]{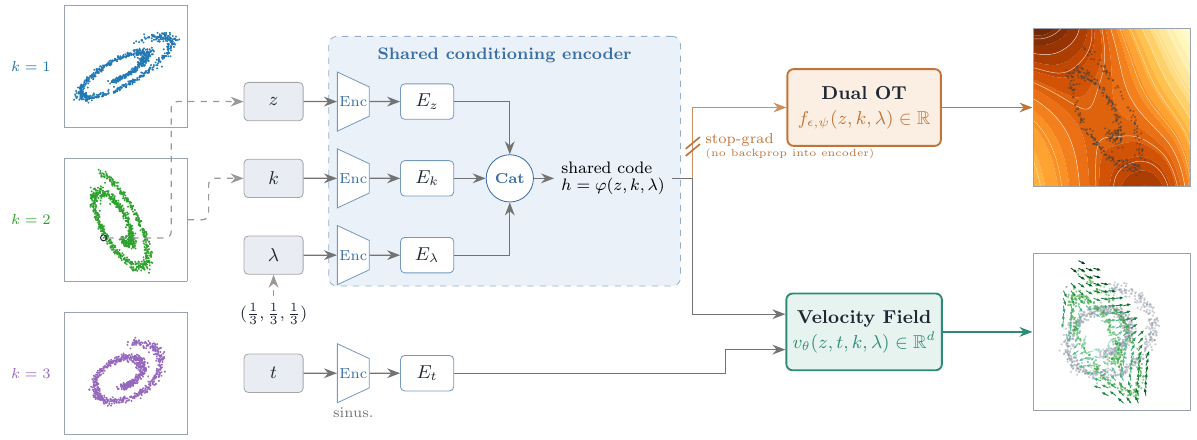}
    \caption{BaryFM architecture. A shared encoder maps $(z, k, \lambda)$ into a latent code $h$, which is used by the velocity field estimation branch, and the dual \gls{ot} branch.}
    \label{fig:arch}
\end{figure}

In this section, we present our \texttt{BaryFM} strategy. In a nutshell, our goal is to build an end-to-end amortized barycenter sampler. Our architecture is based on a 2 head network. The first head estimates the \gls{ot} problem between multiple marginals $p_{1},\cdots,p_{K}$ and the barycenter $q_{\lambda}^{(L_{\text{FP}})}$. The second head estimates a velocity field, moving samples from the marginals $p_{1},\cdots,p_{K}$ to the estimated barycenter $\hat{q}_{\lambda}^{(L_{\text{FP}})}$. We provide an illustration of this architecture in Figure~\ref{fig:arch}.

\textbf{Architecture.} Our method is a single network, composed of a shared encoder and 2 task-specific heads. Our flow has $K$ marginals, $p_{1:K}$, and a moving target $\hat{q}_{\lambda}^{(L_{\text{FP}})}$ (c.f. equation~\ref{eq:fp-neural}) since $\lambda$ is free. Therefore, we condition the network on $(k, \lambda)$. We thus build a shared code $h = \text{Cat}(E_{z}, E_{k}, E_{\lambda})$, where $\text{Cat}$ denotes the concatenation operation, and $E_{z}$ (resp. $k$, $\lambda$) denote the embedding of $z \in \Omega$, $E_{z} = \text{Enc}(z)$. Our first head, $f_{\epsilon,\psi}$, approximates the dual \gls{ot} variable $f_{\epsilon}^{\star}$ of the pair $(p_k, \hat{q}_{\lambda}^{(L_{\text{FP}})})$. We estimate the 2nd dual via the $c-$transform (c.f. equation~\ref{eq:c-transform}). The second head, $v_{\theta}$, estimates the velocity field flowing $p_k$ into $\hat{q}_{\lambda}^{(L_{\text{FP}})}$. Its input consists of $h_{t} = \text{Cat}(h, E_{t})$, where $E_{t} = \text{Enc}(t)$ is a sinusoidal positional encoding~\citep{vaswani2017attention}. Therefore, our design is different from~\cite[Appendix B]{fan2020scalable}, who adds $\lambda$ as an input to the $(2\times K+1)-$\glspl{icnn}.

\textbf{The Dual OT Estimation.} The dual \gls{ot} head estimates the Kantorovich potential of the \gls{ot} problem involving $(p_k, \hat{q}_{\lambda}^{(L_{\text{FP}})})$. This is a \emph{multi-source, moving target problem}, since $\lambda$ is not fixed, which forces $f_{\epsilon,\psi}$ to depend on $\lambda$. We learn the parameters $\psi^{\star}$ of $f_{\epsilon,\psi}$ via stochastic gradient ascent~\citep{genevay2016stochastic,seguy2018large} of the dual loss,
\begin{align}
    \mathcal{L}_{D}(\psi, \lambda) = \sum_{k=1}^{K}\lambda_{k}\biggr(\mathbb{E}_{z}[f_{\epsilon,\psi}(z,k,\lambda) ] + \mathbb{E}_{\bar z}[f_{\epsilon,\psi}^{(c)}(\bar z,k,\lambda)] - \epsilon(\mathbb{E}_{z,\bar{z}}[\gamma_{\epsilon,\psi}(z,\bar{z},k,\lambda)]-1)\biggr),\label{eq:loss-dual}
\end{align}
where we sample $\bar{z} \sim \hat{q}_{\lambda}^{(L_{\text{FP}})}$, $z \sim p_k$ and $(z,\bar{z}) \sim p_k \otimes \hat{q}_{\lambda}^{(L_{\text{FP}})}$. The transport plan $\gamma_{\epsilon,\psi}$ is defined via equation~\ref{eq:transport-plan}, and $f_{\epsilon,\psi}^{(c)}$ is the $c$-transform of $f_{\epsilon,\psi}$ (c.f. equation~\ref{eq:c-transform}).

\noindent\textbf{Estimating the CFM Targets.} The dual (c.f. equation~\ref{eq:loss-dual}) and \gls{cfm} (c.f. equation~\ref{eq:loss-cfm}) losses depend on the samples from the approximate barycenter measure, $\bar{z}$. Ideally, these samples would be acquired by iterating~\cite{alvarez2016fixed} operator $q_{\lambda}^{(\ell+1)} = G(q_{\lambda}^{(\ell)})$. That is not possible, since they depend on the Monge maps $T_{k}^{\star}$, pushing $q_{\lambda}^{(\ell)}$ to $p_k$. These mappings are not available in closed-form, which makes the computation of $G$ intractable. An approximation for $T_{k}^{\star}$ is available, via the barycentric mapping (c.f. equation~\ref{eq:continuous-barymap}),
\begin{align}
    \hat{T}_{\epsilon,\psi}(\bar{z}_{i}^{(\ell)}, k, \lambda) &= \underset{z \sim \gamma_{\epsilon,\psi}(\cdot|\bar{z}_{i}^{(\ell)},k,\lambda)}{\mathbb{E}}[z],\quad\dfrac{d\gamma_{\epsilon,\psi}}{dp_k}(z|\bar{z},k,\lambda) \propto \exp\biggl( \frac{f_{\epsilon,\psi}(z,k,\lambda) - c(z,\bar{z})}{\epsilon} \biggr),\label{eq:soft-bary-map}
\end{align}
In practice, given a batch $\{ z_{k,j} \}_{j=1}^{B}$, $z_{k,j} \sim p_k$, the conditional expectation in equation~\ref{eq:soft-bary-map} is estimated via a softmax $\hat{T}_{\epsilon,\psi}(\bar{z}_{i}^{(\ell)}, k, \lambda) = \sum_{j}\text{softmax}_{j}((f_{\epsilon,\psi}(z_{k,j}, k, \lambda) - c(z_{k,j}, \bar{z}_{i}^{(\ell)}))/\epsilon)z_{k,j}$, where $\text{softmax}_{j}$ denotes the softmax operation over the batch index $j$.

\begin{wrapfigure}[16]{r}{0.35\textwidth}
  \centering
  \includegraphics[width=\linewidth]{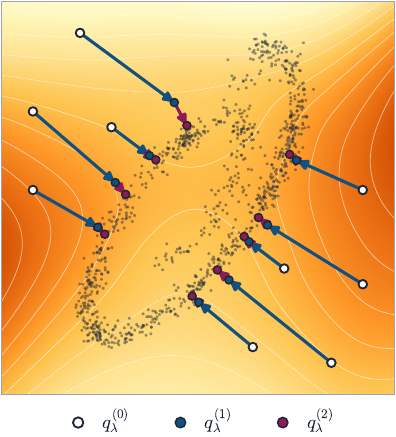}
  \caption{Illustration of the convergence of the fixed-point iterations using the neural barycentric map in equation~\ref{eq:soft-bary-map}.}
  \label{fig:fp-illustration}
\end{wrapfigure}
The barycentric map is an approximation of the Monge map, when the latter exists~\citep{deb2021rates,pooladian2021entropic}, for $\epsilon \to 0$. Effectively, we have,
\begin{align}
    \bar{z}_{i}^{(\ell+1)} &= \sg\biggl( \sum_{k=1}^{K}\lambda_{k}\hat{T}_{\epsilon,\psi}(\bar{z}_{i}^{(\ell)},k,\lambda) \biggr),\label{eq:fp-neural}\\
    \hat{q}_{\lambda}^{(\ell+1)} &= \hat{G}_{\epsilon,\psi}(\hat{q}_{\lambda}^{(\ell)}),\quad \hat{G}_{\epsilon,\psi}(q) = \biggl( \sum_{k=1}^{K}\lambda_k\hat{T}_{\epsilon,\psi}(\cdot,k,\lambda) \biggr)_{\sharp}q\nonumber
\end{align}
where $\bar{z}_{i}^{(0)} = \sum_{k=1}^{K}\lambda_{k}z_{k,i}$, $\hat{q}_{\lambda}^{(\ell)}$ denotes an empirical measure with $B$ samples in its support $\{\bar{z}_{i}^{(\ell)}\}_{i=1}^{B}$, and $\sg$ denotes the stop-grad operation. The dual \gls{ot} head amortizes \gls{ot}, that is, we avoid solving $K$ \gls{ot} problems for estimating $q_{\lambda}^{(L_{\text{FP}})}$. An illustration is shown in Figure~\ref{fig:fp-illustration}.

\begin{figure}[ht]
    \centering
    \begin{subfigure}{0.3\linewidth}
        \centering
        \includegraphics[width=0.8\linewidth]{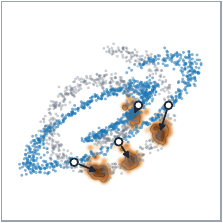}
    \end{subfigure}
    \begin{subfigure}{0.3\linewidth}
        \centering
        \includegraphics[width=0.8\linewidth]{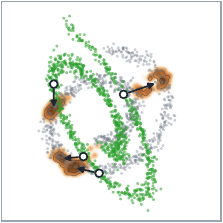}
    \end{subfigure}
    \begin{subfigure}{0.3\linewidth}
        \centering
        \includegraphics[width=0.8\linewidth]{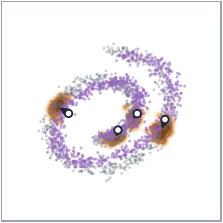}
    \end{subfigure}
    \caption{Sampling of pairs $(z_0, z_1)$ for the \gls{cfm} training. In each panel, $z_0 \sim p_k$, $k=1,\cdots, 3$ respectively. The amber shaded region corresponds to $\gamma_{\epsilon,\psi}(\cdot | z_0, k)$, from which $z_1$ is sampled.}
    \label{fig:cfm-sampling}
\end{figure}

\noindent\textbf{Conditional Flow Matching.} We follow~\cite{tong2023improving}, by drawing pairs $(z_0, z_1) \sim \gamma_{\epsilon,\psi}(\cdot,\cdot,k,\lambda) \in \Gamma(p_{k}, \hat{q}_{\lambda}^{(L_{\text{FP}})})$. More specifically, we sample $z_0 \sim p_k$, $z_{1} \sim \gamma_{\epsilon,\psi}(\cdot \lvert z_0, k, \lambda)$,
\begin{align*}
    \dfrac{d\gamma_{\epsilon,\psi}}{d\hat{q}_{\lambda}^{(L_{\text{FP}})}}(\bar{z}\lvert z_0, k, \lambda) \propto \exp\biggl(\dfrac{f_{\epsilon,\psi}^{(c)}(\bar{z}, k, \lambda) - c(z_{0}, \bar{z})}{\epsilon}\biggr),
\end{align*}
which, again, is estimated via a softmax. We provide an illustration of the sampling strategy just described in Figure~\ref{fig:cfm-sampling}. Given this sampling strategy, we minimize,
\begin{align}
    \mathcal{L}_{C}(\theta,\lambda,\psi^{\star}) &= \underset{t\sim \mathcal{U}[0,1],k\sim\text{Cat}(\lambda),z_0 \sim p_k, z_1\sim \gamma_{\epsilon,\psi^{\star}}(\cdot|z_0,k,\lambda)}{\mathbb{E}}\biggl[ \lVert v_{\theta}(z_t, t, k, \lambda) - (z_1 - z_0) \rVert_{2}^{2} \biggr],\label{eq:loss-cfm}
\end{align}
where $z_t = (1-t)z_0 + tz_1$. We illustrate the sampling strategy in Figure~\ref{fig:cfm-sampling}.

\noindent\textbf{Barycentric Flow Matching.} In summary, we put forth a shared encoder for the variables $(z, k, \lambda)$. $(k, \lambda)$ are necessary to make the flow and \gls{ot} problems tractable under a \emph{multi-source, moving target} setting. On top of the shared encoder, we use two task-specific heads: a \emph{velocity head} $v_{\theta}(z, t, k, \lambda)$ that takes the embeddings of $(z, k, \lambda)$ and $t$ and produces the velocity moving the samples from $p_k$ towards $q_{\lambda}^{(L_{\text{FP}})}$, and a dual \gls{ot} head $f_{\psi}(z,k,\lambda)$, taking the embeddings of $(z, k, \lambda)$ and producing \gls{ot} objects, such as conditional \gls{ot} plans and barycentric mappings. We introduced two losses, $\mathcal{L}_{\text{D}}(\psi, \lambda)$ and $\mathcal{L}_{\text{C}}(\theta, \lambda)$, which are optimized w.r.t. neural net parameters $(\theta, \psi)$,
\begin{equation}
    \begin{aligned}
        \psi^{\star} = \argmax{\psi\in\Psi}\,
    \mathbb{E}_{\lambda\sim\text{Dir}(\alpha)}[\mathcal{L}_{D}(\psi,\lambda)],
    \qquad
    \theta^{\star} = \argmin{\theta\in\Theta}\,
    \mathbb{E}_{\lambda\sim\text{Dir}(\alpha)}[\mathcal{L}_{C}(\theta,\lambda;\psi^{\star})],
    \end{aligned}\label{eq:nuabfm}
\end{equation}
where the expectation over $\lambda \sim \text{Dir}(\alpha)$ ensures $v_{\theta}$ and $f_{\psi}$ are trained over the whole simplex. Our training strategy is summarized in Algorithm~\ref{alg:nuabfm}. $(\theta,\psi)$ are learned jointly via block-coordinate updates. At each iteration, $\psi$ is updated to maximize $\mathcal{L}_{D}$, which is independent of $\theta$, and $\theta$ is updated to minimize $\mathcal{L}_{C}$ for a fixed $\psi$, which boils down to conventional \gls{ot}-\gls{cfm}.

\begin{algorithm}[th]
\caption{\texttt{BaryFM}: Barycentric Flow Matching Training}
\label{alg:nuabfm}
\begin{algorithmic}[1]
\Require Measures $p_{1},\cdots,p_{K}$, velocity field $v_\theta$, dual potential $f_\psi$, batch size $B$, iterations $N$, \# of fixed-point iterations $L_{\text{FP}}$, \# of dual iterations $L_{\text{Dual}}$, entropic regularization $\epsilon$
\For{$\text{iter} = 1, \dots, N$}
    \State Sample $\lambda \sim \text{Dir}(\alpha)$ and $\{\{z_{k,j}\}_{j=1}^{B}\}_{k=1}^{K}$, where $z_{k,j}\sim p_{k}$
    \State Initialize $\bar{z}^{(0)}_{i} \leftarrow \sum_{k=1}^{K} \lambda_k z_{k,i}$
    \For{$\ell=1,\cdots,L_{\text{FP}}$}
        \State $\bar{z}^{(\ell)}_{i} \leftarrow \sg\Bigl(\sum_{k=1}^{K} \lambda_{k} \hat{T}_{\epsilon,\psi}(\bar{z}_{i}^{(\ell-1)},k,\lambda)\Bigr)$ \Comment{No backprop through $\psi$}
    \EndFor
    \For{$s = 1, \dots, L_{\text{Dual}}$}
        \State $\mathcal{L}_{\text{D}} \leftarrow \sum_{k=1}^{K}\lambda_{k}\biggr(\mathbb{E}_{z}[f_{\epsilon,\psi}(z,k,\lambda) ] + \mathbb{E}_{\bar z}[f_{\epsilon,\psi}^{(c)}(\bar z,k,\lambda)] - \epsilon(\mathbb{E}_{z,\bar{z}}[\gamma_{\epsilon,\psi}(z,\bar{z},k,\lambda)]-1)\biggr)$,
        \State $\psi \leftarrow \psi + \text{LR}_{D}\nabla_{\psi}\mathcal{L}_{\text{D}}$
    \EndFor
    \State $\mathcal{L}_{\text{C}}\leftarrow \mathbb{E}_{t\sim \mathcal{U}[0,1],k\sim\text{Cat}(\lambda),  z_0 \sim p_k, z_1 \sim \gamma_{\epsilon,\psi}(\cdot|z_0,k,\lambda)}\biggl[ \lVert v_{\theta}(z_t, t, k, \lambda) - (z_1 - z_0) \rVert_{2}^{2} \biggr]$
    \State $\theta \leftarrow \theta - \text{LR}_{\text{C}}\, \nabla_\theta \mathcal{L}_{\text{C}}$
\EndFor
\State \Return $\theta,\, \psi$
\end{algorithmic}
\end{algorithm}

\begin{wrapfigure}[20]{r}{0.35\textwidth}
  \centering
  \includegraphics[width=\linewidth]{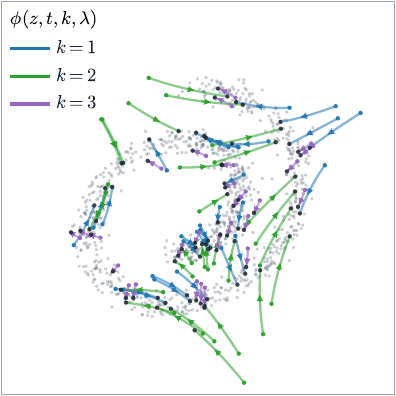}
  \caption{Inference. One ODE solve carries each $z \sim p_k$ (coloured) along the paths shown to $\phi(z,1,k,\lambda)$ (dark), landing on the barycenter support (gray).}
  \label{fig:flow-illustration}
\end{wrapfigure}
\noindent\textbf{Inference.} Once trained, we can sample from any measure in $\mathcal{W}(p_{1:K}) = \{\text{Bar}(\lambda, p_{1:K}):\lambda \in \Delta_{K}\}$ by using the velocity field $v_{\theta}(z,t,k,\lambda)$. For instance, using the forward Euler discretization of the \gls{ode} $\partial_{t}z(t) = v_{\theta}(z,t,k,\lambda)$, we have
\begin{equation}
\begin{aligned}
    &z_{r+1,i} = z_{r,i} + v_{\theta}(z_{r,i},r dt,k,\lambda)dt,\\
    &\text{where }r=0,\cdots,T-1\text{, }dt=T^{-1}\text{, and }z_{0,i} \sim p_k.
\end{aligned}\label{eq:ode-inference}
\end{equation}
In practice, we are integrating the velocity field with $\phi(z,t,k,\lambda) = z+\int_{0}^{t}v_{\theta}(\phi(z,s,k,\lambda),s,k,\lambda)ds$. An illustrative example is shown in Figure~\ref{fig:flow-illustration}. The main advantage of this technique is amortizing the sampling of the barycenter support with respect particles $z$, and barycentric coordinates $\lambda \in \Delta_{K}$, that is, our method is, by design, \emph{universal}.

\noindent\textbf{Flows over joint measures.} Neural methods so far have treated the case of barycenters over $\Omega = \mathbb{R}^{d}$. In this work we take a step forward, and define a flow over $\Omega = \mathbb{R}^{d}\times\Delta_{C}$, i.e., the space over features $\times$ soft-labels joint variables. This has been previously done for empirical solvers~\citep{montesuma2023multi,montesuma2025computing}, and is made possible in the continuous case by flow matching. We let $z=(x,y)$, so the flow is defined over $\Omega = \mathbb{R}^{d}$ and $\Delta_{C}$ with metric,
\begin{align}
    d(z,z')^2 = \lVert x - x' \rVert_{2}^{2} + \beta \lVert y - y' \rVert_{2}^{2},
\end{align}
where $\beta > 0$ is an importance factor for the label terms. Furthermore, since $\hat{T}_{\epsilon,\psi}$ is a convex combination, labels stay in the simplex when computing the \gls{cfm} targets. The only difference comes when solving the \gls{ode} in equation~\ref{eq:ode-inference}, where we add a projection step $\pi_{\Omega}(z_{r,i} + v_{\theta}(z_{r,i}, r dt, k, \lambda)dt)$, where $\pi_{\Omega}(z) = \pi_{\Omega}(x,y) = (x, \pi_{\Delta_{C}}(y))$, and $\pi_{\Delta_{C}}(y)$ denotes the orthogonal projection into $\Delta_{C}$.

\section{Experiments}\label{sec:experiments}

Our experiments compare different barycenter solvers over downstream tasks. In the main paper, we cover domain adaptation (Section~\ref{sec:exp-da}) and generalization~ (Section~\ref{sec:exp-dg}), following the methodologies established in~\cite{montesuma2021wasserstein} and~\cite{wang2020heterogeneous}. In our appendix, we extend our experiments to the Swiss roll measures of~\cite{korotin2021continuous} (Section~\ref{sec:swiss-roll}), Bayesian posterior aggregation (Section~\ref{sec:additional-bpa}) and algorithmic fairness (Section~\ref{sec:additional-fairness}).

\subsection{Domain Adaptation}\label{sec:exp-da}

\begin{figure}[ht]
    \centering
    \begin{subfigure}{0.45\linewidth}
        \centering
        \includegraphics[width=\linewidth]{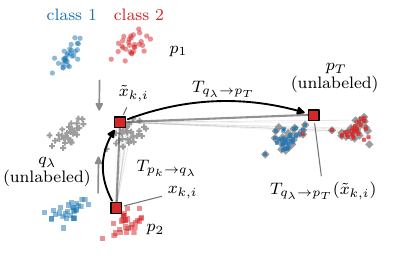}
        \caption{Unsupervised WBT}
    \end{subfigure}
    \begin{subfigure}{0.45\linewidth}
        \centering
        \includegraphics[width=\linewidth]{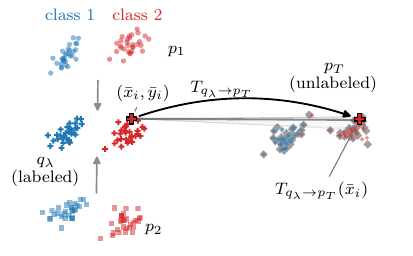}
        \caption{Supervised WBT}
    \end{subfigure}
    \caption{Unsupervised (a, $\mathcal{X} \times \mathcal{Y} = $\xmark), and supervised (b, $\mathcal{X}\times\mathcal{Y}=$\cmark) \gls{wbt} methodologies.}
    \label{fig:WBT-comparison}
\end{figure}

In this experiment, we test the ability of Wasserstein barycenter solvers in the context of multi-source domain adaptation~\citep{sun2015survey}, where a model is learned from multiple source domains $p_{1:K}$, and tested in an unlabeled target domain $p_{T}$, hence, the i.i.d. hypothesis underlying \gls{erm}~\citep{vapnik2013nature} does not hold, a case known as distribution or dataset shift~\citep{quinonero2008dataset}. We follow the \gls{wbt} pipeline~\citep{montesuma2021wasserstein}, a method that aligns the multiple sources to the target via an intermediate barycentric domain. This pipeline was originally proposed for synthesized, labeled barycenter. For comparing unsupervised barycenters, we use the variant proposed in~\cite{montesuma2025computing}. An example is shown in Figure~\ref{fig:WBT-comparison}. Therefore, besides ranking the alignment proposed by barycenter solvers, this task rewards barycenters that respect the underlying class structure.

We test barycenter methods in 10 benchmarks, along \textcolor{myblue}{computer vision}, \textcolor{myorange}{neuroscience} and \textcolor{myteal}{chemical engineering}: \textcolor{myblue}{Office 31}~\citep{saenko2010adapting}, \textcolor{myblue}{Office Home}~\citep{venkateswara2017deep}, \textcolor{myblue}{Domain Net}~\citep{peng2019moment}, \textcolor{myorange}{BCI-CIV-2a}~\citep{brunner2008bci}, \textcolor{myorange}{FACED}~\citep{chen2023large}, \textcolor{myorange}{SEED-VIG}~\citep{zheng2017multimodal}, \textcolor{myorange}{SHU-MI}~\citep{ma2022large},
\textcolor{myorange}{Mumtaz}~\citep{mumtaz2016eeg}, \textcolor{myorange}{Physio}~\citep{schalk2004bci2000}, \textcolor{myteal}{TEP}~\citep{reinartz2021extended,montesuma2024benchmarking}. In total, we test 11 methods, with a total of 14 variants and the source-only baseline. Empirical solvers: discrete~\citep{cuturi2014fast} and WGF~\citep{montesuma2025computing}. Neural solvers: NWB~\citep{fan2020scalable}, CW2B~\citep{korotin2021continuous}, WIN~\citep{korotin2022wasserstein}, NOT~\citep{kolesov2024estimating}, U-NOT~\citep{gazdieva2024robust} and NormFlow~\citep{visentin2025computing}. Diffusion based: TDSB~\citep{noble2023tree} and TSBM~\citep{howard2025schrdinger}. Parametric: GMM~\citep{montesuma2024lighter}. We show the average rank of barycenter methods on these benchmarks in Figure~\ref{fig:ranking-msda}, and report the detailed results in Appendix~\ref{sec:additional-da}.

\begin{figure}[ht]
    \centering
    \includegraphics[width=\linewidth]{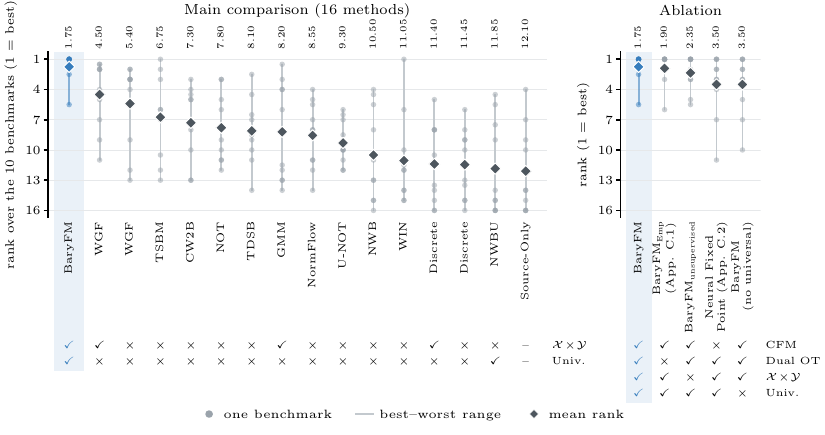}
    \caption{Comparison of \texttt{BaryFM} with 15 methods, and ablation of \texttt{BaryFM} components. We report the average rank over 10 benchmarks. Detailed results are shown in Table~\ref{tab:results-msda} and Appendix~\ref{sec:additional-da}.}
    \label{fig:ranking-msda}
\end{figure}

From Figure~\ref{fig:ranking-msda} (left), we see that \texttt{BaryFM} achieves the best average rank over the tested benchmarks, followed by \texttt{WGF}~\citep{montesuma2025computing}. Our \texttt{BaryFM} consistently achieves top 1 in 6 out of 10 benchmarks (c.f. Table~\ref{tab:results-msda}), being consistent on all benchmarks. In Figure~\ref{fig:ranking-msda} (right), we ablate different components of \texttt{BaryFM}, especially each head, and the universal training. We go into further detail on these variants in Appendix~\ref{appx:alternatives}. Overall, the components that have the most impact in \texttt{BaryFM} performance are the \gls{cfm} head, which amortizes the sampling from $q_{\lambda}^{\star}$, and universality.

\subsection{Domain Generalization}\label{sec:exp-dg}

Domain generalization~\citep{gulrajani2020search} seeks to generalize to an unseen domain from multi-domain source data $p_{1:K}$, with no target samples available. We tackle it through \emph{data augmentation}\citep{zhong2025survey}, that is, we generate data blending the source domains, and minimize
\begin{align}
    \theta^{\star} = \argmin{\theta \in \Theta}\dfrac{1-\rho}{K}\sum_{k=1}^{K}\dfrac{1}{n_k}\sum_{i=1}^{n}\ell(y_{k,i}, f_{\theta}(x_{k,i})) + \dfrac{\rho}{mK'}\sum_{k' =1}^{K' }\sum_{j=1}^{m}\ell(\tilde{y}_{k' ,j}, f_{\theta}(\tilde{x}_{k' ,j})),
\end{align}
where $\ell$ is a loss function, $f_{\theta}$ is a neural net, such as a ResNet 50~\citep{he2016deep}, and $(\tilde{x}_{k',j }, \tilde{y}_{k', j})$ is an augmented sample. In this experiment the augmented samples are obtained either through (1) MixUp~\citep{zhang2017mixup}, (2) Domain MixUp~\citep{wang2020heterogeneous}, (3) BaryFM$_{\text{Random}}$ or (4) BaryFM$_{\text{Vertex}}$ (both ours). In the first case, $\lambda_{k'} \sim \text{Beta}(\alpha, \alpha)$, the 2nd and 3rd cases $\lambda_{k'} \sim \text{Dir}(\alpha)$. The 4th case is different, as we sample from the vertices of $\mathcal{W}(p_{1:K})$, meaning, $\lambda_{k'} = e_{k' }$, $k' = 1,\cdots, K$. More importantly, MixUp and its variants mixes samples linearly, while BaryFM mixes samples nonlinearly through Wasserstein barycenters. Using Wasserstein barycenters for the augmentation task demands 3 properties of solvers: (i) \emph{amortization over samples}, meaning, we must sample $z \sim \hat{q}_{\lambda}^{(L_{\text{FP}})}$, (ii) \emph{universality}, meaning, we must be able to sample from $\hat{q}_{\lambda}^{(L_{\text{FP}})},\forall \lambda \in \Delta_{K}$ and (iii) joint barycenters, meaning, ours samples must be $z = (x, y)$, $x \in \mathbb{R}^{d}$, $y \in \Delta_{C}$. To the best of our knowledge, \texttt{BaryFM} is the only solver satisfying all three (Table~\ref{tab:sota-comparison}).

To better capture the semantics of complex data, we perform augmentations in a latent space. For image data, we use a \gls{vae}~\citep{kingma2013auto,rombach2022high}, while for EEG we use \gls{pca} fit on source domains. An illustration of our approach is shown in Figure~\ref{fig:augmentation-dg}. Further details are available in Appendix~\ref{sec:additional-dg}.

\begin{figure}[ht]
    \centering
    \includegraphics[width=0.85\linewidth]{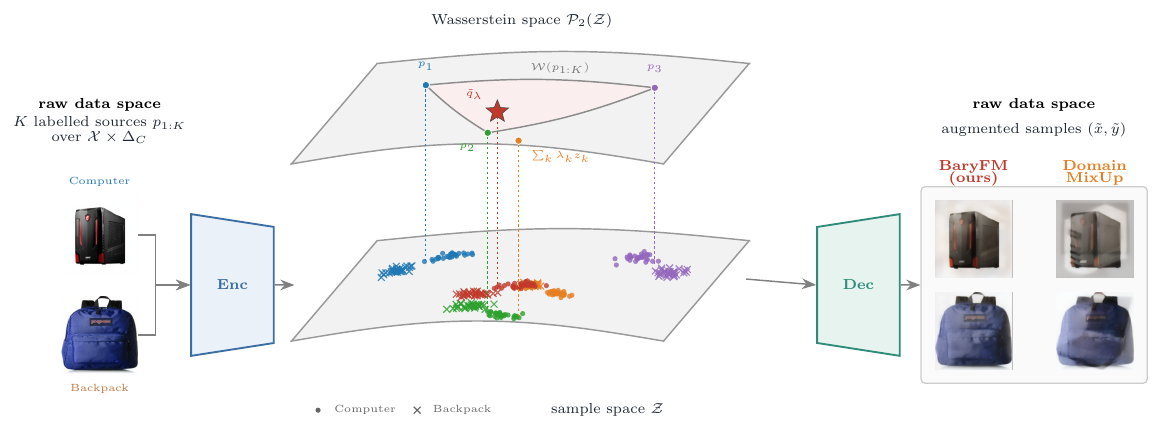}
    \caption{Data augmentation strategy for domain generalization. On the left, samples from different domains are encoded into a common latent space. In the center, synthetic samples are generated in the Wasserstein simplex of source domains. On the right, the latent synthetic samples are decoded back into raw data space.}
    \label{fig:augmentation-dg}
\end{figure}

Our results are summarized in Table~\ref{tab:dg-results}, which reports the downstream performance of ResNets~\citep{he2016deep} and CBraMod~\citep{wang2025cbramod} finetuned on 5 multi-domain benchmarks. On average, BaryFM$_{\text{Random}}$ ranks best (1.40), against BaryFM$_{\text{Vertex}}$ (2.80), Domain MixUp (3.00), MixUp (3.20) and ERM (4.60). Overall, these results illustrate a couple of findings. First, augmentations broadly help generalization to new domains. Second, interpolating under the Wasserstein geometry (our BaryFM vs. MixUp variants) generalizes better. Third, sampling from the Wasserstein simplex (BaryFM$_{\text{Random}}$ vs. BaryFM$_{\text{Vertex}}$) works better overall. These findings illustrate the usefulness of an \emph{universal, amortized, joint} barycenter algorithm for domain generalization.

\begin{table}[ht]
    \centering
\resizebox{\linewidth}{!}{
    \begin{tabular}{lcccccc}
        \toprule
        Method & Office 31 & Office Home & DomainNet & BCI-CIV-2a & SEED-VIG & Avg. Rank \\
        \midrule
        Backbone & ResNet 50 & ResNet 50 & ResNet 101 & CBraMod & CBraMod & -\\
        Latent Space & \multicolumn{3}{c}{\gls{vae}} & \multicolumn{2}{c}{PCA} & -\\
        \midrule
        ERM & \result{84.38}{0.53} & \result{69.74}{0.57} & \result{49.57}{0.24} & \result{50.63}{3.45} & \result{58.30}{2.44} & 4.60\\
        MixUp & \result{85.66}{0.69} & \result{70.43}{0.34} & \result{49.53}{0.11} & \result{52.37}{3.11} & \result{61.11}{2.57} & 3.20\\
        Domain MixUp & \result{84.97}{0.49} & \result{68.86}{0.82} & \textbf{\result{50.63}{0.20}} & \result{52.70}{2.78} & \result{61.04}{3.15} & 3.00\\
        \midrule
        BaryFM$_{\text{Vertex}}$ & \result{84.89}{0.58} & \result{70.55}{0.47} & \result{49.69}{0.34} & \textbf{\result{54.18}{3.48}} & \result{59.25}{5.63} & 2.80\\
        BaryFM$_{\text{Random}}$ & \textbf{\result{85.94}{0.66}} & \textbf{\result{71.17}{0.76}} & \result{50.19}{0.21} & \result{53.73}{3.43} & \textbf{\result{62.08}{2.38}} & 1.40\\
        \bottomrule
    \end{tabular}
}
    \caption{Downstream performance of data augmentation strategies on domain generalization benchmarks. Results denote $\mu_{\pm \sigma}$ over seeds (c.f. Section~\ref{sec:additional-dg}).}
    \label{tab:dg-results}
\end{table}

\section{Conclusion}\label{sec:conclusion}

In this paper, we study the \emph{universal Wasserstein barycenter problem}, that is, approximating the family of measures in the \emph{Wasserstein simplex} (c.f. Equation~\ref{eq:wasserstein-simplex}), which contains the Wasserstein barycenters $\text{Bar}(\lambda, p_{1:K})$ for general $\lambda \in \Delta_{K}$. We propose a new method based on a neural network, composed of a shared encoding layer, and 2 heads (c.f. Figure~\ref{fig:arch}): a dual \gls{ot} head approximating the \gls{ot} problem between each $p_k$ and the barycenter target $q_{\lambda}^{\star}$, and a flow matching~\citep{lipman2022flow} head, modeling the flow between the aforementioned measures. To the best of our knowledge, our method is the first universal, neural method that can handle flows over feature-label joint measures.

We validate our proposed \texttt{BaryFM} on 4 downstream applications of Wasserstein barycenters, namely, domain adaptation~\citep{montesuma2021wasserstein} (c.f. Section~\ref{sec:exp-da}), generalization (c.f. Section~\ref{sec:exp-dg}), algorithmic fairness~\citep{gordaliza2019obtaining} (c.f. Appendix~\ref{sec:additional-fairness}) and Bayesian posterior aggregation~\citep{srivastava2018scalable} (c.f. Appendix~\ref{sec:additional-bpa}). Overall, in domain adaptation, our method has the best average rank among 15 tested methods, over 10 benchmarks. Our method enables a new application of barycenter solvers, that is domain generalization. We demonstrate empirically that augmenting the source domain data with Wasserstein barycenters improves performance over the \gls{erm} baseline, and is better, or competitive against Domain MixUp~\citep{wang2020heterogeneous}.

\noindent\textbf{Limitations.} As we cover in Appendix~\ref{appx:related-work}, our method computes a \emph{regularized} barycenter, which carries inherent biases with respect the true Wasserstein barycenter. While some of these biases (mini-batching, entropic regularization) serve as a way of making the barycenter problem tractable, further analysis on how the computed barycenter differs from the ground-truth is needed, and is the subject of future work. In restricted cases (e.g., Location-Scatter families in Appendix~\ref{sec:swiss-roll}) we show that our method remains accurate to a known ground-truth despite these biases.

\newpage

\bibliography{refs}
\bibliographystyle{arxiv}

\newpage

\appendix

\begin{center}
    {\LARGE\bfseries Supplementary Materials for\\[0.3em]
    Towards Universal Wasserstein Barycenters through Flow Matching\par}
    \vspace{1.5em}
\end{center}

\tableofcontents

\newpage

\section{Introduction}\label{sec:intro}

In this appendix, we provide much more information about the implementation, hyper-parameters, and experiments done in our work. Our appendix is organized as follows. Section~\ref{appx:related-work} discusses how our algorithm fits within the broader literature of Wasserstein barycenters. Section~\ref{appx:alternatives} provides the description of 2 alternative proposed methods to \texttt{BaryFM}, namely \emph{empirical} \texttt{BaryFM} and \texttt{NeuralFP}. These methods are compared in the domain adaptation ablation in Table~\ref{tab:ablation}. Next, we denote 5 sections to additional experimental details. Section~\ref{sec:swiss-roll} covers the Swiss roll toy example. Section~\ref{sec:additional-bpa} covers Bayesian Posterior Aggregation. Section~\ref{sec:additional-fairness} covers algorithmic fairness. Section~\ref{sec:additional-da} covers domain adaptation. Section~\ref{sec:additional-dg} covers domain generalization. In each of these sections, we provide details about how barycenters are applied to these tasks, as well as detailed results.

\section{Related Work}\label{appx:related-work}

\paragraph{Empirical Barycenter Solvers.} Empirical barycenter solvers parametrize the barycentric measure, $q_{\lambda}^{\star}$ via the samples in their support. This hinders the possibility of drawing new samples from this measure, as each new drawn requires solving the optimization problem from scratch, which by it self limits their scalability. Besides this issue, early algorithms such as~\cite{cuturi2014fast} and~\cite[Algorithm 1]{montesuma2023multi} assume access to the complete set of samples in the marginal measures. This issue has been partly solved by~\cite{montesuma2025computing} who proposed a gradient flow-based algorithm that supports mini-batching.

\paragraph{Parametric Barycenter Solvers.} These include Gaussian and Gaussian Mixture approximations to the concerned measures. In the first case,~\cite{agueh2011barycenters} and~\cite{alvarez2016fixed} show that the barycenter can be fold via an iterative fixed-point algorithm with complexity that depends on the dimensionality $\mathcal{O}(d^3)$. While this complexity is sample-free (thus scalable to large-scale datasets), it scales poorly with the data dimensionality.~\cite{montesuma2024lighter} presents a different approach based on \gls{gmm} \gls{ot}~\citep{chen2018optimal,delon2020wasserstein}. In practice, the authors propose a method based on \glspl{gmm} with diagonal covariances, which scales both with number of samples and dimensionality. Nonetheless, as with all parametric models, these algorithms rely on strong hypothesis which are seldom met in real datasets.

\paragraph{Neural Barycenter Solvers.} These methods include a recent wave of works in the intersection of machine learning and \gls{ot}. As such, they rely on different strategies. We cover mainly 3 families of strategies. The first family relies on multi-level optimization. For instance~\cite{fan2020scalable} propose a 3-loop min-max-min problem with $2K+1$ \glspl{icnn}.~\cite{korotin2021continuous} avoid the 3-level optimization problem, by combining congruence and cycle-consistency soft constraints over $2K$ \glspl{icnn}.

The second family relies on a neural parametrization of the barycenter measure itself.~\cite{korotin2022wasserstein} represents $q_{\lambda}^{(\ell)}$ as the map from a latent space to the barycentric measure. The optimization problem proposed by the authors is a fixed-point strategy inspired in~\cite{alvarez2016fixed}.~\cite{kolesov2024estimating} and~\cite{gazdieva2024robust} learn transport maps paired with potentials, under a bi-level max-min objective, the latter being unbalanced by design.

A third family, more similar to our proposed \texttt{BaryFM}, parametrizes the dynamics from a initial measure $q_0$ towards the barycenter $q_{\lambda}^{\star}$. For instance,~\cite{noble2023tree} and~\cite{howard2025schrdinger} view the barycenter problem through the lens of a start-shaped tree \gls{sbm}, and~\cite{visentin2025computing} learn a conditional normalizing flow.

\paragraph{Universal Barycenter Solvers.} All of the aforementioned families share a limitation: for each new set of barycentric coordinates $\lambda \in \Delta_{K}$, the corresponding optimization problems need to be solved \emph{from scratch}. For instance, in~\cite{korotin2022wasserstein}, $\lambda$ enters the regression targets. In~\cite{noble2023tree} and~\cite{howard2025schrdinger}, it enters the \gls{sbm} problem itself. To the best of our knowledge, the only prior \emph{universal} solver is~\cite[Appendix B]{fan2020scalable}, who adds $\lambda$ as an input to the $2K+1$ \glspl{icnn}. The authors validated their approach on low-dimensional, toy examples.

We summarize our discussion in Table~\ref{tab:sota-comparison}, which shows that most existing methods do not consider the question of approximating barycenters over the Wasserstein simplex. Based on the limitations of existing methods, and in order to make the overall architecture scalable with respect number of marginals $K$, samples $n$ and \emph{universal} we use a single network with 2 heads, whose outputs are conditioned on $(k, \lambda)$, $k \in \{1,\cdots,K\}$ and $\lambda \in \Delta_{K}$. The closest precedent in the literature is~\cite{visentin2025computing}, who condition their learned \gls{ot} maps via a domain-index $k$. While this is similar in spirit to our architecture, the conditioning maps a latent measure to each marginal $p_k$. Effectively, this means that while our conditioning expresses the \gls{ot} between $(p_k, q_{\lambda}^{\star})$, theirs concern $(q_0, p_k)$, where $q_0 \in \mathcal{P}_{2}(\Omega)$ for a latent space $\Omega$.

\begin{table}[ht]
    \centering
    \resizebox{\linewidth}{!}{
    \begin{tabular}{lllccccccc}
        \toprule
        Method & Reference & Parametrization & Learned object & \# nets & \# params & Inner opt. & Amort. sampling & Universal & $\mathcal{X}\times\mathcal{Y}$ \\
        \midrule
        \multicolumn{10}{l}{\textit{Free-support (empirical)}}\\
        \texttt{Discrete} & \citep{cuturi2014fast} & Particles & support & --- & $nd$ & --- & \xmark & \xmark & \xmark \\
        \texttt{DaDiL} & \citep{montesuma2023multi} & Particles & support (atoms) & --- & $nd$ & --- & \xmark & \xmark & \cmark \\
        \texttt{WGF} & \citep{montesuma2025computing} & Particles & support & --- & $nd$ & --- & \xmark & \xmark & \cmark \\
        \midrule
        \multicolumn{10}{l}{\textit{Restricted parametric family}}\\
        \texttt{GMM-WBT} & \citep{montesuma2024lighter} & GMM & means, covariances & --- & $md^{2}$ & --- & \cmark & \xmark & \cmark \\
        \midrule
        \multicolumn{10}{l}{\textit{Neural: dual potentials}}\\
        \texttt{NWB} & \citep{fan2020scalable} & \glspl{icnn} & potentials, map & $2K+1$ & $(2K+1)P$ & min-max-min & \cmark & \xmark & \xmark \\
        \texttt{NWBU} & \citep[App. B]{fan2020scalable} & \glspl{icnn} & potentials, map & $2K+1$ & $(2K+1)P$ & min-max-min & \cmark & \cmark & \xmark \\
        \texttt{CW2B} & \citep{korotin2021continuous} & \glspl{icnn} & potentials & $2K$ & $2KP$ & --- & \cmark & \xmark & \xmark \\
        \midrule
        \multicolumn{10}{l}{\textit{Neural: measure and maps}}\\
        \texttt{WIN} & \citep{korotin2022wasserstein} & NNs & generator, maps & $2K+1$ & $(2K+1)P$ & min-max-min & \cmark & \xmark & \xmark \\
        \texttt{NOT} & \citep{kolesov2024estimating} & NNs & maps, potentials & $2K$ & $2KP$ & max-min & \cmark & \xmark & \xmark \\
        \texttt{U-NOT} & \citep{gazdieva2024robust} & NNs & maps, potentials & $2K$ & $2KP$ & max-min & \cmark & \xmark & \xmark \\
        \midrule
        \multicolumn{10}{l}{\textit{Neural: dynamics and flows}}\\
        \texttt{TDSB} & \citep{noble2023tree} & NNs & \gls{sde} drifts & $2K$ & $2KP$ & --- & \cmark & \xmark & \xmark \\
        \texttt{TSBM} & \citep{howard2025schrdinger} & NNs & vector fields & $2K$ & $2KP$ & --- & \cmark & \xmark & \xmark \\
        \texttt{NormFlow} & \citep{visentin2025computing} & NNs & bijections & $1$ & $P + \mathcal{O}(KH)$ & --- & \cmark & \xmark & \xmark \\
        \midrule
        \multicolumn{10}{l}{\textit{Ours}}\\
        \texttt{NeuralFP} & (ours) & NNs & dual potential & $1$ & $P+\mathcal{O}(KH)$ & --- & \xmark & \cmark & \cmark \\
        \texttt{BaryFM} & (ours) & NNs & dual, velocity & $1$ & $P+\mathcal{O}(KH)$ & --- & \cmark & \cmark & \cmark \\
        \bottomrule
    \end{tabular}
    }
    \caption{Comparison of different approaches in the state-of-the-art along parametrization, objects, number of neural nets and parameters, optimization strategy, amortization, universality, and handling of joint measures.}
    \label{tab:sota-comparison}
\end{table}

\paragraph{A Taxonomy of Biases.} Given the different existing solvers and their inner workings, we provide, in Table~\ref{tab:bias-taxonomy}, a taxonomy of the different \emph{biases} in the barycenter solvers. We understand as biases, different design choices that make algorithms compute a barycenter $\hat{q}_{\lambda}^{(L_{\text{FP}})}$ that is different from the ground-truth $q_{\lambda}^{\star}$. We divide these biases into 3 categories,
\begin{itemize}
    \item \textbf{Input biases} concern the way the marginal measures are approximated. These include mini-batch \gls{ot}~\citep{fatras2019learning} and parametric \gls{ot} (e.g., Gaussian or \glspl{gmm}).
    \item \textbf{Algorithmic biases} include several factors, such as entropic \gls{ot}~\citep{cuturi2013sinkhorn}, unbalanced \gls{ot}~\citep{sejourne2023unbalanced}, expressing the barycenter as a push-forward of a latent measure (as in~\cite{korotin2022wasserstein}), relaxing constraints (as in~\cite{korotin2021continuous}) or fixed-point truncation (most methods based on~\cite{alvarez2016fixed}).
    \item \textbf{Output biases} which roughly include the time-discretization of an \gls{ode} as in ours and~\cite{montesuma2025computing}.
\end{itemize}

\begin{table}[ht]
    \centering
    \resizebox{\linewidth}{!}{
    \begin{tabular}{llcccccccc}
        \toprule
        & & \multicolumn{2}{c}{Input} & \multicolumn{5}{c}{Algorithmic} & Output \\
        \cmidrule(lr){3-4}\cmidrule(lr){5-9}\cmidrule(lr){10-10}
        Method & Reference
        & \shortstack{Mini-Batch\\OT}
        & \shortstack{Parametric\\OT}
        & \shortstack{Entropic\\OT}
        & \shortstack{Unbalanced\\OT}
        & \shortstack{Latent\\Pushforward}
        & \shortstack{Constraint\\Relaxation}
        & \shortstack{Fixed-Pt.\\Truncation}
        & \shortstack{Time\\Discretization} \\
        \midrule
        \texttt{Discrete} & \citep{cuturi2014fast} & \xmark & \xmark & \cmark (\xmark\,$\text{ }\epsilon = 0$) & \xmark & \xmark & \xmark & \cmark & \xmark \\
        \texttt{WGF} & \citep{montesuma2025computing} & \cmark & \xmark & \cmark (\xmark $\text{ }\epsilon = 0$)  & \xmark & \xmark & \xmark & \xmark & \cmark \\
        \texttt{GMM} & \citep{montesuma2024lighter} & \xmark & \cmark & \xmark & \xmark & \xmark & \xmark & \cmark & \xmark \\
        \midrule
        \texttt{NWB} & \citep{fan2020scalable} & \xmark & \xmark & \xmark & \xmark & \cmark & \xmark & \xmark & \xmark \\
        \texttt{NWBU} & \citep[App. B]{fan2020scalable} & \xmark & \xmark & \xmark & \xmark & \cmark & \xmark & \xmark & \xmark \\
        \texttt{CW2B} & \citep{korotin2021continuous} & \xmark & \xmark & \xmark & \xmark & \xmark & \cmark & \xmark & \xmark \\
        \texttt{WIN} & \citep{korotin2022wasserstein} & \xmark & \xmark & \xmark & \xmark & \cmark & \xmark & \cmark & \xmark \\
        \texttt{NOT} & \citep{kolesov2024estimating} & \xmark & \xmark & \xmark & \xmark & \xmark & \xmark & \xmark & \xmark \\
        \texttt{U-NOT} & \citep{gazdieva2024robust} & \xmark & \xmark & \xmark & \cmark & \xmark & \xmark & \xmark & \xmark \\
        \texttt{TDSB} & \citep{noble2023tree} & \xmark & \xmark & \cmark & \xmark & \xmark & \xmark & \cmark & \cmark \\
        \texttt{TSBM} & \citep{howard2025schrdinger} & \xmark & \xmark & \cmark & \xmark & \xmark & \xmark & \cmark & \cmark \\
        \texttt{NormFlow} & \citep{visentin2025computing} & \xmark & \xmark & \xmark & \xmark & \xmark & \cmark & \xmark & \xmark \\
        \midrule
        \texttt{NeuralFP} & (ours) & \cmark & \xmark & \cmark & \xmark & \xmark & \xmark & \cmark & \xmark \\
        \texttt{BaryFM} & (ours) & \cmark & \xmark & \cmark & \xmark & \xmark & \xmark & \cmark & \cmark \\
        \bottomrule
    \end{tabular}
    }
    \caption{Taxonomy of barycenter solvers biases, along input, algorithmic and output biases.}
    \label{tab:bias-taxonomy}
\end{table}

\newpage

\section{Alternative Versions of \texttt{BaryFM}}\label{appx:alternatives}

As described in section~\ref{sec:methodology}, our method is based on 2 pieces, a dual \gls{ot} estimator head, and a flow matching head. In our experiments, we ablate these 2 components. In this section, we briefly describe other 2 methods,
\begin{itemize}
    \item Empirical \texttt{BaryFM} (Section~\ref{sec:empirical-baryfm}), where we replace the dual estimation head by an inner \gls{ot} problem approximating the barycenter,
    \item \texttt{NeuralFP} (Section~\ref{sec:neural-fp}), where we remove the flow matching component of our algorithm and use the dual \gls{ot} head in the fixed-point strategy of~\cite{alvarez2016fixed}.
    \item Semi-Unbalanced \texttt{BaryFM} (Section~\ref{sec:su-baryfm}), where we change the semi-dual \gls{ot} loss with the semi-unbalanced objective developed in~\cite{gazdieva2024robust}.
\end{itemize}
We summarize the different versions of our method in Table~\ref{tab:ablation-summary}. Note that these different formulations make an ablation ladder, starting from the discrete barycenter of~\cite{cuturi2014fast}, and progressively converging towards \texttt{BaryFM}. We highlight the advantages of each component,
\begin{itemize}
    \item The \textbf{Dual \gls{ot} Head} allows us to amortize \gls{ot} between mini-batches from $p_k$ and the moving target $q_{\lambda}^{\star}$. Therefore, instead of computing $K-$\gls{ot} problem at each optimization step, with complexity $\mathcal{O}(Kn^{2})$, we are able to compute the \gls{ot} variables with a forward pass of a neural network.
    \item The \textbf{Velocity Head} allows us to amortize the \emph{sampling} of $q_{\lambda}^{\star}$, for an arbitrary $\lambda \in \Delta_{K}$. This sampling is efficient, since it relies on solving an \gls{ode}. In general, it requires $T$ forward passes of the velocity field (c.f. equation~\ref{eq:ode-inference}).
\end{itemize}
Besides these components, we explore in Section~\ref{sec:su-baryfm} a semi-unbalanced version of our method inspired by~\cite{gazdieva2024robust}.

\begin{table}[ht]
    \centering
    \begin{tabular}{lccc}
        \toprule
        Method & Conditional Flow Matching & Neural Dual Estimation & Balanced\\
        \midrule
        Discrete Barycenter & \xmark & \xmark & \cmark\\
        \midrule
        \texttt{NeuralFP} & \xmark & \cmark & \cmark\\
        Empirical \texttt{BaryFM} & \cmark & \xmark & \cmark\\
        \texttt{BaryFM} & \cmark & \cmark & \cmark\\
        \midrule
        Semi-unbalanced \texttt{BaryFM} & \cmark & \cmark & \xmark \\
        \bottomrule
    \end{tabular}
    \caption{Summary of different methods proposed in this paper. The discrete barycenter corresponds to the algorithm of~\cite{cuturi2014fast}, which does not use neither \gls{cfm} nor the dual \gls{ot} head.}
    \label{tab:ablation-summary}
\end{table}

\subsection{Empirical \texttt{BaryFM}}\label{sec:empirical-baryfm}

This version of our algorithm estimates the barycenter targets with empirical \gls{ot}. In a nutshell, we initialize the barycenter measure with $\bar{z}_i^{(0)} = \sum_{k=1}^{K}\lambda_k z_{k,i}$, where $z_{k,i} \sim p_k$. Note that we sample a fixed number of $B$ samples from each marginal measure. For $\ell=1,\cdots,L$, we progressively update the support of $q_{\lambda}^{(\ell)}$ via the fixed-point approach of~\cite{alvarez2016fixed},
\begin{align}
    \underbrace{\bar{z}_{i}^{(\ell+1)} = \sum_{k=1}^{K}\lambda_{k}\hat{T}_{k}(\bar{z}_{i}^{(\ell)})}_{\text{Barycenter support update}}\text{, where }\underbrace{\hat{T}_{k}(\bar{z}_{i}^{(\ell)}) = \sum_{j=1}^{B}\frac{\gamma_{k,i,j}^{(\ell)}}{\sum_{j'=1}^{B}\gamma_{k,i,j'}^{(\ell)}}z_{k,j}}_{\text{Barycentric mapping}}\text{, and }\gamma_{k}^{(\ell)}=\text{OT}_{\epsilon}(p_{k},q_{\lambda}^{(\ell)}),\label{eq:emp-fixed-pt}
\end{align}
where $\text{OT}_{\epsilon}$ denotes the Sinkhorn problem~\citep{cuturi2013sinkhorn} between empirical measures $p_k$ with support $\{z_{k,j}\}_{j=1}^{B}$ and barycenter support $\{\bar{z}_{i}^{(\ell)}\}_{i=1}^{B}$. Note that equation~\ref{eq:emp-fixed-pt} is an empirical approximation to the mapping $G$ defined in equation~\ref{eq:fixed-point-ae}. Indeed, the barycentric mapping defines in equation~\ref{eq:emp-fixed-pt} is an approximation to the Monge map $T_k^{\star}$~\citep{deb2021rates}, as long as it exists (e.g., $p_k$ assumed to be absolutely continuous).

The computation of the \gls{cfm} targets has complexity $\mathcal{O}(KB^{2})$ per fixed-point iterations for both the neural and empirical solvers. The main difference is that, within the fixed-point iterations, the empirical solver needs to solve $K-$\gls{ot} problems, with a complexity that scales with the number of Sinkhorn iterations. Meanwhile, the neural solver requires a single forward pass, being able to estimate in \emph{one-shot} the \gls{ot} variables. Therefore, complexity generally reduces from $\mathcal{O}(L_{\text{FP}}L_{\text{Sinkhorn}}KB^{2})$ to $\mathcal{O}(L_{\text{FP}}KB^{2})$. We present the empirical version of \texttt{BaryFM} in Algorithm~\ref{alg:eubfm}.

\begin{algorithm}[ht]
\caption{Empirical Barycentric Flow Matching}
\label{alg:eubfm}
\begin{algorithmic}[1]
\Require Measures $\{p_k\}_{k=1}^K$, velocity field $v_\theta$,
         batch size $B$, iterations $N$, entropic regularization $\epsilon$
\For{$\text{iter} = 1, \dots, N$}
    \State Sample $\lambda \sim \text{Dir}(\alpha)$ and $\{\{z_{k,j}\}_{j=1}^{B}\}_{k=1}^{K}$, where $z_{k,j}\sim p_{k}$
    \State $\bar{z}^{(0)}_{i} \leftarrow \sum_{k=1}^{K} \lambda_k z_{k,i}$
    \For{$\ell=1,\cdots,L_{\text{FP}}$}
        \State $C_{k,i,j} \leftarrow \lVert \bar{z}_{i}^{(\ell-1)} - z_{k,j} \rVert_{2}^{2}$
        \State $\gamma_{1:K}^{(\ell)} \leftarrow \text{Sinkhorn}\left(a_{1:K},b, C_{1:K},\epsilon\right)$\Comment{$a_{k,i} = B^{-1}$, $b_{j} = B^{-1}$}
        \State $\bar{z}^{(\ell)}_{i} \leftarrow \sum_{k=1}^{K} \lambda_{k}\sum_{j=1}^{B}\tfrac{\gamma_{k,i,j}^{(\ell)}}{\sum_{j'=1}^{B}\gamma_{k,i,j'}^{(\ell)}}z_{k,j}$
    \EndFor
    \State $\mathcal{L}_{\text{C}}(\theta) = \mathbb{E}_{t\sim \mathcal{U}[0,1], k \sim \text{Cat}(\lambda), (z_{0},z_{1})\sim \gamma_{k}}\biggl[ \lVert v_{\theta}(z_t, t, k, \lambda) - (z_1 - z_0) \rVert_{2}^{2} \biggr]$
    \State $\theta \leftarrow \theta - \text{LR}_{\text{C}}\, \nabla_\theta \mathcal{L}_{\text{C}}(\theta)$
\EndFor
\State \Return $\theta$
\end{algorithmic}
\end{algorithm}

\subsection{Neural Fixed-Point}\label{sec:neural-fp}

We recall that the fixed-point operator of~\cite{alvarez2016fixed} is,
\begin{align*}
    G(q) = \biggr( \sum_{k=1}^{K}\lambda_k T_k^{\star} \biggr)_{\sharp} q,
\end{align*}
where $T_{k}^{\star}$ is the Monge map from $q$ to $p_k$. Since the barycentric map $\hat{T}_{\epsilon,\psi}$ is an approximation of the Monge map, when it exists~\citep{deb2021rates,pooladian2021entropic}, we devise a variant of our \texttt{BaryFM} method that relies exclusively on the dual \gls{ot} head, by approximating $G$ via,
\begin{align*}
    \hat{G}_{\epsilon,\psi}(q) = \biggr( \sum_{k=1}^{K}\lambda_k \hat{T}_{\epsilon,\psi}(\cdot,k,\lambda) \biggr)_{\sharp}q
\end{align*}
We summarize its learning algorithm in Algorithm~\ref{alg:neuralfp}.

\begin{algorithm}[ht]
\caption{\texttt{NeuralFP}: Neural Fixed-Point Barycenters. Sampling re-uses lines 4--8 with $n$ samples per marginal and any $\lambda \in \Delta_{K}$, returning $\{\bar{z}_{i}^{(L_{\text{FP}})}\}_{i=1}^{n}$.}
\label{alg:neuralfp}
\begin{algorithmic}[1]
\Require Measures $p_{1},\cdots,p_{K}$, dual potential $f_\psi$, batch size $B$, iterations $N$, \# of fixed-point iterations $L_{\text{FP}}$, \# of dual iterations $L_{\text{Dual}}$, entropic regularization $\epsilon$
\For{$\text{iter} = 1, \dots, N$}
    \State Sample $\lambda \sim \text{Dir}(\alpha)$ and $\{\{z_{k,j}\}_{j=1}^{B}\}_{k=1}^{K}$, where $z_{k,j}\sim p_{k}$
    \State $f_{k,j} \leftarrow f_{\epsilon,\psi}(z_{k,j},k,\lambda)$\Comment{Single pass: $f_{\epsilon,\psi}$ does not depend on $\bar{z}$}
    \State Initialize $\bar{z}^{(0)}_{i} \leftarrow \sum_{k=1}^{K} \lambda_k z_{k,i}$
    \For{$\ell=1,\cdots,L_{\text{FP}}$}
        \State $C_{k,i,j} \leftarrow c(z_{k,j},\bar{z}_{i}^{(\ell-1)})$
        \State $\gamma_{j\lvert i, k} \leftarrow \softmax_{c}\left(\nicefrac{(f_{k,j} - C_{k,i,j})}{\epsilon}\right)$\Comment{$\gamma_{\epsilon,\psi}(z_{k,j}|\bar{z}_{i}^{(\ell-1)},k,\lambda)$, eq.~\ref{eq:soft-bary-map}}
        \State $\bar{z}^{(\ell)}_{i} \leftarrow \sg\Bigl(\sum_{k=1}^{K} \lambda_{k}\sum_{j=1}^{B}\gamma_{j\lvert i,k}z_{k,j}\Bigr)$\Comment{No backprop through $\psi$}
    \EndFor
    \For{$s = 1, \dots, L_{\text{Dual}}$}
        \State $\mathcal{L}_{\text{D}} \leftarrow \sum_{k=1}^{K}\lambda_{k}\biggr(\mathbb{E}_{z}[f_{\epsilon,\psi}(z,k,\lambda) ] + \mathbb{E}_{\bar z}[f_{\epsilon,\psi}^{(c)}(\bar z,k,\lambda)] - \epsilon(\mathbb{E}_{z,\bar{z}}[\gamma_{\epsilon,\psi}(z,\bar{z},k,\lambda)]-1)\biggr)$,
        \State $\psi \leftarrow \psi + \text{LR}_{\text{D}}\nabla_{\psi}\mathcal{L}_{\text{D}}$
    \EndFor
\EndFor
\State \Return $\psi$
\end{algorithmic}
\end{algorithm}

Since this variant does not have a velocity field, inference is done via the fixed-point strategy in equation~\ref{eq:emp-fixed-pt}, with $\hat{T}_{k}$ replaced by the neural barycentric mapping $\hat{T}_{\epsilon,\psi}$,
\begin{equation}
    \begin{aligned}
        \bar{z}_{i}^{(\ell+1)} &= \sum_{k=1}^{K}\lambda_{k}\hat{T}_{\epsilon,\psi}(\bar{z}_{i}^{(\ell)},k,\lambda),\\
        \hat{T}_{\epsilon,\psi}(\bar{z}_{i}^{(\ell)}) &= \sum_{j=1}^{B}\text{softmax}_{j}\biggr(\frac{f_{\epsilon,\psi}(z_{k,j},k,\lambda) - c(z_{k,j},\bar{z}_{i}^{(\ell)})}{\epsilon}\biggr)z_{k,j}.
    \end{aligned}\label{eq:neural-fixed-pt}
\end{equation}
At inference, we draw samples $z_{k,i} \sim p_k$, initialize $\bar{z}_{i}^{(0)} = \sum_{k=1}^{K}\lambda_{k}z_{k,i}$ then execute the iterations in equation~\ref{eq:neural-fixed-pt}. This procedure is shown in Algorithm~\ref{alg:neuralfp-inference}, and can be interpreted as a neural version of the algorithms of~\cite{cuturi2014fast} and~\cite{montesuma2023multi}.

\begin{algorithm}[ht]
\caption{\texttt{NeuralFP} Inference}
\label{alg:neuralfp-inference}
\begin{algorithmic}[1]
\Require Measures $p_{1:K}$, dual potential $f_{\psi}$, number of samples $n$, \# of fixed-point iterations $L_{\text{FP}}$, weights $\lambda$, entropic regularization $\epsilon$
\State Sample $\{\{z_{k,j}\}_{j=1}^{n}\}_{k=1}^{K}$, where $z_{k,j} \sim p_{k}$
\State $f_{k,j} \leftarrow f_{\epsilon,\psi}(z_{k,j},k,\lambda)$
\State $\bar{z}_{i}^{(0)} \leftarrow \sum_{k=1}^{K}\lambda_{k}z_{k,i}$
\For{$\ell=1,\cdots,L_{\text{FP}}$}
        \State $C_{k,i,j} \leftarrow c(z_{k,j},\bar{z}_{i}^{(\ell-1)})$
        \State $\gamma_{j\lvert i, k} \leftarrow \softmax_{c}\left(\nicefrac{(f_{k,j} - C_{k,i,j})}{\epsilon}\right)$
        \State $\bar{z}^{(\ell)}_{i} \leftarrow \sum_{k=1}^{K} \lambda_{k}\sum_{j=1}^{B}\gamma_{j\lvert i,k}z_{k,j}$
\EndFor
\State \Return $\{\bar{z}_{i}^{(L_{\text{FP}})}\}_{i=1}^{n}$\Comment{Support of $\hat{q}_{\lambda}^{(L_{\text{FP}})}$}
\end{algorithmic}
\end{algorithm}

\paragraph{Related Work.} The \texttt{NeuralFP} version of our method sits within the existing literature of neural solvers for the fixed-point approach of~\cite{alvarez2016fixed}, such as~\cite{korotin2022wasserstein} and~\cite{howard2025schrdinger}. The shared goal is approximating the fixed-points of the operator $G$. We divide these methods into 2 families.

\emph{The first family} parametrize $q$ and $p_k$ empirically, meaning, they are mixtures of Diracs. As a consequence, the sample-level Monge map $T_{k}^{\star}$ may not exist (e.g., when the supports have different size) but they can be approximated via the barycentric mapping in equation~\ref{eq:emp-fixed-pt}. This is the case of works such as~\cite{cuturi2013sinkhorn,montesuma2023multi,Lindheim2023} and~\cite{montesuma2025computing}.

\emph{The second family} parametrize the iterate $q_{\lambda}^{(\ell)}$ through a neural network. This is the case of \texttt{WIN}~\citep{korotin2022wasserstein}. These authors propose a generative model for the barycenter iterate, that is, $q_{\lambda}^{(\ell)} = \xi_{\sharp}q_0$, where $q_0$ is a latent measure. Their method works by alternating 2 steps. First, they estimate $K-$transport maps $T_{\theta_k}$ from $\xi_{\sharp}q_0$ to $p_k$. Second, they regress $\xi$ into the approximated fixed-point operator $\sum_{k}\lambda_kT_{\theta_k}\circ \xi$.

In comparison with these approaches, our \texttt{NeuralFP} parametrizes the dual \gls{ot} parameters through a single network $f_{\psi}$. While this approach computes an entropic barycenter, it only needs $K-$neural networks and its parametrization is independent of the support of $q_{\lambda}^{\star}$. In contrast, the 1st family of methods parametrize the barycentric measure directly through their support, and the 2nd family, namely, \texttt{WIN}, uses $(2K+1)$ neural networks.

\subsection{Semi-Unbalanced \texttt{BaryFM}}\label{sec:su-baryfm}

Taking inspiration from~\cite{gazdieva2024robust}, we compute a semi-unbalanced version of our \texttt{BaryFM}. In a nutshell, unbalanced \gls{ot} is an optimization over a coupling with relaxed constraints. Recalling equations~\ref{eq:kantorovich} and~\ref{eq:sinkhorn}, the unbalanced \gls{ot} problem~\citep{chizat2018unbalanced,liero2018optimal} is,
\begin{align*}
    \text{UOT}(p,q) = \arginf{\gamma\in\mathcal{M}_{+}(\Omega\times\Omega)}\mathbb{E}_{(z,z')\sim\gamma}[c(z,z')] - \epsilon H(\gamma) + D_{\eta}(\pi_{1,\sharp}\gamma\lVert p) + D_{\xi}(\pi_{2,\sharp}\gamma\lVert q),
\end{align*}
where $D_\eta$ and $D_\xi$ are~\cite{csiszar1975divergence} divergences between measures over $\Omega$ (e.g., the Kullback-Leibler divergence), $\eta$ and $\xi$ are convex functions with $\eta(1) = \xi(1) = 0$, and $\pi_{i}(z_1,\cdots,z_n) = z_{i}$ is the projection onto the $i-$th coordinate. This problem takes the hard-constraint \gls{ot} problem, and regularizes it so $\gamma$ is simply a non-negative measure, while $D_{\eta}$ and $D_{\xi}$ act as a soft-constraint type of regularization. We are interested in the \emph{semi-unbalanced case}~\citep[Equation 4]{gazdieva2024robust}, in which only one marginal is penalized,
\begin{align}
    \text{SUOT}(p,q) = \arginf{\gamma\in\Gamma(q)}\mathbb{E}_{(z,z')\sim\gamma}[c(z,z')] - \epsilon H(\gamma) + \tau D_{\eta}(\pi_{1,\sharp}\gamma\lVert p),\label{eq:suot}
\end{align}
where $\Gamma(q) = \{\gamma \in \mathcal{M}_{+}(\Omega\times\Omega):\pi_{2,\sharp}\gamma = q\}$ is the set of plans with the 2nd marginal constraint, and $\tau > 0$ is the 1st marginal violation penalty weight. This formulation is adequate for barycenters, as in practice we want to allow the mass constraint over $p_{1:K}$ to be violated. The mass constraint on the barycenter side, $q_{\lambda}^{\star}$, must be respected, otherwise we could have degenerate solutions (e.g., barycenters with null mass). In light of equation~\ref{eq:suot}, we consider a modified semi-unbalanced dual loss~\citep[Equation 5]{gazdieva2024robust},
\begin{align}
    \sum_{k=1}^{K}\lambda_{k}\biggr(\mathbb{E}_{z}[-\tau\eta^{*}(-f_{\epsilon,\psi}(z,k,\lambda)/\tau) ] + \mathbb{E}_{\bar z}[f_{\epsilon,\psi}^{(c)}(\bar z,k,\lambda)] - \epsilon(\mathbb{E}_{z,\bar{z}}[\gamma_{\epsilon,\psi}(z,\bar{z},k,\lambda)]-1)\biggr).\label{eq:loss-dual-unbalanced}
\end{align}
where $\eta^{*}(u)$ denotes the Fenchel conjugate $\eta^{*}(u) = \text{sup}_{s \geq 0}\{su - \eta(s) \}$ evaluated at the dual $f_{\epsilon,\psi}(z,k,\lambda) \in \mathbb{R}$.

\subsection{Other ODE Integration Strategies}\label{sec:integration}

In the main paper (c.f. equation~\ref{eq:ode-inference}), we described the forward Euler strategy for solving the \gls{ode},
\begin{align*}
    \partial_t z(t) = v_{\theta}(z(t), t, k, \lambda),
\end{align*}
by approximating $\partial_t z(t) = \nicefrac{dz}{dt} \approx \nicefrac{(z_{r+1} - z_{r})}{dt}$, $r = \nicefrac{t}{dt}$. Other numerical approximation strategies exist~\citep[Chapter 5]{burden2010numerical}. In our experiments (c.f. Section~\ref{sec:swiss-roll}), we ablate the Euler strategy against the midpoint strategy,
\begin{align*}
    z_{\text{mid},i} &= z_{r,i} + v_{\theta}(z_{r,i}, rdt, k, \lambda)\dfrac{dt}{2},\\
    z_{r+1, i} &= z_{r, i}+v_{\theta}(z_{\text{mid},i}, rdt + \nicefrac{dt}{2}, k, \lambda)dt
\end{align*}
It is a general fact from numerical analysis that the error involved with discretizing the \gls{ode} depends on the strategy (e.g., Euler, or midpoint), and that midpoint the solver generally has a smaller error for the same time discretization parameter $dt$. See Figure~\ref{fig:comparison-methods-swiss-roll} for a comparison in our context.

\newpage

\section{Swiss Roll Measures}\label{sec:swiss-roll}

In this experiment, we build a toy example where each member of the family $\mathcal{W}(p_{1:K})$ has a closed-form ground-truth. To that end, we use the theory developed in~\cite[Section 4]{alvarez2016fixed}, where the authors extend the fixed-point barycenter computation algorithm to location-scatter families. Briefly, let $p_0$ be a reference measure, such as a Gaussian measure, or, in the context of this experiment, the Swiss roll measure. We assume, without loss of generality, that $p_0$ is centered with identity covariance. Given $T_{1},\cdots,T_{K}$ affine transformations, we consider the family,
\begin{align}
    \text{Location-Scatter}(p_0) = \{ T_{k,\sharp}p_0: T_{k}(x) = A_{k}x + b_k, A_k \in \text{SPD}(d)\text{, and }b_k \in \mathbb{R}^{d} \}_{k=1}^{K}.\label{eq:loc-scatter}
\end{align}

\begin{wrapfigure}[12]{r}{0.6\textwidth}
    \centering
  \begin{subfigure}{0.32\linewidth}\centering
    \includegraphics[width=\linewidth]{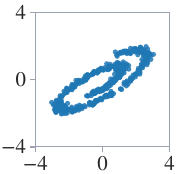}\caption{$p_1$}\end{subfigure}\hfill
  \begin{subfigure}{0.32\linewidth}\centering
    \includegraphics[width=\linewidth]{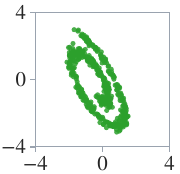}\caption{$p_2$}\end{subfigure}\hfill
  \begin{subfigure}{0.32\linewidth}\centering
    \includegraphics[width=\linewidth]{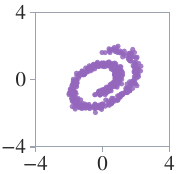}\caption{$p_3$}\end{subfigure}
  \caption{The $K=3$ Swiss-roll marginals.}
  \label{fig:marginals}
\end{wrapfigure}
Each measure $p_k \in \text{Location-Scatter}(p_0)$ is the image measure of $p_0$ by the affine transformation $T_k$.  We show an example of the marginal measures in Figure~\ref{fig:marginals}, which are used throughout the examples in this paper. We use $K=3$ for visualizing these measures in the simplex

This example is similar to that proposed by~\cite{korotin2021continuous}, and is interesting for several reasons. First, the Swiss roll measure itself has a non-linear structure that makes the question of barycenter accuracy interesting. Second, its barycenters have analytical solution in terms of the parameters $\{(A_k, b_k)\}_{k=1}^{K}$, so methods can be scored against a known ground-truth. Third, and more importantly to us, the family $\mathcal{W}(p_{1:K})$ is also known analytically. For reference, following~\cite[Section 4]{alvarez2016fixed}, barycenters of $p_{1:K}$ are $q_{\lambda}^{\star} = T_{\lambda,\sharp}p_0$, where $T_{\lambda}(x) = A_{\lambda}x + b_{\lambda}$ is affine with coefficients that depend on $\lambda \in \Delta_{K}$.

\begin{wrapfigure}[12]{r}{0.45\textwidth}
  \centering
  \includegraphics[width=\linewidth]{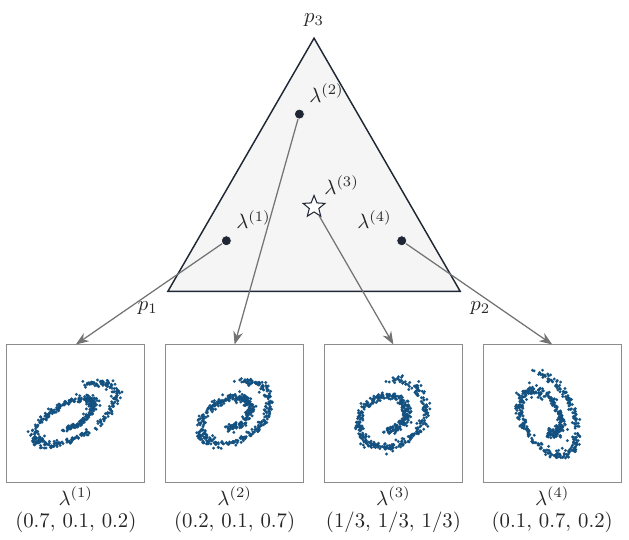}
  \caption{Wasserstein simplex $\mathcal{W}_{p_{1:3}}$ with 4 measures highlighted. Due the location-scatter structure, all measures in $\mathcal{W}(p_{1:3})$ are known in closed-form.}
  \label{fig:simplex-illustration}
\end{wrapfigure}
In this experiment, we use $K=3$ measures with the following coefficients,
\begin{align*}
    A_1 &= \begin{bmatrix}
        1.2292 & 0.7474\\
        0.7474 & 1.4325
    \end{bmatrix}\quad b_1 = \begin{bmatrix}
        -0.3807 & -0.2799
    \end{bmatrix},\\
    A_2 &= \begin{bmatrix}
        1.7729 & 0.6365\\
        0.6365 & 0.9066
    \end{bmatrix}\quad b_2 = \begin{bmatrix}
        -0.5273 & -0.2228
    \end{bmatrix},\\
    A_3 &= \begin{bmatrix}
        0.5725 & 0.1488\\
        0.1488 & 2.1387
    \end{bmatrix}\quad b_3 = \begin{bmatrix}
        -0.1675 & -0.1474
    \end{bmatrix},\\
\end{align*}
As mentioned earlier, the main interest of this family is that barycenters have analytical solution in terms of $T_{\lambda}$ and $p_0$,
\begin{equation}
    \begin{aligned}
        b_{\lambda} = \sum_{k=1}^{K}\lambda_{k}b_{k},\,\,
        A_{\lambda}^{2} = \sum_{k=1}^{K}\lambda_{k}\bigl(A_{\lambda}A_{k}^{2}A_{\lambda}\bigr)^{1/2}.
    \end{aligned}
\end{equation}
We compare 3 methods over this family of measures: \texttt{WGF}~\citep{montesuma2025computing}, which computes, from scratch, $q_{\lambda}^{\star}$, \texttt{NWBU}~\citep[Appendix B]{fan2020scalable}, which uses partial \glspl{icnn}, accepting $\lambda$ as input, and \texttt{BaryFM} (ours). We rank methods according the BW2-UVP\% metric of~\cite{korotin2021continuous}, which itself depends on the Wasserstein distance over Gaussian measures, known as the Bures-Wasserstein metric~\citep{takatsu2011wasserstein},
\begin{align}
    \text{BW}_2(p,q)^2 = \lVert \mu_p - \mu_q \rVert_{2}^{2} + \text{Tr}\Bigl( \Sigma_p + \Sigma_q - 2\bigl( \Sigma_p^{1/2}\Sigma_q\Sigma_p^{1/2} \bigr)^{1/2} \Bigr).
\end{align}
From the BW metric,
\begin{align}
    \text{BW2-UVP\%}(p\lvert q) = 100\dfrac{\text{BW2}(p,q)^2}{\frac{1}{2}\text{Var}(q)}\%,\label{eq:bw-uvp}
\end{align}
where $\text{Var}(q) = \int_{\Omega} \lVert z-\mu_q \rVert^2 dq$ is the variance of the probability measure $q$.

\begin{figure}[ht]
    \centering
    \includegraphics[width=0.5\linewidth]{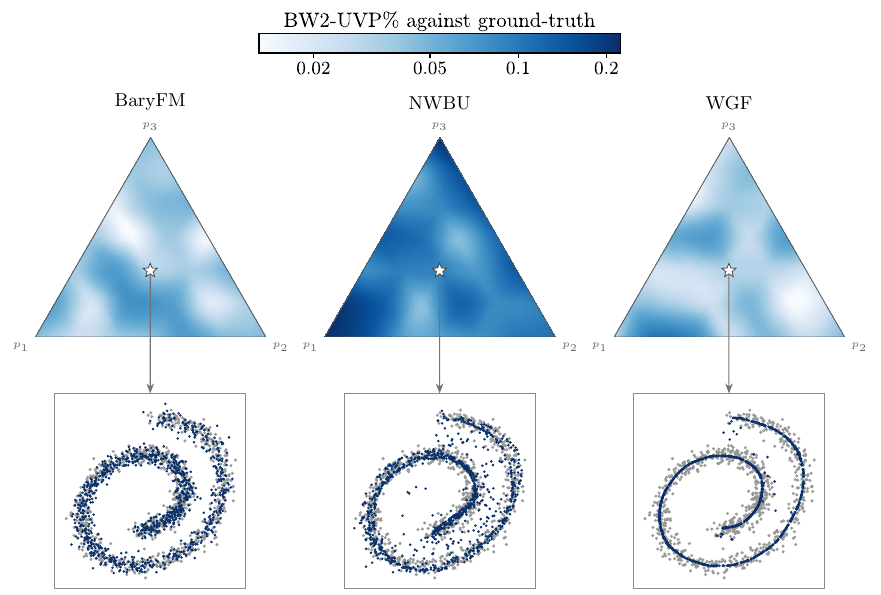}
    \caption{BW2-UVP\% over the simplex for \texttt{WGF}, \texttt{NWBU}, and \texttt{BaryFM} (ours) in the Swiss Roll toy example. Blue points represent the calculated barycenter by each method, gray points represent points from the ground-truth.}
    \label{fig:universal-swiss-roll}
\end{figure}

\paragraph{Main Results.} We summarize our main results in Figure~\ref{fig:universal-swiss-roll}, which shows the universal approximation of the Swiss roll measures across the simplex. We highlight the barycenter under uniform weights, at the center of the simplex. Overall, \texttt{WGF} and \texttt{BaryFM} barycenters are more accurate than \texttt{NWBU}, by the intensity of colors of the heatmaps across the simplex. Qualitatively, the 2nd row of Figure~\ref{fig:universal-swiss-roll} shows that \texttt{NWBU} barycenters are correct, but considerably more noisy. Curiously, \texttt{WGF}'s barycenters have a very tight support, which is a consequence of entropic regularization. In contrast, our \texttt{BaryFM} fits the underlying Swiss roll manifold very well.

\begin{figure}[ht]
    \centering
    \includegraphics[width=\linewidth]{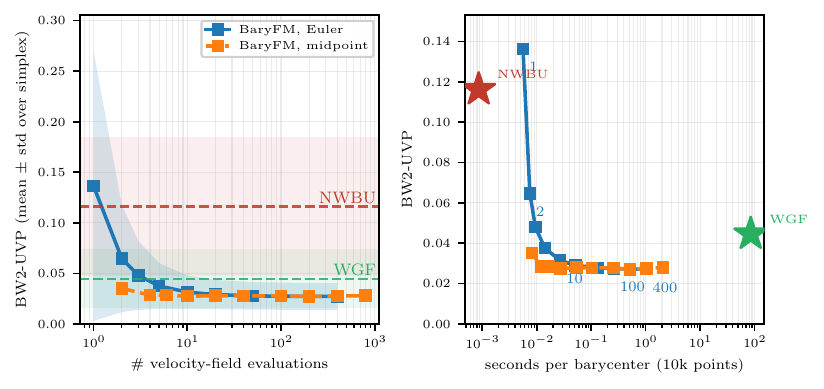}
    \caption{BW2-UVP\% as a function of \# of velocity field evaluations and inference time.}
    \label{fig:comparison-methods-swiss-roll}
\end{figure}

\paragraph{Sampling Speed.} An immediate advantage of amortizing the sampling of the family $\mathcal{W}(p_{1:K})$ is computational efficiency. To that end we explore, in Figure~\ref{fig:comparison-methods-swiss-roll}, the average accuracy across the simplex as a function of \# of velocity-field evaluations, and seconds per barycenter. Our method sits between \texttt{NWBU} and \texttt{WGF}, while being around 4 order of magnitude faster than the latter, while at parity with respect BW2-UVP\%.

\newpage

\section{Bayesian Posterior Aggregation}\label{sec:additional-bpa}

\paragraph{Background.} Let $\mathcal{D}$ denote a dataset (e.g., a collection of samples $\{(x_i, y_i)\}_{i=1}^{n}$). Bayesian inference is concerned with estimating the \emph{posterior measure}, $p(\theta|\mathcal{D})$, where $\theta \in \Theta$ are model parameters (e.g., weights of neural network, or coefficients of a linear regression model) given the dataset $\mathcal{D}$. The naming \emph{posterior}, comes from Bayes rule,
\begin{align*}
    p(\theta|\mathcal{D}) = \dfrac{p(\theta)p(\mathcal{D}|\theta)}{p(\mathcal{D})},
\end{align*}
where $p(\theta)$ is called the prior, $p(\mathcal{D}|\theta)$ is the likelihood, and $p(\mathcal{D})$ is the probability of observing the dataset $\mathcal{D}$. Therefore, given a set of datasets $\{\mathcal{D}_{k}\}_{k=1}^{K}$, Bayesian posterior aggregation seeks to combine the posterior measures $\{p(\theta|\mathcal{D}_{k})\}_{k=1}^{K}$. Samples from these posteriors, $\{ \{ \theta_{k,j} \}_{j=1}^{n_k} \}_{k=1}^{K}$ can be obtained, for instance, via \gls{mcmc}.

In this context,~\cite{srivastava2015wasp,srivastava2018scalable} were the first to aggregate posterior via Wasserstein barycenters, that is, via,
\begin{align*}
    q^{\star} = \text{Bar}(\bar{\lambda},p_{1:K}) = \argmin{q \in \mathcal{P}_{2}(\Omega)}\dfrac{1}{K}\sum_{k=1}^{K}W_{2}(q,p(\theta|\mathcal{D}_{k}))^{2}.
\end{align*}
In other words, given $p_{k} = p(\theta|\mathcal{D}_{k})$, they propose $q^{\star}_{\bar{\lambda}} = \text{Bar}(\bar{\lambda}, p_{1:K})$, with $\bar{\lambda} = (\nicefrac{1}{K},\cdots,\nicefrac{1}{K})$. This task has a well defined ground-truth, as the true posterior corresponds to sampling from $q_{0} = p(\theta \lvert \mathcal{D})$, for instance, running \gls{mcmc} on the full dataset. Therefore, methods are ranked with respect their BW2-UVP\% against the ground-truth, $\text{BW2-UVP\%}(\hat{q}_{\bar{\lambda}}, q_{0})$. Note, however, that the BW2-UVP compares measures based on their 1st and 2nd moments only. We therefore include in our comparison the $2-$Sliced-Wasserstein distance,
\begin{align}
    \text{SW}_{2}(p,q)^{2} = \int_{\mathbb{S}^{d-1}} W_{2}(\pi_{u,\sharp}p, \pi_{u,\sharp}q)^{2} du \approx \dfrac{1}{N_{\text{proj}}} \sum_{i=1}^{N_{\text{proj}}}W_{2}( \pi_{u_{i},\sharp}p, \pi_{u_{i},\sharp}q )^{2},\label{eq:sw2}
\end{align}
where $\mathbb{S}^{d-1}$ is the unit sphere in $\mathbb{R}^{d}$, and $u_{i} \sim \mathcal{U}(\mathbb{S}^{d-1})$ is an unit norm vector. We use $N_{\text{proj}} = 500$ projections. The Sliced Wasserstein distance relies on the Wasserstein distance between 1-dimensional measures, and enjoys a dimension independent statistical complexity. We refer readers to~\cite{montesuma2024recent} and~\cite{peyre2019computational} for further discussion.

\paragraph{Dataset.} For this experiment we consider the dataset introduced in~\cite{noble2023tree}. These authors have obtained samples from a logistic regression model on the UCI\footnote{\url{https://archive.ics.uci.edu/dataset/109/wine}} \texttt{Wine} dataset of~\cite{aeberhard1994comparative}. More specifically, the authors have used $3$ partitions of the original \texttt{Wine} dataset for obtaining 9,900 samples from each $p_k = p(\theta\lvert \mathcal{D}_k)$, $k=1,\cdots,3$. Likewise, the 9,900 samples are obtained from the true posterior $q_0 = p(\theta \lvert \mathcal{D})$. We refer readers to~\cite[Appendix G.2]{noble2023tree} for further details on the specifics of their numerical simulations. All methods are evaluated with respect the BW2-UVP\% and SW$_{2}$ metrics, using all the 29,700 transported marginal samples to the computed barycenter.

\paragraph{Main Results.} Our main results are shown in Table~\ref{tab:comparison_bayesian}. We compare our \texttt{BaryFM} against several barycenter solvers. Every method is run on 3 seeds (42, 43 and 44), and we report their \result{$\mu$}{\sigma}. Overall, our semi-unbalanced variant (\texttt{BaryFM-U}) performs better than the its balanced version (\texttt{BaryFM}). For instance, in the homogeneous benchmark, \texttt{BaryFM-U} ranks 5th, while \texttt{BaryFM} ranks 7th. We are able to improve performance with \gls{cfg} (see analysis below), matching the best previous method on the homogeneous benchmark, and surpassing it on the hterogeneous one.

In the homogeneous benchmark, with $k-$\gls{cfg}, our \texttt{BaryFM-U} achieves \result{8.02}{0.07} against \texttt{U-NOT} \result{8.08}{0.10} (2nd best, also unbalanced method), thus being statistically tied. In the heterogeneous benchmark, we our \texttt{BaryFM-U} achieves \result{9.93}{0.17}, against \result{10.26}{0.08} of \texttt{U-NOT} (3rd best, also unbalanced method). Overall, the main drivers of performance in this benchmark are (i) universal training, (ii) \gls{cfg}, and (iii) semi-unbalanced \gls{ot}. Universal training is our particular proposal, \gls{cfg} is made possible by \gls{cfm}, and semi-unbalanced \gls{ot} is made possible via the dual \gls{ot} head.

\begin{table}[ht]
    \centering
\resizebox{\linewidth}{!}{
    \begin{tabular}{lccccccccc}
    \toprule
    & & \multicolumn{4}{c}{Homogeneous} & \multicolumn{4}{c}{Heterogeneous}\\
    \cmidrule(lr){3-6}\cmidrule(lr){7-10}
    Method & Universal & BW2-UVP\% & Rank & SW$_{2}$ & Rank & BW2-UVP\% & Rank & SW$_{2}$ & Rank\\
    \midrule
    Discrete & \xmark & \result{17.70}{0.02} & 16 & \result{0.310}{0.004} & 17 & \result{18.92}{0.08} & 17 & \result{0.317}{0.007} & 17 \\
    WGF & \xmark & \result{20.42}{0.02} & 18 & \result{0.333}{0.004} & 18 & \result{20.93}{0.00} & 18 & \result{0.338}{0.004} & 18 \\
    \midrule
    TDSB & \xmark & \result{14.22}{1.73} & 14 & \result{0.234}{0.019} & 15 & \result{17.03}{1.46} & 15 & \result{0.256}{0.011} & 14 \\
    TSBM & \xmark & \result{8.09}{0.03} & 3 & \result{0.179}{0.007} & 8 & \result{10.34}{0.07} & 5 & \result{0.202}{0.007} & 9 \\
    CW2B & \xmark & \result{11.58}{0.69} & 12 & \result{0.222}{0.012} & 14 & \result{16.27}{1.54} & 14 & \result{0.264}{0.011} & 15 \\
    WIN & \xmark & \result{12.79}{0.72} & 13 & \result{0.207}{0.010} & 12 & \result{15.23}{1.67} & 12 & \result{0.227}{0.022} & 13 \\
    NOT & \xmark & \result{8.54}{0.18} & 9 & \result{0.183}{0.008} & 10 & \result{11.10}{0.20} & 11 & \result{0.209}{0.009} & 11 \\
    U-NOT & \xmark & \result{8.08}{0.10} & 2 & \result{0.176}{0.007} & 5 & \result{10.26}{0.08} & 3 & \result{0.197}{0.008} & 6 \\
    NF & \xmark & \result{8.41}{0.02} & 7 & \result{0.183}{0.007} & 10 & \result{10.79}{0.03} & 9 & \result{0.206}{0.007} & 10 \\
    NWB & \xmark & \result{15.65}{4.10} & 15 & \result{0.217}{0.045} & 13 & \result{15.34}{1.90} & 13 & \result{0.219}{0.032} & 12 \\
    GMM & \xmark & \result{8.28}{0.15} & 6 & \result{0.178}{0.008} & 7 & \result{10.69}{0.15} & 8 & \result{0.201}{0.009} & 8 \\
    \texttt{BaryFM} (ours) & \xmark & \result{9.31}{0.17} & 11 & \result{0.182}{0.007} & 9 & \result{10.80}{0.08} & 10 & \result{0.199}{0.008} & 7 \\
    \texttt{BaryFM-U} (ours) & \xmark & \result{8.97}{0.10} & 10 & \result{0.175}{0.007} & 4 & \result{10.25}{0.17} & 2 & \result{0.191}{0.007} & 2 \\
    \midrule
    NWBU & \cmark & \result{20.30}{2.36} & 17 & \result{0.265}{0.028} & 16 & \result{18.40}{1.28} & 16 & \result{0.268}{0.019} & 16 \\
    \texttt{BaryFM} (ours) & \cmark & \result{8.41}{0.06} & 7 & \result{0.176}{0.006} & 5 & \result{10.51}{0.23} & 7 & \result{0.195}{0.009} & 5 \\
    \texttt{BaryFM-U} (ours) & \cmark & \result{8.27}{0.06} & 5 & \result{0.169}{0.005} & 1 & \result{10.35}{0.25} & 6 & \result{0.192}{0.006} & 3 \\
    \midrule
    \texttt{BaryFM} (ours) + $k$-CFG & \cmark & \result{8.19}{0.07} & 4 & \result{0.174}{0.009} & 3 & \result{10.31}{0.46} & 4 & \result{0.194}{0.012} & 4 \\
    \texttt{BaryFM-U} (ours) + $k$-CFG & \cmark & \result{8.02}{0.07} & 1 & \result{0.170}{0.008} & 2 & \result{9.93}{0.17} & 1 & \result{0.189}{0.009} & 1 \\
    \bottomrule
    \end{tabular}
}
    \caption{Bayesian Posterior Aggregation results. We report BW2-UVP\% and SW$_{2}$ over seeds 42, 43 and 44 for each method.}
    \label{tab:comparison_bayesian}
\end{table}

\paragraph{On the benefits of universality.} One result that stands out from Table~\ref{tab:comparison_bayesian} is the effect of universality. Comparing \texttt{BaryFM} universal versus non-universal, its ranks improve from 11th in the Homogeneous setting to 7th, and from 10th to 7th in the Heterogeneous setting. In the semi-unbalanced version, the gains are more mixed. In the Homogeneous benchmark rank improves from 10th to 5th, and in the Heterogeneous benchmark rank degrades from 2nd to 6th. Except for this degradation, universality tends to improve the downstream score in posterior aggregation (i.e., similarity with the posterior).

\begin{figure}[ht]
    \centering
    \begin{subfigure}{0.24\linewidth}
        \includegraphics[width=\linewidth]{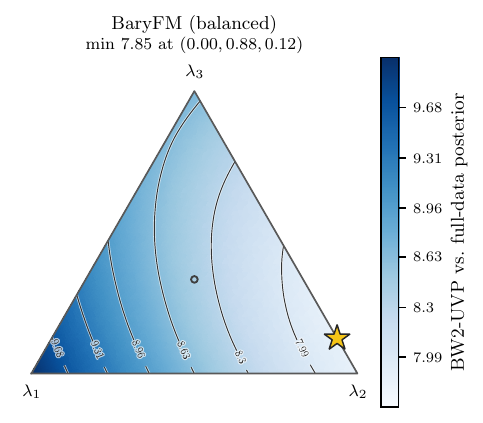}
        \caption{Homogeneous.}
    \end{subfigure}
    \begin{subfigure}{0.24\linewidth}
        \includegraphics[width=\linewidth]{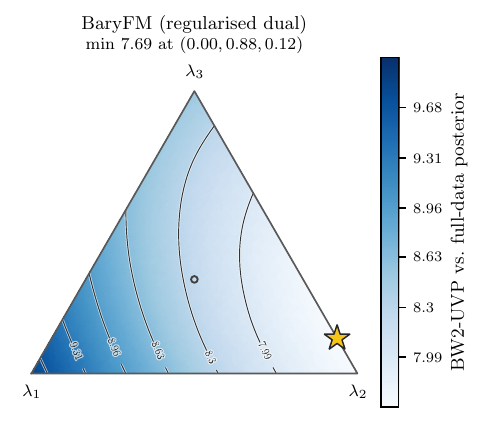}
        \caption{Homogeneous.}
    \end{subfigure}
    \begin{subfigure}{0.24\linewidth}
        \includegraphics[width=\linewidth]{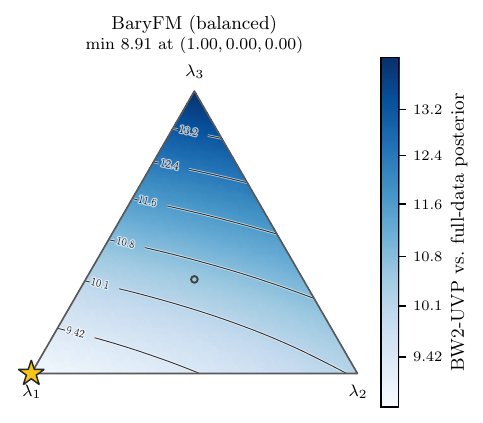}
        \caption{Heterogeneous.}
    \end{subfigure}
    \begin{subfigure}{0.24\linewidth}
        \includegraphics[width=\linewidth]{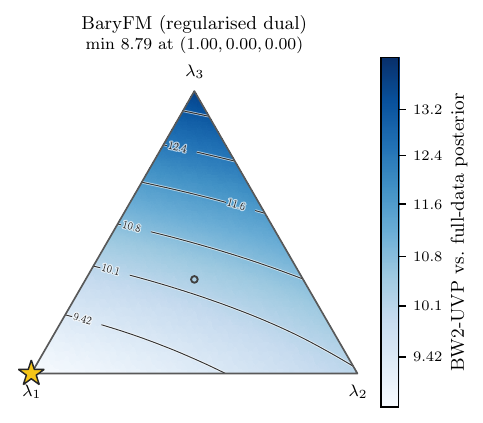}
        \caption{Heterogeneous.}
    \end{subfigure}
    \caption{BW2-UVP\% between $\hat{q}_{\lambda}^{(L_{\text{FP}})}$, the $\lambda$-weighted Wasserstein barycenter of marginals, against the ground truth full posterior $q_0$ over the simplex $\Delta_{3}$.}
    \label{fig:simplex-bpa}
\end{figure}

Besides the ranking improvement, there is a second advantage of approximating the Wasserstein simplex $\mathcal{W}(p_{1:K})$, as we can actually explore $\lambda \mapsto \text{BW2-UVP}\%(\hat{q}_{\lambda}^{(L_{\text{FP}})}, q_0)$. We show this exploration in Figure~\ref{fig:simplex-bpa}, which shows that the minimum BW2-UVP\% is not actually in the uniform barycenter $\bar{\lambda}$. Indeed, our algorithm is able to obtain much better metrics in the homogeneous (best BW2-UVP\% 7.69) and heterogeneous settings (best BW2-UVP\% 8.79).

\paragraph{Classifier Free Guidance.} A nice possibility made available by flow matching is \gls{cfg}~\citep{ho2022classifier}, based on dropping out the conditioning factors of $v_{\theta}(z,t,k,\lambda)$. We experiment with the dropout of 2 conditioning variables, the domain index $k$, and the barycentric coordinates $\lambda \in \Delta$,
\begin{align}
    v(z,t,k,\lambda,s) &= v_{\theta}(z,t,\emptyset,\lambda) + s (v_{\theta}(z,t,k,\lambda) - v_{\theta}(z,t,\emptyset,\lambda)),\label{eq:k-dropout}\\
    v(z,t,k,\lambda,s) &= v_{\theta}(z,t,k,\emptyset) + s (v_{\theta}(z,t,k,\lambda) - v_{\theta}(z,t,k,\emptyset)),\label{eq:vanilla-lambda-cfg}
\end{align}
where $s$ is the so-called guidance scale. As commonly done in the flow matching and diffusion literature~\citep{ho2022classifier}, we define the embedding vectors $E_{k}(\emptyset)$ and $E_{\lambda}(\emptyset)$, and train our \texttt{BaryFM} with a dropout probability on the conditionings $k$ and $\lambda$.

\begin{figure}[ht]
    \centering
    \begin{subfigure}{0.48\linewidth}
        \includegraphics[width=\linewidth]{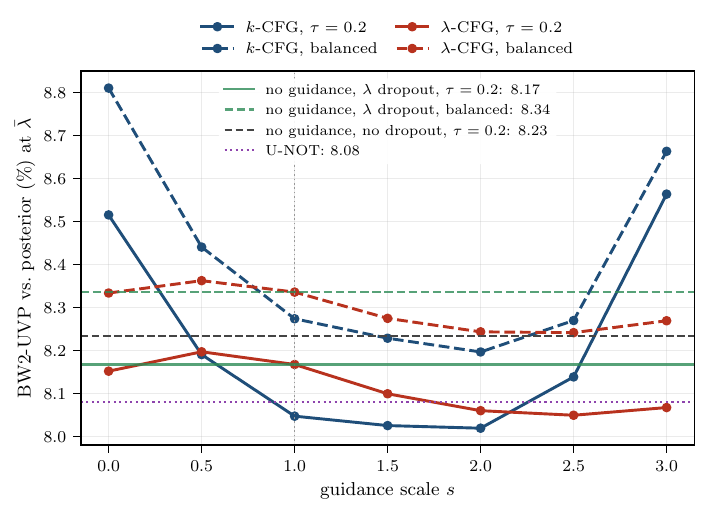}
        \caption{Homogeneous}
    \end{subfigure}
    \begin{subfigure}{0.48\linewidth}
        \includegraphics[width=\linewidth]{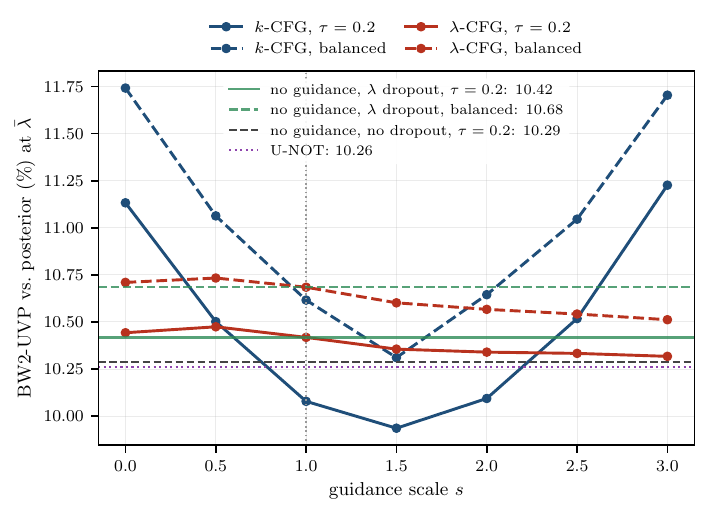}
        \caption{Heterogeneous}
    \end{subfigure}
    \caption{Different \gls{cfg} strategies ($k$-\gls{cfg} and $\lambda$-\gls{cfg}) for Bayesian Posterior Aggregation. Through \gls{cfg} ($s \neq 1$, especially $s \in (1, 2]$ our proposed \texttt{BaryFM} improves over the best previous baseline, \texttt{U-NOT}~\citep{gazdieva2024robust}.}
    \label{fig:cfg-bpa}
\end{figure}

We call equation~\ref{eq:k-dropout} $k-$conditioning, and equation~\ref{eq:vanilla-lambda-cfg} $\lambda-$conditioning. From Figure~\ref{fig:cfg-bpa}, \gls{cfg} on the $k-$conditioning has a clear $U-$shape curve. Performance improves with a reasonable $s>1.0$, thus with \gls{cfg} active, achieving its best at $s = 2.0$ (homogeneous, both variants) and $s=1.5$ (heterogeneous, both variants). Performance improvements from $\lambda-$dropout are somewhat more mixed. On the one hand, it is able to improve to similar levels of $k-$\gls{cfg} in the Homogeneous benchmark. In the heterogeneous, gains are almost flat as a function of $s$. Interestingly, this \gls{cfg} strategy is monotonic in $s$.

\begin{table}[ht]
    \centering
    \begin{tabular}{lcccc}
    \toprule
    & \multicolumn{2}{c}{\texttt{BaryFM}} & \multicolumn{2}{c}{\texttt{BaryFM-U}}\\
    \cmidrule(lr){2-3}\cmidrule(lr){4-5}
    & -- & $+k$-CFG & -- & $+k$-CFG\\
    \midrule
    Semi-unbalanced penalty $\tau$              & --     & --     & 0.2    & 0.2\\
    Batch size $B$                              & 1\,024 & 4\,096 & 1\,024 & 4\,096\\
    $k-$dropout (Homogeneous/Heterogeneous)       & 0.0      & 0.8 / 0.1 & 0.0   & 0.7 / 0.2\\
    Guidance scale $s$ (Homogeneous/Heterogeneous)           & 1.0      & 2.0 / 1.5 & 1.0   & 2.0 / 1.5\\
    \midrule
    Entropic regularisation $\epsilon$          & \multicolumn{4}{c}{2}\\
    Fixed-point iterations $L_{\text{FP}}$      & \multicolumn{4}{c}{1}\\
    Dual iterations $L_{\text{Dual}}$           & \multicolumn{4}{c}{1}\\
    Divergence $\eta$                       & \multicolumn{4}{c}{KL}\\
    \midrule
    Iterations $N$                              & \multicolumn{4}{c}{3\,000}\\
    Optimizer (field and dual)                  & \multicolumn{4}{c}{Adam, $\text{LR}=10^{-5}$}\\
    Learning-rate schedule                      & \multicolumn{4}{c}{cosine, $\text{LR}_{\text{min}}=10^{-7}$}\\
    EMA decay                                   & \multicolumn{4}{c}{0.999}\\
    Gradient clipping                           & \multicolumn{4}{c}{1.0}\\
    \midrule
    Network                                     & \multicolumn{4}{c}{densely connected, 2 layers, width 512}\\
    Embeddings (feature / time / domain / weight) & \multicolumn{4}{c}{128 / 32 / 32 / 32}\\
    \midrule
    $n_{\text{gen}}$                            & \multicolumn{4}{c}{29\,700 (full support, $3 \times 9\,900$)}\\
    ODE solver                                  & \multicolumn{4}{c}{Midpoint, 100 steps}\\
    \bottomrule
    \end{tabular}
    \caption{Hyperparameters of the universal \texttt{BaryFM} rows in Table~\ref{tab:comparison_bayesian}.}
    \label{tab:bpa-hparams}
\end{table}

\begin{figure}[ht]
    \centering
    \begin{subfigure}{0.32\linewidth}
        \includegraphics[width=\linewidth]{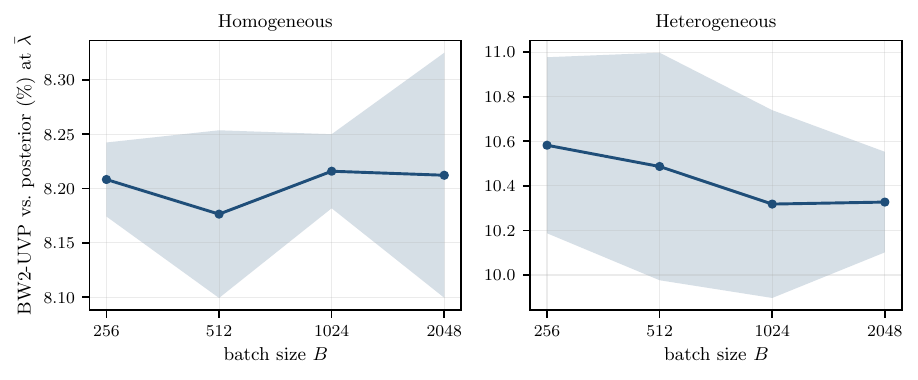}
        \caption{Sensitivity to $B$.}
    \end{subfigure}
    \begin{subfigure}{0.32\linewidth}
        \includegraphics[width=\linewidth]{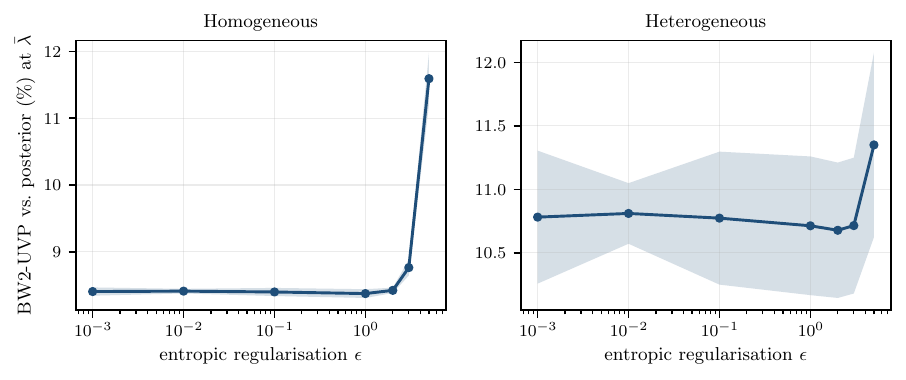}
        \caption{Sensitivity to $\epsilon$.}
    \end{subfigure}
    \begin{subfigure}{0.32\linewidth}
        \includegraphics[width=\linewidth]{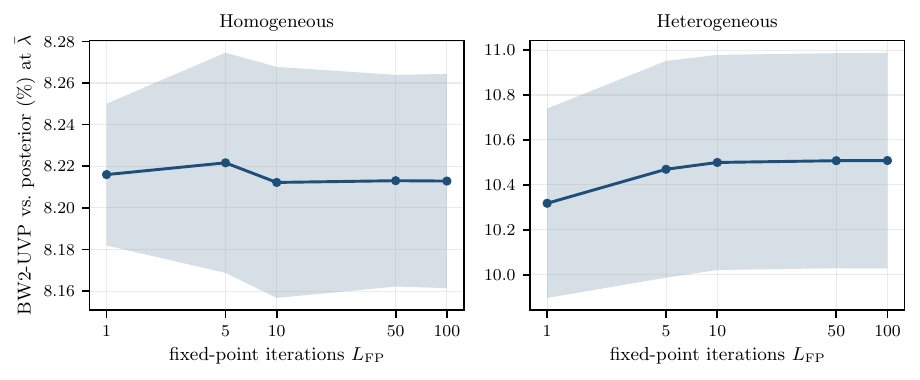}
        \caption{Sensitivity to $L_{\text{FP}}$.}
    \end{subfigure}
    \caption{Sensitivity to batch size $B$, entropic regularization $\epsilon$ and \# of fixed point iterations $L_{\text{FP}}$.}
    \label{fig:hp-sensitivity-bpa}
\end{figure}

\paragraph{Hyperparameters and their sensitivity.} We present, in Table~\ref{tab:bpa-hparams}, the hyperparameters used in our method, and its sensitivity to hyperparameter choice. We perform this study on the following axis: entropic regularization $\epsilon \in \{10^{-3}, 10^{-2}, 10^{-1}, 10^{0}, 2\times 10^{0}, 3\times 10^{0}, 5\times 10^{0}\}$, $B \in \{256, 512, 1024, 2048\}$, $L_{\text{FP}} \in \{ 1, 5, 10, 50, 100 \}$. Our results are summarized in Figure~\ref{fig:hp-sensitivity-bpa}. Overall, our algorithm is robust to both $B$ and $L_{\text{FP}}$, having a smaller variation than seed standard deviation. Sensitivity to $\epsilon$ is more pronounced, and our algorithm tends to prefer smaller $\epsilon$ (i.e., $\epsilon \leq 2$).

\paragraph{Conclusion.} Overall, our results demonstrate 2 interesting effects. First, the best posterior approximation is not attained at the center of simplex, meaning $\hat{q}_{\bar{\lambda}}$. Indeed, exploring the whole Wasserstein simplex $\mathcal{W}(p_{1:K})$ allows us to improve the BW2-UVP\% against the true posterior from 10.51 to 8.91 (heterogeneous, balanced \texttt{BaryFM}), with similar gains of our method in other settings. As a result, we argue that, at least in the context of this benchmark, the quality of barycenter approximation is only one axis of exploration, as there are other measures within the family of barycenters (i.e., the Wasserstein simplex) that do this task better. Second, \gls{cfg} is an important mechanism for posterior approximation, although is bias the calculated barycenter.

\clearpage
\newpage

\section{Fairness}\label{sec:additional-fairness}

\paragraph{Background.} In the context of machine learning, fairness~\citep{mehrabi2021survey} studies the biases of developed models or algorithms against sub-populations in the data. Mathematically, this task is formalized by a protected variable, $S$, taking values among a discrete set $\{1,\cdots,K\}$. For instance, a protected variable can concern gender, in which case $S\in\{1,2\}$, where $1$ corresponds to \emph{male}, and $2$ corresponds to \emph{female}. Let $h:\mathcal{X}\to \mathcal{Y}$ denote a predictor, mapping a feature space $\mathcal{X}$ to a label space $\mathcal{Y}$. Among the different notions of fairness, we consider \gls{di},
\begin{align}
    \text{DI}(h,X,S) = \frac{1}{K-1}\sum_{k=2}^{K} \frac{\text{Pr}(h(X)=1|S=k)}{\text{Pr}(h(X)=1|S=1)},\label{eq:disparate-impact}
\end{align}
where $\text{Pr}$ denotes the probability of the event in its argument. $h(X) = 1$ denotes the \emph{positive outcome} (e.g., obtaining a loan or a scholarship), and $S=1$ is the privileged group (e.g., \emph{male}). In a nutshell, \gls{di} measures the ratio, in probability, of getting a positive outcome from the rule $h$. Intuitively, a rule $h$ is \emph{fairer} as $\text{DI}\to 1$.

Wasserstein barycenters were first proposed to achieve fairness in machine learning by~\cite{gordaliza2019obtaining}. These authors proposed to \emph{repair} the measures of each sub-population into the Wasserstein barycenter weighted by the population proportions. In these experiments, we repair into the uniform barycenter,
\begin{align*}
    q_{\bar{\lambda}}^{\star} = \text{Bar}(\bar{\lambda}, p_{1:K})\text{, where }\bar{\lambda}=(\nicefrac{1}{K},\cdots,\nicefrac{1}{K}).
\end{align*}
The underlying idea is that the barycenter will correct the potential shifts within the protected variable-conditioned measures $p_{k}$. As a result, trained on the repaired measure (called total repair by~\cite{gordaliza2019obtaining}), the model will become insensitive to the protected variable. Uniform weights are used to ensure all groups are treated equally.

\paragraph{Experimental Protocol.} All barycenter algorithms are compared along 2 axis. First, we measure how fair models trained under the estimated barycenter $\hat{q}_{\bar{\lambda}}$. We measure fairness via equation~\ref{eq:disparate-impact}. Second, we measure how performance is degraded by repairing the probability measures. This is a consequence of repairing, and downstream classification accuracy is usually at a trade-off with respect fairness~\citep{zhao2022inherent}. In  that regard a we fit a logistic regression model to the repaired data obtained via each barycenter solver.

\paragraph{Datasets.} We consider $4$ datasets fore evaluating fairness: Adult Income~\citep{kohavi1996scaling}, COMPAS~\citep{angwin2022machine}, Credit~\citep{yeh2009comparisons} and the National Longitudinal Survey of Youth (NLSY)~\citep{bls_nlsy}. We summarize the information about each dataset in Table~\ref{tab:fairness-datasets}, among which the positive outcome $h(X)=1$, the considered protected variables, groups and the favored group.

\begin{table}[ht]
\centering
\resizebox{\linewidth}{!}{%
\begin{tabular}{rcccclccc}
\toprule
Dataset & Task ($Y{=}1$) & \# Samples & \# Features & Protected Variable & Groups & \# Samples per group & \% & $\Pr(Y{=}1\mid S{=}s)$ \\
\midrule
\multirow{2}{*}{Adult} & \multirow{2}{*}{income $>50$K} & \multirow{2}{*}{45,222} & \multirow{2}{*}{5} & \multirow{2}{*}{gender} & \textbf{Male} & 30,527 & 67.5 & 0.312 \\
 &  &  &   &  & Female & 14,695 & 32.5 & 0.114 \\
\midrule
\multirow{2}{*}{COMPAS} & \multirow{2}{*}{two-year \textbf{non-recidivism}} & \multirow{2}{*}{5,278} & \multirow{2}{*}{8} & \multirow{2}{*}{race} & Afr.-Am. & 3,175 & 60.2 & 0.477 \\
 &  &  &   &  & \textbf{Caucasian} & 2,103 & 39.8 & 0.609 \\
\midrule
\multirow{2}{*}{Credit} & \multirow{2}{*}{\textbf{no} payment default} & \multirow{2}{*}{30,000} & \multirow{2}{*}{22}  & \multirow{2}{*}{gender} & \textbf{Female} & 18,112 & 60.4 & 0.792 \\
 &  &  &   &  & Male & 11,888 & 39.6 & 0.758 \\
\midrule
\multirow{4}{*}{NLSY} & \multirow{4}{*}{income $\geq$ median} & \multirow{4}{*}{209,863} & \multirow{4}{*}{5}  & \multirow{4}{*}{gender $\times$ race} & Female-Non-Bl. & 78,554 & 37.4 & 0.452 \\
 &  &  &   &  & \textbf{Male-Non-Bl.} & 76,561 & 36.5 & 0.576 \\
 &  &  &   &  & Male-Black & 26,979 & 12.9 & 0.503 \\
 &  &  &   &  & Female-Black & 27,769 & 13.2 & 0.426 \\
\bottomrule
\end{tabular}}
\caption{Fairness datasets. On each dataset, we highlight the task according to the positive outcome $Y=1$. We further list the demographic groups considered on each dataset, and denote in bold the \textbf{favored group}.}
\label{tab:fairness-datasets}
\end{table}

\paragraph{Main Results.} We summarize our results in Table~\ref{tab:fairness-results}, which reports the classification accuracy (Acc.) and \gls{di} per dataset and barycenter solver, from which we draw a few observations. First, fairness is improved across the board for most solvers. For instance, the logistic regression fit on \texttt{Adult} has $\text{DI} = 0.509$, whereas a fit on repaired data through \texttt{BaryFM} gets $\text{DI} = 1.013$ at a cost of $1.16\%$ in classification accuracy. Second, the fairness axis makes a clear split between empirical (\texttt{DWB}, \texttt{WGF}) and neural solvers (e.g., \texttt{NormFlow}). Indeed, empirical solvers tend to compensate on the fairness axis while leaving classification accuracy behind, whereas neural solvers tend to preserve accuracy at the cost of a residual fairness gap. One notable example is \texttt{NWBU}, that preserves classification performance on \texttt{Adult} while degrading \gls{di}.

\begin{table}[ht]
\centering
\resizebox{\linewidth}{!}{%
\begin{tabular}{lcccccccc}
\toprule
Method & \multicolumn{2}{c}{Adult} & \multicolumn{2}{c}{COMPAS} & \multicolumn{2}{c}{Credit} & \multicolumn{2}{c}{NLSY} \\
\cmidrule(lr){2-3} \cmidrule(lr){4-5} \cmidrule(lr){6-7} \cmidrule(lr){8-9}
 & Acc & DI & Acc & DI & Acc & DI & Acc & DI \\
\midrule
\textit{Baseline} & \result{80.84}{0.19} & \result{0.509}{0.014} & \result{69.19}{2.25} & \result{0.662}{0.007} & \result{80.77}{0.28} & \result{0.974}{0.010} & \result{83.69}{0.12} & \result{0.840}{0.003} \\
\midrule
DWB & \result{78.82}{0.33} & \result{1.019}{0.013} & \result{67.02}{0.33} & \result{1.013}{0.008} & \result{80.02}{0.03} & \result{0.998}{0.009} & \result{81.27}{0.13} & \result{0.960}{0.006} \\
WGF & \result{78.67}{0.32} & \result{1.030}{0.029} & \result{65.94}{1.26} & \result{0.981}{0.022} & \result{77.03}{0.70} & \result{0.990}{0.004} & \result{81.00}{0.10} & \result{0.998}{0.004} \\
\midrule
NormFlow & \result{80.54}{0.19} & \result{0.652}{0.071} & \result{67.42}{2.32} & \result{0.871}{0.036} & \result{80.72}{0.31} & \result{0.984}{0.014} & \result{83.05}{0.26} & \result{0.942}{0.025} \\
NOT & \result{79.52}{0.33} & \result{1.074}{0.180} & \result{68.09}{0.33} & \result{0.916}{0.083} & \result{80.78}{0.30} & \result{1.002}{0.014} & \result{82.34}{0.22} & \result{1.014}{0.011} \\
NWB & \result{78.68}{0.70} & \result{0.499}{0.037} & \result{61.65}{3.53} & \result{1.505}{1.041} & \result{79.36}{0.36} & \result{0.944}{0.051} & \result{74.60}{3.56} & \result{0.944}{0.164} \\
NWBU & \result{80.87}{0.81} & \result{0.357}{0.222} & \result{68.40}{1.26} & \result{0.692}{0.047} & \result{80.52}{0.52} & \result{0.966}{0.042} & \result{81.91}{1.38} & \result{0.754}{0.138} \\
TDSB & \result{78.82}{0.18} & \result{0.830}{0.121} & \result{66.16}{2.52} & \result{0.759}{0.060} & \result{80.22}{0.13} & \result{0.983}{0.038} & \result{79.88}{0.20} & \result{0.877}{0.078} \\
TSBM & \result{79.67}{0.09} & \result{0.898}{0.016} & \result{66.51}{2.03} & \result{0.920}{0.006} & \result{80.52}{0.22} & \result{0.993}{0.015} & \result{80.52}{0.04} & \result{0.984}{0.009} \\
UNOT & \result{80.23}{0.07} & \result{0.780}{0.029} & \result{68.05}{0.98} & \result{0.853}{0.029} & \result{80.81}{0.38} & \result{0.987}{0.013} & \result{83.00}{0.18} & \result{0.947}{0.017} \\
CW2B & \result{79.50}{0.08} & \result{1.142}{0.035} & \result{67.17}{1.12} & \result{0.967}{0.036} & \result{80.72}{0.34} & \result{0.997}{0.010} & \result{83.19}{0.10} & \result{0.936}{0.012} \\
WIN & \result{78.86}{1.68} & \result{0.859}{0.484} & \result{68.18}{1.16} & \result{0.862}{0.079} & \result{80.70}{0.36} & \result{0.975}{0.005} & \result{81.68}{0.13} & \result{1.001}{0.010} \\
GMM & $80.67_{\pm 0.10}$ & $0.599_{\pm 0.023}$ & $67.33_{\pm 1.55}$ & $1.047_{\pm 0.073}$ & $80.77_{\pm 0.28}$ & $1.108_{\pm 0.175}$ & $83.56_{\pm 0.11}$ & $0.881_{\pm 0.004}$ \\
\midrule
\textbf{BaryFM} & \result{79.68}{0.76} & \result{1.013}{0.168} & \result{67.55}{2.08} & \result{1.023}{0.013} & \result{80.75}{0.18} & \result{0.992}{0.012} & \result{82.39}{0.16} & \result{1.001}{0.003} \\
\bottomrule
\end{tabular}}
\caption{Performance of total repair, per barycenter solver, across 4 fairness datasets.}
\label{tab:fairness-results}
\end{table}

\begin{wrapfigure}[24]{r}{0.6\textwidth}
  \centering
  \includegraphics[width=\linewidth]{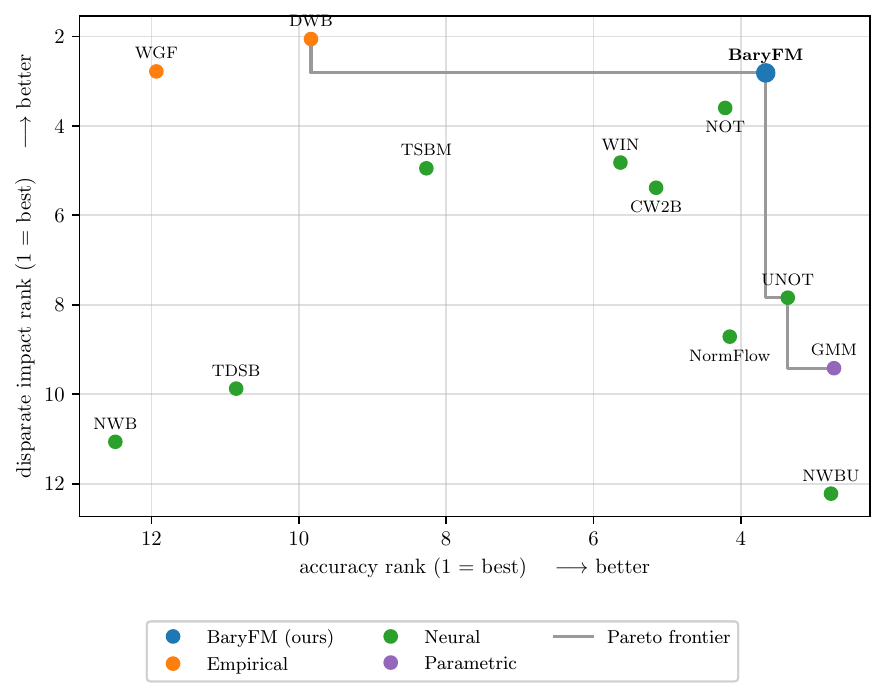}
  \caption{Fairness vs. classification accuracy tradeoff. For having a single view of methods performance across different datasets, we use the average rank on each axis of methods across datasets.}
  \label{fig:fairness-tradeoff}
\end{wrapfigure}
\paragraph{Fairness vs. Accuracy Tradeoff.} Figure~\ref{fig:fairness-tradeoff} summarizes the results of Table~\ref{tab:fairness-results} by computing the ranking of each metric obtained by of each method on each dataset, then averaging out over datasets. This averaged ranking gives a global view of how methods perform over the 4 tested datasets. Intuitively, methods are better by ranking smaller on both accuracy and fairness ranks, that is, being on the upper right corner of Figure~\ref{fig:fairness-tradeoff} (both axes are \textbf{decreasing}). This figure provides another view of our previous claim. Especially, empirical solvers such as \texttt{DWB} and \texttt{WGF} occupy the upper left corner of the figure (better fairness, worse downstream accuracy), while neural methods are more scattered over the plane, with a few (\texttt{NormFlow}, \texttt{UNOT} and \texttt{NWBU}) occupying the bottom right corner (better downstream accuracy, worse fairness). Our proposed \texttt{BaryFM} occupies the upper right corner, having strong fairness metric and downstream accuracy performance.

\begin{figure}[ht]
    \centering
    \begin{subfigure}{0.48\linewidth}
        \includegraphics[width=\linewidth]{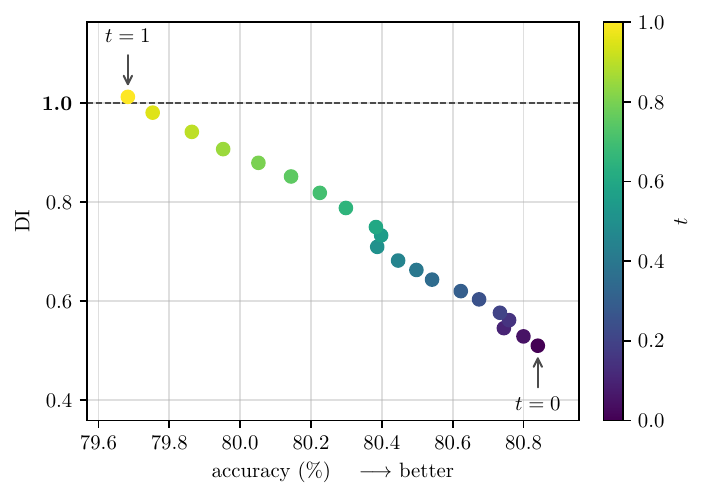}
        \caption{Adult}
    \end{subfigure}
    \begin{subfigure}{0.48\linewidth}
        \includegraphics[width=\linewidth]{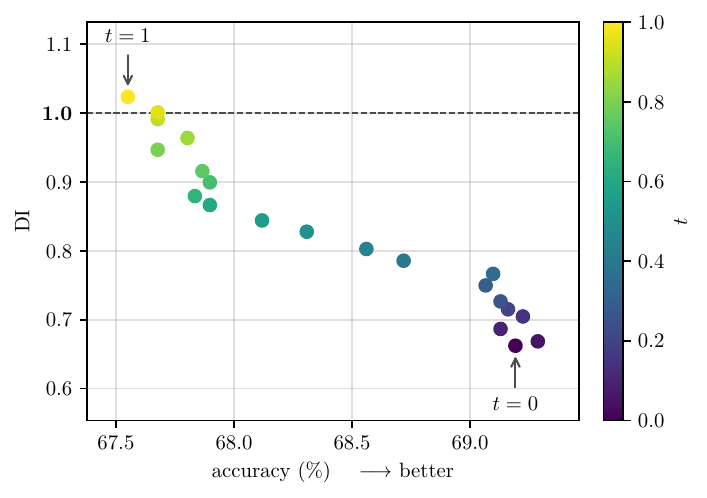}
        \caption{COMPAS}
    \end{subfigure}\\
    \begin{subfigure}{0.48\linewidth}
        \includegraphics[width=\linewidth]{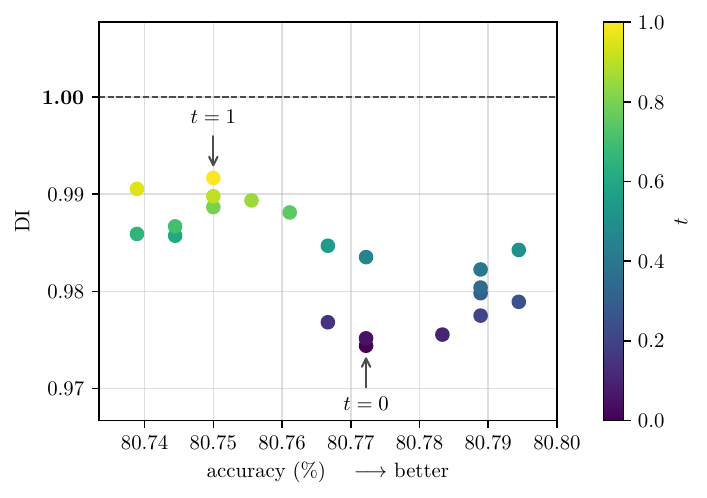}
        \caption{Credit}
    \end{subfigure}
    \begin{subfigure}{0.48\linewidth}
        \includegraphics[width=\linewidth]{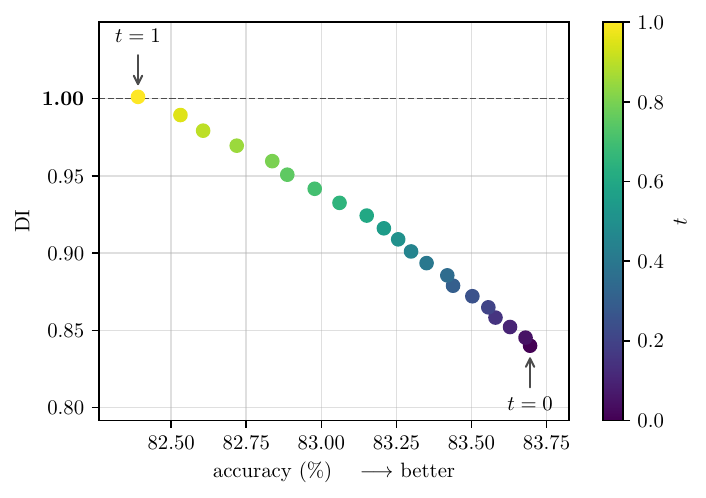}
        \caption{NLSY}
    \end{subfigure}
    \caption{Disparate Impact (DI) vs. classification accuracy as a function of flow time. At $t=0$, we have the starting point of samples at each $p_k$. As time evolves, these samples are progressively transported to the barycenter, which usually improves \gls{di} (models become more fair) at the cost of accuracy.}
    \label{fig:trajectory-fairness}
\end{figure}

\paragraph{Fairness vs. Accuracy across \gls{cfm} trajectories.} An interesting consequence of computing a barycenter through \gls{cfm} is that our algorithm is inherently dynamical. Since in our methodology we move each $p_k$ to $\hat{q}_{\lambda}^{(L_{\text{FP}})}$ through a flow for $t \in [0, 1]$, we can measure, at each step of the \gls{ode} trajectory, the \gls{di} and classification accuracy of partially repaired (in the transport sense). We show these trajectories in Figure~\ref{fig:trajectory-fairness}. At $t=0$, we retrieve the baseline performance (i.e., no repair). as time progresses, we see a progressive overall improvement of \gls{di} towards values close to $1.0$, and a degradation of downstream accuracy, again, highlighting the fairness-vs-task performance trade-off~\citep{zhao2022inherent}.

\paragraph{Hyperparameters and their sensitivity.} We use the same training and inference configurations for each dataset, namely: $B=1024$ (except for COMPAS, $B=512$), $N=30\,000$, $L_{\text{FP}} = L_{\text{D}} = 1$, Adam~\citep{kingma2014adam} optimizer with $\text{LR}_{\text{C}} = \text{LR}_{\text{D}} = 10^{-3}$, entropic regularization $\epsilon = 10^{-3}$, \# of \gls{ode} steps $T=100$, and an Euler solver. Every dataset uses an \gls{mlp} architecture with $\text{dim}_{h} = 256$ and $4$ layers in each head. These result in $645\,062$, $646\,601$, $653\,783$ and $645\,318$ parameters for Adult, COMPAS, Credit and NLSY respectively.

\begin{wrapfigure}[20]{r}{0.5\textwidth}
  \centering
  \includegraphics[width=0.8\linewidth]{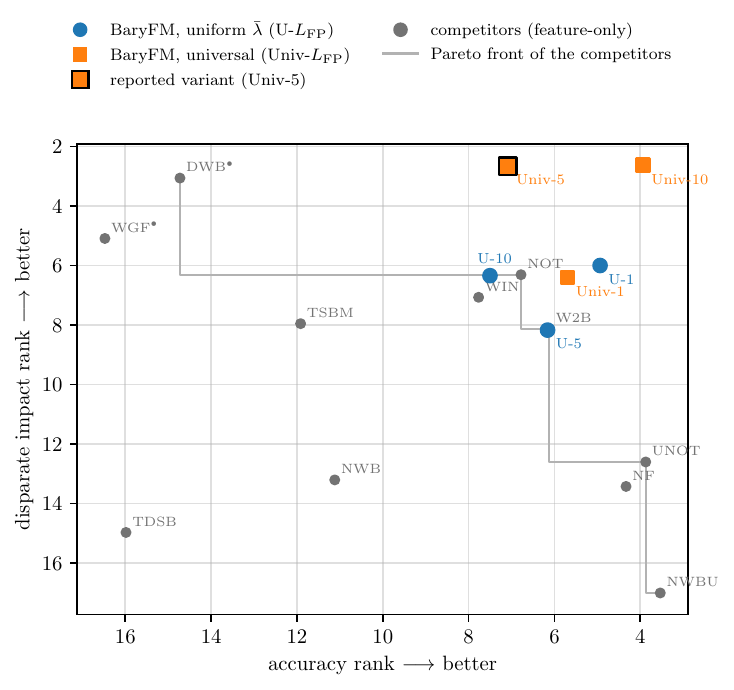}
  \caption{Fairness vs. classification accuracy tradeoff for \texttt{BaryFM} under different hyper-parameters.}
  \label{fig:hp-tradeoff}
\end{wrapfigure}
We study the parameter sensitivity of our \texttt{BaryFM} across 2 axis: universal vs. non-universal solver, and number of fixed point iterations $L_{\text{FP}}$. We show an overview in Figure~\ref{fig:hp-tradeoff}, which has $6$ variants, named $\{\text{U,Univ}\}-L_{\text{FP}}$, for $L_{\text{FP}} \in \{1, 5, 10\}$. Overall, 4 of the 6 variants rank consistently above the pareto-frontier defined by \texttt{DWB} (best fairness rank) -- \texttt{NOT} -- \texttt{CW2B} -- \texttt{U-NOT} -- \texttt{NWBU} (best accuracy rank). Especially, our $\text{Univ}-10$ (universal solver, $L_{\text{FP}} = 10$) has the best geometric mean fairness rank (1.86) against \texttt{DWB} (2.21), and is close to the methods with the best geometric mean accuracy ranks (ours 3.94 vs. \texttt{U-NOT} and \texttt{NWBU} with 3.87 and 3.53 respectively). In the end, we report our $\text{Univ}-5$ variant in Table~\ref{tab:fairness-results}, as it strikes the best balanced between the fairness metric (\gls{di}) and the downstream task metric (classification accuracy).

\section{Domain Adaptation}\label{sec:additional-da}

We evaluate barycenter algorithms in terms of their \emph{domain alignment performance} of extracted embeddings. In a nutshell, we fine-tune a model with the available source-domain data. This is either a pre-trained foundation model, such as CBraMod~\citep{wang2025cbramod} for EEG data, ResNets~\citep{he2016deep} pretrained on ImageNet for image data, or a convolutional neural net from scratch for chemical engineering data. This initial pre-training minimizes the risk over the pooled source domain data, which can be written as,
\begin{align}
    \theta^{\star} = \argmin{\theta \in \Theta} \dfrac{1}{K}\sum_{k=1}^{K}\dfrac{1}{n_{k}}\sum_{i=1}^{n_{k}}\ell(y_{k,i},h(x_{k,i})),\label{eq:erm-pooled-sources}
\end{align}
where $x_{k,i}$ is the $i-$th embedding from the $k-$th domain (resp. $y$ for labels), and $\Theta$ is the space of neural net weights. This first fine-tuning stage serves the purpose of separating the classes in the different datasets. As explored in~\cite[Appendix E]{montesuma2025computing}, the domain alignment performed by barycenter algorithms happens in 2 different ways, depending on whether the barycenter measure is labeled or not.

\subsection{Wasserstein Barycenter Transport}

Originally proposed by~\cite{montesuma2021wasserstein}, the \gls{wbt} algorithm solves multi-source domain adaptation in a 2 stage procedure. First, source domains are aggregated into the Wasserstein barycenter. Second, the Wasserstein barycenter is transported into the target domain. The original method was proposed with a labeled variant of the discrete algorithm of~\cite{cuturi2014fast}, later formalized as a barycentric transport over the joint continuous variable $z = (x, y)$, where $x \in \mathbb{R}^{d}$ and $y \in \Delta_{C}$, for $C-$classes by~\cite{montesuma2023multi}.

In~\cite{montesuma2025computing}, the authors have extended the \gls{wbt} methodology for other barycenter solvers, allowing the comparison of barycenter solvers for the downstream task of domain adaptation. Therefore, the comparison along this application judge barycenter in terms of (1) their ability to align heterogeneous domains, and (2) their ability to respect class structure. There is a supplementary, 3rd axis that is how well the algorithms approximate the mappings from the marginal measures $p_k$ to the barycenter support $q_{\lambda}^{\star}$. An illustration of the unsupervised and supervised mechanisms is shown in Figure~\ref{fig:WBT-comparison}.

In both unsupervised and supervised cases, the barycenter points must be aligned with the target domain. We isolate the quality of the calculated barycenter by subjecting all methods to the same alignment mechanism. Here, note that \gls{wbt} is effectively reducing a multi-source domain adaptation problem to a single-source one. Therefore, we use~\cite{courty2016optimal}, which transports the support of the estimated barycenter $\hat{q}_{\lambda}$ to that of $p_{T}$,
\begin{align}
    T_{q_{\lambda} \rightarrow p_{T}}(\tilde{x}_{k,i}) = n_{T}\sum_{j=1}^{n_{T}}\gamma_{ij}^{\star}x_{T,j}.\qquad\gamma^{\star} = \text{OT}(q_{\lambda}^{\star}, p_{T}).\label{eq:domain-alignment-target}
\end{align}

However, this method has limited scalability. For instance, in the context of DomainNet, domains such as \texttt{quickdraw} and \texttt{real} have around $170,000$ points, making the required \gls{ot} solve unfeasible. We take a way around this issue by solving a \gls{cfm} problem for aligning the barycenter with the target domain, meaning,
\begin{align*}
    \theta_{T}^{\star} = \argmin{\theta \in \Theta} \mathbb{E}_{t,z_0,z_1}[\lVert v_{\theta}(x_t, t) - (x_1 - x_0) \rVert_{2}^{2}]\text{, where }t\sim\mathcal{U}[0,1]\text{ and }(x_0, x_1)\sim \text{OT}(\hat{q}_{\lambda},p_T),
\end{align*}
where $\text{OT}(\hat{q}_{\lambda},p_T)$ is an \gls{ot} solve between minibatches from the estimated barycenter $\hat{q}_{\lambda}$ and from $p_{T}$. Once trained, the barycenter samples are transported into the target domain via the same \gls{ode} defined in equation~\ref{eq:ode-inference}.

\subsection{Experimental Details}

\noindent\textbf{Adaptation Protocol.} Algorithms are evaluated in 2 protocols, depending on the established literature. For instance, in computer vision, datasets have a small number of domains (e.g., Office 31 has 3 domains), and thus, algorithms are evaluated in the \gls{lodo} setting. Taking Office 31 as an example, this dataset has 3 domains: Amazon (\texttt{A}), dSLR (\texttt{D}) and Webcam (\texttt{W}). In the \gls{lodo} protocol, one has 3 adaptation tasks: $\{\texttt{D}, \texttt{W}\} \to \texttt{A}$, $\{\texttt{A}, \texttt{W}\} \to \texttt{D}$, and $\{\texttt{A}, \texttt{W}\} \to \texttt{W}$. Performance is evaluated and ranked in terms of \emph{average target domain performance}.

The second adaptation protocol is \emph{fixed target}, which commonly appears in cross-subject adaptation studies. In that case, there is a fixed partition of source subjects (i.e., subjects from which we have available labeled data and from which the barycenter is calculated) and target subjects (i.e., subjects to which adaptation is done). To keep results comparable to those reported by the backbone~\citep{wang2025cbramod}, we report \emph{pooled target domain performance}.

\begin{table}[ht]
    \centering
    \resizebox{\linewidth}{!}{
        \begin{tabular}{cccccccc}
            \toprule
            Dataset & Adaptation Protocol & Backbone & \# Features & \# Classes & \# Domains & \# Samples & Seeds\\
            \midrule
            Office 31 & LODO & ResNet-50 & 2048 & 31 & 3 & 4,110 & 42, 43, 44\\
            Office-Home & LODO & ResNet-101 & 2048 & 65 & 4 & 15,500 & 42, 43, 44\\
            DomainNet & LODO & ResNet-101 & 2048 & 345 & 6 & 586,575 & 42, 43, 44\\
            \midrule
            BCI-CIV-2a & Fixed Target & CBraMod & 200 & 4 & 9 & 5,184 & 42, 43, 44\\
            FACED & Fixed Target & CBraMod & 200 & 9 & 123 & 10,332 & 42, 43, 44\\
            SEED-VIG & Fixed Target & CBraMod & 200 & -- & 21 & 20,355 & 42, 43, 44\\
            Mumtaz & Fixed Target & CBraMod & 200 & 2 & 5 & 7,143 & 42, 43, 44\\
            Physio & Fixed Target & CBraMod & 200 & 4 & 109 & 9,837 & 42, 43, 44\\
            SHU-MI & Fixed Target & CBraMod & 200 & 2 & 25 & 11,988 & 42, 43, 44\\
            \midrule
            TEP & LODO & CNN & 128 & 29 & 6 & 17,289 & 42 (pre-defined 5 folds)\\
            \bottomrule
        \end{tabular}
    }
    \caption{Summary of datasets used in our experiments.}
    \label{tab:datasets}
\end{table}

\paragraph{Architecture.} All domain adaptation experiments use the same network architecture, which follows the design presented in Section~\ref{sec:methodology}. Our networks are composed of a shared encoder, that different layers to map $x, y, k, \lambda$ into a shared latent space $\mathbb{R}^{\text{dim}_{h}}$. In a nutshell, continuous inputs such as $x \in \mathbb{R}^{\text{dim}_{x}}, y \in \Delta_{C}, \lambda \in \Delta_{K}$ are mapped into $\mathbb{R}^{\text{dim}_{h}}$ through a dense layer. Since $k \in \{1,\cdots,K\}$ is inherently discrete, we use a lookup table with $K$ vectors in $\mathbb{R}^{\text{dim}_h}$.

On top of the latent vector $h = \text{Cat}(E_{x}, E_{y}, E_{k}, E_{\lambda})$, we use 2 heads, a \gls{cfm} and a dual \gls{ot} head. The \gls{cfm} head receives, additionally $h_t = \text{Cat}(h, E_{t})$, where $E_{t}$ is a sinusoidal position encoding of the time variable $t\in [0, 1]$. Both heads implement the same architecture, that is, a \gls{mlp} with 4 layers, and SiLU activations. We summarize this discussion in Table~\ref{tab:arch-da}.

\begin{table}[ht]
    \centering
        \begin{tabular}{lccccccccc}
            \toprule
            Dataset & Hidden width & \gls{mlp} embedding & $x$ & $t$ & $y$ & $k$ & $\lambda$ & \# Params\\
            \midrule
            Office 31   & 512 & \cmark & 64  & 64 & 64  & 64  & 64 & 4.09M\\
            Office-Home & 512 & \xmark & 128 & 32 & 32  & 64  & 64 & 3.24M\\
            DomainNet   & 512 & \xmark & 128 & 64 & 128 & 64  & 64 & 3.54M\\
            \midrule
            BCI-CIV-2a  & 512 & \cmark & 64  & 64 & 64  & 64  & 64 & 2.18M\\
            FACED       & 512 & \cmark & 128 & 64 & 64  & 64  & 64 & 2.36M\\
            SEED-VIG    & 512 & \cmark & 128 & 64 & 32  & 128 & 64 & 2.38M\\
            Mumtaz      & 512 & \cmark & 64  & 64 & 64  & 64  & 64 & 2.18M\\
            Physio      & 512 & \cmark & 64  & 64 & 64  & 64  & 64 & 2.19M\\
            SHU-MI      & 512 & \cmark & 64  & 64 & 64  & 64  & 64 & 2.18M\\
            \midrule
            TEP         & 512 & \cmark & 64  & 64 & 64  & 64  & 64 & 2.12M\\
            \bottomrule
        \end{tabular}
    \caption{Architecture of the network used in domain adaptation. Columns $x$, $t$, $y$, $k$ and $\lambda$ give the embedding dimensions of the features, time, labels, domain index and barycentric weights. The last column gives the total number of trainable parameters, including the dual \gls{ot} head.}
    \label{tab:arch-da}
\end{table}

\paragraph{Hyperparameters.} We divide the hyperparameters of our algorithm in 2 families. First, there are the training hyperparameters, which are: batch size $B$, number of iterations $N$, number of fixed point iterations $L_{\text{FP}}$, number of dual iterations $L_{\text{D}}$, learning rate $\text{LR} = \text{LR}_{C} = \text{LR}_{D}$ (that is, we use a single learning rate for the whole network), and entropic regularization $\epsilon$. We use the same architecture on all benchmarks, consisting of a densely connected multi-layer perceptron. We train this network using the AdamW~\citep{kingma2014adam, loshchilov2017decoupled} optimizer with weight decay $\omega=10^{-2}$. We show an overview in Table~\ref{tab:train-hps-da}.

\begin{table}[ht]
    \centering
    \begin{tabular}{lccccccc}
        \toprule
        Benchmark & $B$ & $N$ & $L_{\text{Dual}}$ & $L_{\text{FP}}$ & $\text{LR}$ & Weight Decay & $\varepsilon$ \\
        \midrule
        Office 31    &  310 & 10\,000 & 1 & 1 & $10^{-3}$ & 0.01 & $10^{-3}$\\
        Office Home  & 1300 & 10\,000 & 1 & 1 & $10^{-4}$ & 0    & $10^{-3}$\\
        DomainNet    & 2048 & 30\,000 & 1 & 1 & $10^{-4}$ & 0    & $10^{-3}$\\
        \midrule
        BCI-CIV-2a   &  400 & 10\,000 & 1 & 1 & $10^{-3}$ & 0.01 & $10^{-3}$\\
        FACED        &  256 & 10\,000 & 1 & 1 & $10^{-4}$ & 0.01 & $10^{-3}$\\
        SEED-VIG     & 1024 & 10\,000 & 5 & 5 & $10^{-4}$ & 0.01 & $10^{-3}$\\
        Mumtaz       & 1024 & 10\,000 & 1 & 1 & $10^{-4}$ & 0.01 & $10^{-3}$\\
        Physio       & 1024 & 10\,000 & 1 & 1 & $10^{-4}$ & 0.01 & $10^{-3}$\\
        SHU-MI       & 1024 & 10\,000 & 1 & 1 & $10^{-4}$ & 0.01 & $10^{-3}$\\
        \midrule
        TEP          &  400 & 10\,000 & 1 & 1 & $10^{-3}$ & 0.01 & $10^{-2}$\\
        \bottomrule
    \end{tabular}
    \caption{Training hyperparameters of \texttt{BaryFM} across benchmarks.}
    \label{tab:train-hps-da}
\end{table}

\begin{figure}[ht]
    \centering
    \begin{subfigure}{0.45\linewidth}
        \includegraphics[width=\linewidth]{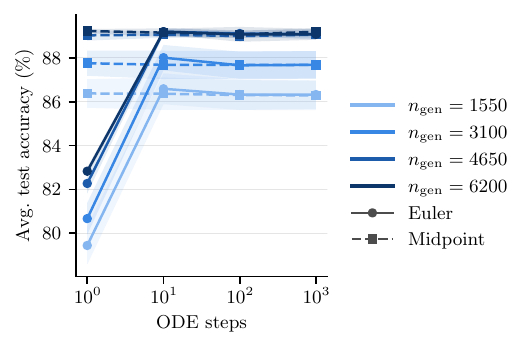}
        \caption{Office 31 $n_{\text{gen}}$ \& ODE Solver.}
    \end{subfigure}
        \begin{subfigure}{0.36\linewidth}
        \includegraphics[width=\linewidth]{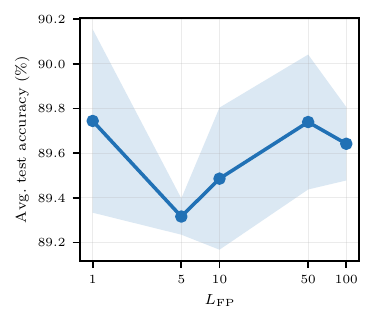}
        \caption{Office 31 ($L_{\text{FP}}$).}
    \end{subfigure}\\
    \begin{subfigure}{0.45\linewidth}
        \includegraphics[width=\linewidth]{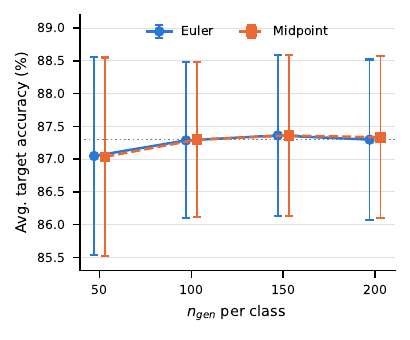}
        \caption{TEP ($n_{\text{gen}}$)}
    \end{subfigure}
    \begin{subfigure}{0.45\linewidth}
        \includegraphics[width=\linewidth]{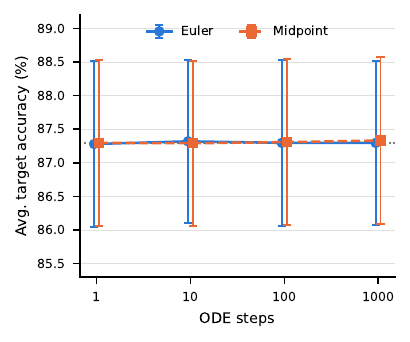}
        \caption{TEP ($T$)}
    \end{subfigure}
    \caption{Hyper-parameter sensitivity to $n_{\text{gen}}$, ODE solver, \# of fixed-point iterations ($L_{\text{FP}}$), and \# of ODE steps ($T$) on Office 31 and TEP benchmarks.}
    \label{fig:office31-inference-ablation}
\end{figure}

Second, there are inference hyper-parameters, which comprise number of generated samples, number of \gls{ode} steps, and \gls{ode} solver. In general, $\text{ODE steps} > 10$ suffice for obtaining the best performance, and there is not much difference in the ODE solver choice. We ablate these results in Figure~\ref{fig:office31-inference-ablation}. For simplicity, we use $100$ steps with an Euler solver. The number of generated points, on the other hand, has more leverage on performance, as sampling more points in the barycenter support increases performance. For the sake of a fair comparison, we use the same number of points in each barycenter solver in Table~\ref{tab:results-msda}.

\begin{table}[ht]
    \centering
\resizebox{\linewidth}{!}{
    \begin{tabular}{lcccccccccc}
        \toprule
        & Office 31 & Office Home & DomainNet & BCI-CIV-2a & FACED & SEED-VIG & Mumtaz & Physio & SHU-MI & TEP \\
        \midrule
        $n_{\text{gen}}$ & 4\,650 & 39\,000 & 103\,500 & 1\,000 & 9\,000 & 53\,100 & 10\,000 & 10\,000 & 10\,000 & 5\,800 \\
        ODE steps       & 100 & 100 & 100 & 100 & 100 & 100 & 100 & 100 & 100 & 100 \\
        ODE Solver      & Euler & Euler & Euler & Euler & Euler & Euler & Euler & Euler & Euler & Euler \\
        Transport       & \texttt{emd} & \texttt{emd} & \texttt{cfm} & \texttt{emd} & \texttt{emd} & \texttt{emd} & \texttt{emd} & \texttt{emd} & \texttt{emd} & \texttt{emd} \\
        \bottomrule
    \end{tabular}
}
    \caption{Inference hyperparameters of \texttt{BaryFM} across benchmarks.}
    \label{tab:inference-hps-da}
\end{table}

\paragraph{Main Results.} Our main results are shown in Table~\ref{tab:results-msda}, which comprehend the aggregate performance on each benchmark. Note that Office 31, Office Home, DomainNet and TEP report average target domain performance, and neuroscience datasets (BCI-CIV-2a, FACED, SEED-VIG, Mumtaz, Physio, SHU-MI) report pooled performance over target subjects. This reporting choices are done to conform to usual practice. See, for instance, how~\cite{wang2025cbramod} report results on multiple held-out  subjects. As we mentioned in the main paper, our \texttt{BaryFM} achieves an average rank of 1.75, surpassing the previous state-of-the-art barycenter solver for domain adaptation, \texttt{WGF}~\citep{montesuma2025computing}. We provide detailed descriptions of these benchmarks and their results per domain in Sections~\ref{sec:o31} through~\ref{sec:tep}.

\begin{table}[ht]
    \centering
\resizebox{\linewidth}{!}{
\begin{tabular}{lccccccccccccc}
    \toprule
    Method & $\mathcal{X}\times\mathcal{Y}$ & Universal & Office31 & OfficeHome & DomainNet & BCI-CIV-2a & FACED & SEED-VIG & Mumtaz & Physio & SHU-MI & TEP & Avg. Rank \\
    \midrule
    Backbone & - & - & ResNet-50 & \multicolumn{2}{c}{ResNet-101} & \multicolumn{6}{c}{CBraMod} & CNN & - \\
    \midrule
    Source-Only & - & - & \result{86.58}{0.07} & \result{76.93}{0.09} & \result{51.23}{0.01} & \result{52.72}{0.05} & \result{54.87}{0.27} & \result{44.72}{2.15} & \result{91.25}{0.36} & \result{64.10}{0.08} & \result{62.68}{0.41} & \result{79.18}{0.18} & 12.10 \\
    \midrule
    Discrete \xmark & \xmark & \xmark & \result{85.76}{0.56} & \result{75.95}{0.05} & \result{50.46}{0.20} & \result{62.62}{0.41} & \result{54.47}{0.32} & \result{34.66}{0.50} & \result{91.30}{0.13} & \result{64.95}{0.29} & \result{55.89}{3.26} & \result{84.54}{1.47} & 11.45 \\
    Discrete \cmark & \cmark & \xmark & \result{85.46}{0.36} & \result{76.09}{0.36} & \result{51.01}{0.10} & \result{62.50}{0.63} & \result{54.43}{0.50} & \result{34.76}{0.60} & \result{91.30}{0.13} & \result{64.97}{0.27} & \result{53.80}{4.03} & \result{85.23}{1.26} & 11.40 \\
    WGF \xmark & \xmark & \xmark & \result{86.69}{0.33} & \result{77.09}{0.28} & \result{49.48}{0.04} & \result{63.02}{0.23} & \result{58.06}{0.24} & \result{52.29}{0.62} & \result{92.79}{0.41} & \result{65.12}{0.40} & \result{62.87}{0.90} & \result{85.52}{1.48} & 5.40 \\
    WGF \cmark & \cmark & \xmark & \result{88.90}{0.74} & \result{78.03}{0.13} & \result{52.08}{0.14} & \result{62.56}{0.13} & \result{56.99}{0.24} & \result{51.74}{0.44} & \result{91.77}{0.56} & \result{65.54}{0.23} & \result{63.70}{0.11} & \result{86.87}{1.27} & 4.50 \\
    \midrule
    CW2B & \xmark & \xmark & \result{87.12}{1.27} & \result{75.93}{0.19} & \result{49.14}{0.69} & \result{62.70}{0.35} & \result{58.07}{0.18} & \result{51.06}{0.74} & \result{91.88}{0.19} & \result{64.97}{0.14} & \result{63.05}{0.20} & \result{83.00}{1.55} & 7.30 \\
    WIN & \xmark & \xmark & \result{86.21}{0.53} & \result{75.88}{0.17} & \result{46.89}{0.49} & \result{59.95}{3.69} & \result{58.83}{0.43} & \result{49.78}{2.47} & \result{91.58}{0.08} & \result{64.60}{0.03} & \result{62.96}{0.32} & \result{82.58}{1.95} & 11.05 \\
    NOT & \xmark & \xmark & \result{87.38}{0.85} & \result{76.35}{0.54} & \result{49.74}{0.07} & \result{62.73}{0.10} & \result{57.94}{0.11} & \result{42.23}{2.74} & \result{91.63}{0.13} & \result{64.97}{0.28} & \result{62.99}{0.52} & \result{83.35}{2.10} & 7.80 \\
    U-NOT & \xmark & \xmark & \result{86.42}{0.67} & \result{76.21}{0.12} & \result{49.84}{0.15} & \result{62.64}{0.81} & \result{57.85}{0.26} & \result{43.98}{1.59} & \result{91.74}{0.22} & \result{64.84}{0.17} & \result{62.92}{0.21} & \result{83.45}{1.92} & 9.30 \\
    TDSB & \xmark & \xmark & \result{86.74}{0.19} & \result{74.97}{0.15} & \result{52.56}{0.13} & \result{62.64}{0.18} & \result{57.04}{0.09} & \result{49.58}{0.87} & \result{91.74}{0.00} & \result{64.87}{0.08} & \result{63.14}{0.21} & \result{83.76}{1.50} & 8.10 \\
    TSBM & \xmark & \xmark & \result{86.96}{0.04} & \result{76.11}{0.30} & \result{52.59}{0.02} & \result{62.50}{0.31} & \result{55.99}{0.65} & \result{55.03}{0.35} & \result{92.07}{0.22} & \result{64.78}{0.34} & \result{63.06}{0.70} & \result{84.42}{1.24} & 6.75 \\
    NormFlow & \xmark & \xmark & \result{86.88}{0.28} & \result{75.91}{0.08} & \result{52.28}{0.07} & \result{62.44}{0.22} & \result{58.01}{0.13} & \result{46.13}{1.83} & \result{91.80}{0.05} & \result{64.74}{0.03} & \result{62.95}{0.26} & \result{83.92}{1.30} & 8.55 \\
    GMM & \cmark & \xmark & \result{87.18}{0.99} & \result{75.62}{0.57} & \result{51.72}{0.02} & \result{61.95}{0.52} & \result{55.49}{0.07} & \result{41.25}{1.87} & \result{91.58}{0.08} & \result{65.54}{0.08} & \result{63.49}{0.46} & \result{85.34}{0.63} & 8.20 \\
    NWB & \xmark & \xmark & \result{85.07}{0.67} & \result{73.31}{0.37} & \result{38.47}{4.23} & \result{62.41}{0.54} & \result{57.38}{0.08} & \result{39.60}{2.25} & \result{91.80}{0.17} & \result{65.04}{0.19} & \result{63.14}{0.60} & \result{83.36}{1.65} & 10.50 \\
    \midrule
    NWBU & \xmark & \cmark & \result{84.71}{0.81} & \result{74.29}{0.36} & \result{37.41}{1.82} & \result{62.70}{0.88} & \result{57.06}{0.21} & \result{37.15}{2.92} & \result{91.14}{0.75} & \result{65.00}{0.06} & \result{63.05}{0.12} & \result{80.57}{1.50} & 11.85 \\
    \rowcolor{lightblue!50} BaryFM & \cmark & \cmark & \result{89.74}{0.23} & \result{78.72}{0.12} & \result{52.56}{0.07} & \result{64.53}{0.65} & \result{58.25}{0.03} & \result{57.42}{0.94} & \result{93.12}{0.59} & \result{65.00}{0.11} & \result{63.70}{0.15} & \result{87.30}{1.39} & 1.75 \\
    \bottomrule
\end{tabular}
}
\caption{Downstream performance of WBT per barycenter solver. Results report $\mu_{\pm \sigma}$ over 3 seeds. Avg. Rank averages the per-dataset ranks.}
\label{tab:results-msda}
\end{table}

\paragraph{Ablations.} In Table~\ref{tab:ablation}, we ablate the components of our \texttt{BaryFM} method, meaning: \gls{cfm}, dual \gls{ot}, joint flow ($\mathcal{X}\times\mathcal{Y}$) and universality. We provide further information on each variant in Section~\ref{appx:alternatives}. Our methodology is as follows: each ablation stage replaces \texttt{BaryFM} in Table~\ref{tab:results-msda} and is thus ranked independently (i.e., one variant is not ranked against the others). There are 2 components that stand out in Table~\ref{tab:ablation}, and degrade performance the most. The first component is the \gls{cfm} head, which validates that learning the flow from each $p_k$ to $\hat{q}_{\lambda}^{(L_{\text{FP}})}$ is necessary. The second is universality, which means that learning universal barycenters leads to a better estimation of the uniform barycenter. A similar result was obtained in the context of Bayesian Posterior Aggregation (c.f. Section~\ref{sec:additional-bpa}, Table~\ref{tab:comparison_bayesian}). Other components (e.g., $\mathcal{X}\times\mathcal{Y}$ and the dual \gls{ot} head) do improve performance, but not as much. In addition to these remarks, note that the worst variants (non-universal \texttt{BaryFM}, and \texttt{NeuralFP}) rank $3.50$, which is still better than the 2nd best (\texttt{WGF}).

\begin{table}[ht]
    \centering
\resizebox{\linewidth}{!}{
\begin{tabular}{ccccccccccccccc}
    \toprule
     & & & & Office31 & OfficeHome & DomainNet & BCI-CIV-2a & FACED & SEED-VIG & Mumtaz & Physio & SHU-MI & TEP & Avg. Rank \\
    \midrule
    CFM & Dual OT & $\mathcal{X}\times\mathcal{Y}$ & Universal & ResNet-50 & \multicolumn{2}{c}{ResNet-101} & \multicolumn{6}{c}{CBraMod} & CNN & - \\
    \midrule
    \xmark & \cmark & \cmark & \cmark & \result{88.53}{0.11} & \result{78.61}{0.02} & \result{52.18}{0.09} & \result{64.93}{0.19} & \result{56.78}{0.10} & \result{47.65}{3.33} & \result{91.94}{0.17} & \textbf{\result{65.15}{0.09}} & \result{63.63}{0.49} & \result{87.07}{1.32} & 3.50 \\
    \cmark & \xmark & \cmark & \cmark & \result{89.09}{0.33} & \result{78.68}{0.04} & \result{52.54}{0.12} & \result{64.47}{0.28} & \textbf{\result{59.27}{0.16}} & \result{53.52}{0.49} & \result{93.12}{0.27} & \result{64.98}{0.21} & \result{63.61}{0.66} & \result{87.26}{1.15} & 1.90 \\
    \cmark & \cmark & \xmark & \cmark & \result{89.03}{0.18} & \result{78.60}{0.07} & \textbf{\result{52.70}{0.04}} & \textbf{\result{65.22}{0.53}} & \result{57.95}{0.39} & \result{52.15}{0.65} & \result{92.05}{0.48} & \result{65.00}{0.17} & \result{63.61}{0.30} & \textbf{\result{87.32}{1.20}} & 2.35 \\
    \cmark & \cmark & \cmark & \xmark & \result{87.88}{0.27} & \result{78.65}{0.09} & \result{52.45}{0.04} & \result{64.79}{0.18} & \result{57.82}{0.23} & \result{49.99}{0.06} & \result{92.35}{0.99} & \result{64.89}{0.17} & \result{63.47}{0.28} & \result{87.14}{1.22} & 3.50 \\
    \cmark & \cmark & \cmark & \cmark & \textbf{\result{89.74}{0.23}} & \textbf{\result{78.72}{0.12}} & \result{52.56}{0.07} & \result{64.53}{0.65} & \result{58.25}{0.03} & \textbf{\result{57.42}{0.94}} & \textbf{\result{93.12}{0.59}} & \result{65.00}{0.11} & \textbf{\result{63.70}{0.15}} & \result{87.30}{1.39} & 1.75 \\
    \bottomrule
\end{tabular}
}
\caption{Architectural ablation of \texttt{BaryFM}. Results report $\mu_{\pm \sigma}$ over 3 seeds. The rank of each method is computed by replacing \texttt{BaryFM} in Table~\ref{tab:results-msda}; Avg. Rank averages the per-dataset ranks.}
\label{tab:ablation}
\end{table}

\paragraph{Hyperparameter Sensitivity.} Besides the specific components of our \texttt{BaryFM}, we investigate how our method is sensitive to inference hyper-parameters, namely $n_{\text{gen}}$, \gls{ode} solver, number of \gls{ode} steps $T$, and number of fixed-point iterations $L_{\text{FP}}$. Furthermore, $L_{\text{FP}}$ does not have much effect on the downstream performance, as it oscillates between an average domain performance of 89.2 and 89.7. For simplicity, we use $L_{\text{FP}} = 1$ on most benchmarks. From the curves in Figure~\ref{fig:office31-inference-ablation} (a), we see that the main driver of performance in Office 31 is the number of generated points, while for TEP, performance is globally stable across $n_{\text{gen}}$. The choice of \gls{ode} solver is also less important, except for extremely small $T$ on Office 31 (c.f. Figure~\ref{fig:office31-inference-ablation} (a), \gls{ode} steps = $10^{0}$). Overall, we conclude that our algorithm is generally robust to the \gls{ode} hyper-parameters, and that $n_{\text{gen}}$ sensitivity depends on the tested dataset.

\paragraph{Number of Parameters.} We count the number of parameters in each tested neural method across the 10 benchmarks. We summarize the results in Table~\ref{tab:n-parameters}. Furthermore, we note that FACED and Physio demand significantly more parameters of neural networks that use $\mathcal{O}(K)$ networks to approximate Wasserstein barycenters, such as \texttt{CW2B}~\citep{korotin2021continuous} or \texttt{TSBM}~\citep{howard2025schrdinger}. For instance, BCI-IV-2a (5 sources) demands $\approx 10^{6}$ parameters for \texttt{CW2B} and $\approx 6.9\times10^{6}$ for \texttt{TSBM}, while FACED demand $\approx 1.6\times 10^{7}$ parameters and $\approx1.11\times10^{8}$, respectively.

\begin{table}[ht]
    \centering
\resizebox{\linewidth}{!}{
\begin{tabular}{lrrrrrrrrrr}
    \toprule
    Method & Office31 & OfficeHome & DomainNet & BCI-IV-2a & FACED & SEED-VIG & Mumtaz & Physio & SHU-MI & TEP \\
    \midrule
    CW2B & 3{,}957{,}504 & 5{,}936{,}256 & 9{,}893{,}760 & 1{,}023{,}360 & 16{,}373{,}760 & 2{,}660{,}736 & 1{,}023{,}360 & 14{,}327{,}040 & 3{,}070{,}080 & 677{,}760 \\
    WIN & 3{,}979{,}140 & 5{,}688{,}966 & 9{,}108{,}618 & 1{,}518{,}882 & 23{,}033{,}832 & 3{,}813{,}810 & 1{,}518{,}882 & 20{,}165{,}172 & 4{,}387{,}542 & 1{,}223{,}178 \\
    NOT & 1{,}643{,}778 & 2{,}465{,}667 & 4{,}109{,}445 & 552{,}045 & 8{,}832{,}720 & 1{,}435{,}317 & 552{,}045 & 7{,}728{,}630 & 1{,}656{,}135 & 413{,}445 \\
    U-NOT & 1{,}643{,}778 & 2{,}465{,}667 & 4{,}109{,}445 & 552{,}045 & 8{,}832{,}720 & 1{,}435{,}317 & 552{,}045 & 7{,}728{,}630 & 1{,}656{,}135 & 413{,}445 \\
    TDSB & 2{,}453{,}760 & 3{,}680{,}640 & 6{,}134{,}400 & 2{,}567{,}760 & 41{,}084{,}160 & 6{,}676{,}176 & 2{,}567{,}760 & 35{,}948{,}640 & 7{,}703{,}280 & 2{,}428{,}800 \\
    TSBM & 8{,}946{,}688 & 13{,}420{,}032 & 22{,}366{,}720 & 6{,}941{,}520 & 111{,}064{,}320 & 18{,}047{,}952 & 6{,}941{,}520 & 97{,}181{,}280 & 20{,}824{,}560 & 3{,}168{,}000 \\
    NormFlow & 3{,}658{,}176 & 3{,}696{,}560 & 1{,}723{,}904 & 201{,}152 & 359{,}552 & 218{,}048 & 201{,}152 & 338{,}432 & 222{,}272 & 463{,}760 \\
    NWB & 2{,}428{,}356 & 3{,}506{,}246 & 5{,}662{,}026 & 674{,}274 & 10{,}275{,}624 & 1{,}698{,}418 & 674{,}274 & 8{,}995{,}444 & 1{,}232{,}424 & 277{,}984 \\
    NWBU & 1{,}959{,}300 & 2{,}801{,}350 & 4{,}486{,}986 & 681{,}954 & 11{,}872{,}104 & 1{,}738{,}354 & 681{,}954 & 10{,}213{,}684 & 2{,}007{,}574 & 533{,}706 \\
    \midrule
    \texttt{NeuralFP} & 2{,}004{,}545 & 1{,}201{,}441 & 1{,}293{,}121 & 1{,}057{,}025 & 1{,}132{,}545 & 1{,}140{,}641 & 1{,}056{,}897 & 1{,}065{,}345 & 1{,}058{,}177 & 1{,}021{,}761 \\
    \texttt{BaryFM}$_{\text{Emp}}$ & 3{,}169{,}439 & 2{,}301{,}281 & 2{,}552{,}985 & 1{,}260{,}044 & 1{,}403{,}729 & 1{,}407{,}721 & 1{,}258{,}890 & 1{,}268{,}364 & 1{,}260{,}170 & 1{,}200{,}669 \\
    \texttt{BaryFM}$_{\text{unsup}}$ & 4{,}005{,}953 & 3{,}169{,}473 & 3{,}186{,}113 & 2{,}112{,}137 & 2{,}285{,}705 & 2{,}343{,}561 & 2{,}112{,}137 & 2{,}120{,}457 & 2{,}113{,}417 & 2{,}038{,}337 \\
    \texttt{BaryFM} (no Universal) & 4{,}089{,}504 & 3{,}237{,}730 & 3{,}538{,}586 & 2{,}180{,}109 & 2{,}356{,}562 & 2{,}376{,}938 & 2{,}178{,}955 & 2{,}188{,}429 & 2{,}180{,}235 & 2{,}120{,}734 \\
    \texttt{BaryFM} & 4{,}089{,}504 & 3{,}237{,}730 & 3{,}538{,}586 & 2{,}180{,}109 & 2{,}356{,}562 & 2{,}376{,}938 & 2{,}178{,}955 & 2{,}188{,}429 & 2{,}180{,}235 & 2{,}120{,}734 \\
    \bottomrule
\end{tabular}
}
    \caption{Number of trainable parameters of neural barycenter solvers.}
    \label{tab:n-parameters}
\end{table}

In comparison, our proposed architectures remain bounded as a function of the number of marginal measures (e.g., $\approx 2.1\times 10^{6}$ for BCI-IV-2a, and $\approx 2.3\times 10^{6}$ for FACED). From Table~\ref{tab:n-parameters}, the only other method that is able to do this is \texttt{NormFlow}~\citep{visentin2025computing}, who relies on the multi-scale architecture of~\cite{dinh2016density}.

Overall, Table~\ref{tab:n-parameters} shows a gap in the architecture design of neural network solvers, as there is no current consensus. Future work could focus on several computational questions around this issue, such as (i) scaling laws, and (ii) theoretically justified architectures.

\clearpage
\newpage

\subsection{Detailed Results and Dataset Descriptions}

\subsubsection{Office 31}\label{sec:o31}

Proposed by~\cite{saenko2010adapting}, this dataset consists of images from 3 domains, Amazon, dSLR and Webcam. Images from Amazon were scraped from the web, while dSLR and Webcam are taken from a high-resolution camera and a webcam, respectively. As a result, there are at least these factors of variation across domains: quality, pose, background, and illumination. These factors induce different shifts in the underlying data, motivating the domain adaptation scenario. There is a grand total of 4652 images in this benchmark, with 2817 on Amazon, 498 in dSLR, and 795 in Webcam. We show the detailed results in Table~\ref{tab:per-domain-office31}. We use ResNets~\citep{he2016deep} (50 variant) as the backbone for extracting embeddings. We finetune the backbone with the available labeled source domain data, then adapt to each target using unlabeled data.

\begin{table}[ht]
    \centering
\resizebox{\linewidth}{!}{
\begin{tabular}{lccccccccc}
    \toprule
    Method & CFM & Dual OT & $\mathcal{X}\times\mathcal{Y}$ & Universal & Amazon & dSLR & Webcam & Avg. & Rank \\
    \midrule
    Source-Only & - & - & - & - & \result{64.23}{0.36} & \result{99.41}{0.51} & \result{96.10}{0.68} & \result{86.58}{0.07} & 10 \\
    \midrule
    Discrete \xmark & - & - & \xmark & \xmark & \result{70.67}{0.40} & \result{92.26}{0.52} & \result{94.35}{0.89} & \result{85.76}{0.56} & 13 \\
    Discrete \cmark & - & - & \cmark & \xmark & \result{69.98}{0.61} & \result{91.67}{1.36} & \result{94.74}{1.01} & \result{85.46}{0.36} & 14 \\
    WGF \xmark & - & - & \xmark & \xmark & \result{69.63}{0.56} & \result{93.75}{0.89} & \result{96.69}{0.34} & \result{86.69}{0.33} & 9 \\
    WGF \cmark & - & - & \cmark & \xmark & \result{70.33}{0.44} & \result{98.51}{1.36} & \result{97.86}{0.68} & \result{88.90}{0.74} & 2 \\
    \midrule
    CW2B & - & - & \xmark & \xmark & \result{69.05}{1.22} & \result{95.83}{2.73} & \result{96.49}{0.58} & \result{87.12}{1.27} & 5 \\
    WIN & - & - & \xmark & \xmark & \result{69.57}{1.92} & \result{93.15}{1.36} & \result{95.91}{1.01} & \result{86.21}{0.53} & 12 \\
    NOT & - & - & \xmark & \xmark & \result{70.03}{0.46} & \result{95.24}{2.25} & \result{96.88}{0.68} & \result{87.38}{0.85} & 3 \\
    U-NOT & - & - & \xmark & \xmark & \result{70.38}{0.63} & \result{93.75}{2.36} & \result{95.13}{0.89} & \result{86.42}{0.67} & 11 \\
    TDSB & - & - & \xmark & \xmark & \result{70.85}{0.10} & \result{94.05}{0.52} & \result{95.32}{0.00} & \result{86.74}{0.19} & 8 \\
    TSBM & - & - & \xmark & \xmark & \result{70.03}{0.46} & \result{94.94}{1.36} & \result{95.91}{1.01} & \result{86.96}{0.04} & 6 \\
    NormFlow & - & - & \xmark & \xmark & \result{70.09}{0.36} & \result{94.05}{1.03} & \result{96.49}{0.58} & \result{86.88}{0.28} & 7 \\
    GMM & - & - & \cmark & \xmark & \result{71.25}{0.91} & \result{95.54}{1.55} & \result{94.74}{0.58} & \result{87.18}{0.99} & 4 \\
    NWB & - & - & \xmark & \xmark & \result{69.57}{0.56} & \result{92.26}{2.06} & \result{93.37}{0.34} & \result{85.07}{0.67} & 15 \\
    NWBU & - & - & \xmark & \cmark & \result{67.83}{1.48} & \result{91.37}{2.25} & \result{94.93}{1.79} & \result{84.71}{0.81} & 16 \\
    \midrule
    NeuralFP & \xmark & \cmark & \cmark & \cmark & \result{68.93}{0.44} & \result{98.81}{0.52} & \result{97.86}{0.34} & \result{88.53}{0.11} & 2 \\
    BaryFM$_{\text{Emp}}$ & \cmark & \xmark & \cmark & \cmark & \result{69.63}{0.53} & \result{98.81}{0.52} & \result{98.83}{0.00} & \result{89.09}{0.33} & 1 \\
    BaryFM$_{\text{unsup}}$ & \cmark & \cmark & \xmark & \cmark & \result{69.05}{0.40} & \result{100.00}{0.00} & \result{98.05}{0.89} & \result{89.03}{0.18} & 1 \\
    BaryFM (no Universal) & \cmark & \cmark & \cmark & \xmark & \result{67.36}{0.36} & \result{98.81}{0.52} & \result{97.47}{0.34} & \result{87.88}{0.27} & 2 \\
    BaryFM & \cmark & \cmark & \cmark & \cmark & \result{70.21}{0.76} & \result{100.00}{0.00} & \result{99.03}{0.89} & \result{89.74}{0.23} & 1 \\
    \bottomrule
\end{tabular}
}
\caption{Per-domain Office 31 results, along average over domains, and rank.}
\label{tab:per-domain-office31}
\end{table}

From Table~\ref{tab:per-domain-office31}, all variants of \texttt{BaryFM} attain either the 1st, or 2nd place among methods, with 2nd best being \texttt{WGF} of~\cite{montesuma2025computing}.  Note that, naturally, the transfer towards dSLR and Webcam are easier, since they have, in the source domains, a closely resembling labeled domain (Webcam for dSLR, and dSLR for Webcam), however, most unlabeled solvers degrade performance in these domains (e.g., \texttt{NWBU} has \result{91.37}{2.25} on dSLR, significantly lower than the source-only baseline at \result{99.41}{0.51}). Our \texttt{BaryFM} is the only method among the compared solvers that manages to improve performance on the most difficult domain (Amazon), while also improving performance on the other domains (dSLR and Webcam).

\newpage

\subsubsection{Office Home}

Proposed by~\cite{venkateswara2017deep}, this benchmark is a visual adaptation benchmark with 4 domains, Art, Clipart, Product, and Real World, with 2427, 4365, 4439 and 4357 images each. Images in this benchmark have a different kind of shift with respect to Office 31. Indeed, the different domains in Office Home account for \emph{stylistic shifts}, that is, a spoon in art is a work of art while in real world, it is a photo of a spoon. This generally induces a stronger shift between domains, as well as some domains (e.g., Clipart) have less separated classes. Our obtained results are shown in Table~\ref{tab:per-domain-office-home}. Like Office 31, we use ResNets~\citep{he2016deep} (101 variant) as the backbone for extracting embeddings. We finetune the backbone with the available labeled source domain data, then adapt to each target using unlabeled data.

\begin{table}[ht]
    \centering
\resizebox{\linewidth}{!}{
\begin{tabular}{lcccccccccc}
    \toprule
    Method & CFM & Dual OT & $\mathcal{X}\times\mathcal{Y}$ & Universal & Art & Clipart & Product & Real World & Avg. & Rank \\
    \midrule
    Source-Only & - & - & - & - & \result{75.55}{0.25} & \result{63.21}{0.15} & \result{83.57}{0.01} & \result{85.40}{0.07} & \result{76.93}{0.09} & 4 \\
    \midrule
    Discrete \xmark & - & - & \xmark & \xmark & \result{69.63}{0.86} & \result{65.29}{0.51} & \result{84.12}{0.64} & \result{84.78}{0.50} & \result{75.95}{0.05} & 9 \\
    Discrete \cmark & - & - & \cmark & \xmark & \result{70.15}{1.28} & \result{65.73}{0.22} & \result{83.28}{0.11} & \result{85.19}{0.39} & \result{76.09}{0.36} & 8 \\
    WGF \xmark & - & - & \xmark & \xmark & \result{75.24}{1.24} & \result{64.76}{0.89} & \result{83.33}{0.23} & \result{85.02}{0.99} & \result{77.09}{0.28} & 3 \\
    WGF \cmark & - & - & \cmark & \xmark & \result{75.45}{0.43} & \result{65.06}{0.11} & \result{85.32}{0.33} & \result{86.28}{0.24} & \result{78.03}{0.13} & 2 \\
    \midrule
    CW2B & - & - & \xmark & \xmark & \result{71.07}{0.82} & \result{64.95}{0.62} & \result{82.81}{1.02} & \result{84.89}{0.63} & \result{75.93}{0.19} & 10 \\
    WIN & - & - & \xmark & \xmark & \result{70.74}{1.19} & \result{65.02}{0.39} & \result{82.62}{0.29} & \result{85.12}{0.74} & \result{75.88}{0.17} & 12 \\
    NOT & - & - & \xmark & \xmark & \result{71.99}{1.38} & \result{64.99}{0.66} & \result{82.81}{0.66} & \result{85.63}{0.49} & \result{76.35}{0.54} & 5 \\
    U-NOT & - & - & \xmark & \xmark & \result{71.40}{0.20} & \result{64.50}{0.30} & \result{83.61}{0.44} & \result{85.34}{0.17} & \result{76.21}{0.12} & 6 \\
    TDSB & - & - & \xmark & \xmark & \result{68.18}{0.30} & \result{63.94}{0.22} & \result{83.10}{0.13} & \result{84.67}{0.17} & \result{74.97}{0.15} & 14 \\
    TSBM & - & - & \xmark & \xmark & \result{71.60}{1.10} & \result{65.02}{0.28} & \result{82.66}{0.33} & \result{85.15}{0.13} & \result{76.11}{0.30} & 7 \\
    NormFlow & - & - & \xmark & \xmark & \result{71.60}{0.20} & \result{64.87}{0.45} & \result{82.48}{0.33} & \result{84.71}{0.45} & \result{75.91}{0.08} & 11 \\
    GMM & - & - & \cmark & \xmark & \result{69.63}{0.90} & \result{63.49}{1.16} & \result{85.25}{0.33} & \result{84.11}{0.68} & \result{75.62}{0.57} & 13 \\
    NWB & - & - & \xmark & \xmark & \result{70.22}{0.79} & \result{59.84}{4.03} & \result{82.51}{1.14} & \result{80.66}{2.71} & \result{73.31}{0.37} & 16 \\
    NWBU & - & - & \xmark & \cmark & \result{67.85}{0.99} & \result{63.31}{0.36} & \result{82.15}{1.05} & \result{83.85}{0.29} & \result{74.29}{0.36} & 15 \\
    \midrule
    NeuralFP & \xmark & \cmark & \cmark & \cmark & \result{76.65}{0.06} & \result{65.84}{0.02} & \result{85.51}{0.17} & \result{86.46}{0.07} & \result{78.61}{0.02} & 1 \\
    BaryFM$_{\text{Emp}}$ & \cmark & \xmark & \cmark & \cmark & \result{77.24}{0.13} & \result{65.79}{0.11} & \result{85.45}{0.00} & \result{86.24}{0.14} & \result{78.68}{0.04} & 1 \\
    BaryFM$_{\text{unsup}}$ & \cmark & \cmark & \xmark & \cmark & \result{76.90}{0.13} & \result{65.76}{0.11} & \result{85.29}{0.10} & \result{86.47}{0.07} & \result{78.60}{0.07} & 1 \\
    BaryFM (no Universal) & \cmark & \cmark & \cmark & \xmark & \result{76.91}{0.29} & \result{66.04}{0.12} & \result{85.39}{0.12} & \result{86.25}{0.11} & \result{78.65}{0.09} & 1 \\
    BaryFM & \cmark & \cmark & \cmark & \cmark & \result{77.12}{0.20} & \result{65.84}{0.29} & \result{85.52}{0.07} & \result{86.38}{0.12} & \result{78.72}{0.12} & 1 \\
    \bottomrule
\end{tabular}
}
\caption{Per-domain Office-Home results, along average over domains, and rank.}
\label{tab:per-domain-office-home}
\end{table}

In this benchmark, \texttt{BaryFM} and all its variants achieve the best performance, being very close to each other (e.g., worst \texttt{BaryFM}$_{\text{unsup}}$ with \result{78.60}{0.07} vs. \texttt{BaryFM} \result{78.72}{0.12}). All of these results cluster well above the 2nd placed method, $\texttt{WGF}$, at \result{78.03}{0.13}. One important point in this comparison, the source-only baseline ranks 4th, surpassing many methods. Indeed, only \texttt{WGF} variants and ours manage to improve over source-only.

\newpage

\subsubsection{Domain Net}

Proposed by~\citep{peng2019moment}, this benchmark is the largest, in number of samples, of the vision adaptation benchmarks. It comprises 6 domains, Clipart, Infograph, Painting, Quickdraw, Real, Sketch, with 48129, 51605, 72266, 172500, 172947, and 69128 images, in a grand total of 586575 images divided across 345 classes. This benchmark pushes the limits of visual adaptation, having a higher diversity of images, classes, and styles. For instance, the most difficult domain, Quickdraw, consists of drawings collected from players of the game \emph{Quick, Draw!} Like other benchmarks, we use ResNets~\citep{he2016deep} (101 variant) as the backbone for extracting embeddings. We finetune the backbone with the available labeled source domain data, then adapt to each target using unlabeled data.

\begin{table}[ht]
    \centering
\resizebox{\linewidth}{!}{
\begin{tabular}{lcccccccccccc}
    \toprule
    Method & CFM & Dual OT & $\mathcal{X}\times\mathcal{Y}$ & Universal & Clipart & Infograph & Painting & Quickdraw & Real & Sketch & Avg. & Rank \\
    \midrule
    Source-Only & - & - & - & - & \result{71.75}{0.02} & \result{28.63}{0.03} & \result{58.73}{0.00} & \result{17.34}{0.01} & \result{71.07}{0.00} & \result{59.86}{0.00} & \result{51.23}{0.01} & 7 \\
    \midrule
    Discrete \xmark & - & - & \xmark & \xmark & \result{69.61}{0.38} & \result{27.33}{0.32} & \result{55.97}{0.32} & \result{21.33}{0.63} & \result{70.84}{0.29} & \result{57.72}{0.36} & \result{50.46}{0.20} & 9 \\
    Discrete \cmark & - & - & \cmark & \xmark & \result{70.81}{0.58} & \result{27.19}{0.23} & \result{56.64}{0.41} & \result{21.28}{0.41} & \result{72.06}{0.17} & \result{58.06}{0.21} & \result{51.01}{0.10} & 8 \\
    WGF \xmark & - & - & \xmark & \xmark & \result{67.88}{0.25} & \result{27.44}{0.06} & \result{55.28}{0.46} & \result{20.70}{0.23} & \result{68.72}{0.04} & \result{56.85}{0.19} & \result{49.48}{0.04} & 12 \\
    WGF \cmark & - & - & \cmark & \xmark & \result{71.28}{0.29} & \result{29.46}{0.42} & \result{58.02}{0.45} & \result{22.41}{0.36} & \result{72.46}{0.19} & \result{58.88}{0.09} & \result{52.08}{0.14} & 5 \\
    \midrule
    CW2B & - & - & \xmark & \xmark & \result{69.01}{1.52} & \result{26.10}{0.36} & \result{55.56}{1.17} & \result{16.98}{0.79} & \result{69.63}{0.35} & \result{57.56}{0.74} & \result{49.14}{0.69} & 13 \\
    WIN & - & - & \xmark & \xmark & \result{66.80}{0.50} & \result{23.76}{0.12} & \result{53.60}{0.63} & \result{14.14}{1.08} & \result{67.67}{0.55} & \result{55.39}{0.93} & \result{46.89}{0.49} & 14 \\
    NOT & - & - & \xmark & \xmark & \result{69.62}{0.36} & \result{26.75}{0.38} & \result{55.71}{0.14} & \result{17.66}{0.57} & \result{70.95}{0.08} & \result{57.77}{0.07} & \result{49.74}{0.07} & 11 \\
    U-NOT & - & - & \xmark & \xmark & \result{69.19}{0.07} & \result{27.43}{0.13} & \result{56.20}{0.09} & \result{18.14}{0.19} & \result{70.28}{0.38} & \result{57.78}{0.27} & \result{49.84}{0.15} & 10 \\
    TDSB & - & - & \xmark & \xmark & \result{71.34}{0.34} & \result{30.64}{0.28} & \result{58.28}{0.14} & \result{23.22}{0.09} & \result{72.04}{0.12} & \result{59.82}{0.13} & \result{52.56}{0.13} & 2 \\
    TSBM & - & - & \xmark & \xmark & \result{71.54}{0.12} & \result{30.11}{0.08} & \result{58.45}{0.16} & \result{23.16}{0.11} & \result{72.14}{0.20} & \result{60.14}{0.10} & \result{52.59}{0.02} & 1 \\
    NormFlow & - & - & \xmark & \xmark & \result{71.06}{0.21} & \result{29.96}{0.19} & \result{58.01}{0.17} & \result{23.09}{0.19} & \result{71.98}{0.13} & \result{59.60}{0.11} & \result{52.28}{0.07} & 4 \\
    GMM & - & - & \cmark & \xmark & \result{70.44}{0.12} & \result{28.05}{0.03} & \result{56.17}{0.05} & \result{24.90}{0.10} & \result{72.98}{0.09} & \result{57.76}{0.10} & \result{51.72}{0.02} & 6 \\
    NWB & - & - & \xmark & \xmark & \result{59.79}{2.90} & \result{20.70}{6.09} & \result{41.58}{11.44} & \result{8.54}{3.53} & \result{62.93}{1.34} & \result{37.28}{12.42} & \result{38.47}{4.23} & 15 \\
    NWBU & - & - & \xmark & \cmark & \result{50.44}{6.13} & \result{18.55}{2.44} & \result{38.02}{10.76} & \result{11.20}{3.05} & \result{55.64}{3.73} & \result{50.61}{1.36} & \result{37.41}{1.82} & 16 \\
    \midrule
    NeuralFP & \xmark & \cmark & \cmark & \cmark & \result{72.16}{0.03} & \result{27.99}{0.21} & \result{57.67}{0.08} & \result{22.83}{0.44} & \result{73.39}{0.12} & \result{59.04}{0.26} & \result{52.18}{0.09} & 4 \\
    BaryFM$_{\text{Emp}}$ & \cmark & \xmark & \cmark & \cmark & \result{72.41}{0.28} & \result{28.69}{0.12} & \result{57.95}{0.18} & \result{23.24}{0.15} & \result{73.23}{0.07} & \result{59.71}{0.11} & \result{52.54}{0.12} & 3 \\
    BaryFM$_{\text{unsup}}$ & \cmark & \cmark & \xmark & \cmark & \result{72.38}{0.04} & \result{29.15}{0.14} & \result{57.94}{0.18} & \result{23.44}{0.07} & \result{73.48}{0.01} & \result{59.84}{0.02} & \result{52.70}{0.04} & 1 \\
    BaryFM (no Universal) & \cmark & \cmark & \cmark & \xmark & \result{72.34}{0.24} & \result{28.66}{0.09} & \result{57.78}{0.15} & \result{23.15}{0.22} & \result{73.22}{0.12} & \result{59.56}{0.10} & \result{52.45}{0.04} & 3 \\
    BaryFM & \cmark & \cmark & \cmark & \cmark & \result{72.36}{0.26} & \result{28.81}{0.23} & \result{57.89}{0.09} & \result{23.33}{0.16} & \result{73.22}{0.09} & \result{59.74}{0.05} & \result{52.56}{0.07} & 3 \\
    \bottomrule
\end{tabular}
}
\caption{Per-domain DomainNet results, along average over domains, and rank.}
\label{tab:per-domain-domain-net}
\end{table}

The difficulty of this benchmark makes the gains overall marginal. For instance, only one empirical solver manages to improve over the source-only baseline, \texttt{WGF}, achieving \result{52.08}{0.14} against \result{51.23}{0.01}. Furthermore, unsupervised solvers perform best: \texttt{TSBM} ranks 1st with \result{52.59}{0.02}, and \texttt{TDSB} ties with our full \texttt{BaryFM} at \result{52.56}{0.13} and \result{52.56}{0.07}, respectively. Consistently with this remark, our \texttt{BaryFM}$_{\text{unsup}}$ variant is the best among our pool of variants (\result{52.70}{0.04}, c.f. Table~\ref{tab:ablation}), which overall suggests an edge of unsupervised barycenter solvers in this benchmark.

\newpage

\subsubsection{BCI-CIV-2a}

This benchmark was proposed by~\cite{brunner2008bci} as the dataset A within the BCI competition IV. This benchmark contains \gls{eeg} recordings, that is, time series of the form $x \in \mathbb{R}^{V \times T}$, of $V$ variates per $T$ time steps. Each \gls{eeg} recording is associated with a \emph{motor imagery task}, namely, imagining a movement. There are 4 movements in this benchmark: left hand, right hand, both feet and tongue. Recordings were acquired with $V=22$ electrodes at 250Hz over 2 sessions on separate days, with 288 trials per session. We refer readers to~\cite{wang2025cbramod} for further details on the preprocessing and overall backbone design.

The main driver of heterogeneity in this benchmark, as with other \gls{eeg} datasets, is \emph{cross-subject variability}, that is, the difference in the measured signals due to the different morphology or state-of-mind of subjects at the moment of data collection. Following~\cite{wang2025cbramod}, we train using data from subjects $1,2,\cdots,5$, use subjects $6, 7$ for checkpoint selection. Subjects 8 and 9 are reserved for test, and are never seen during the fine-tuning of the backbone. We report poled accuracy, which makes Table~\ref{tab:per-domain-bci-iv-2a-crosssub} comparable to the results reported in~\cite{wang2025cbramod}.

\begin{table}[ht]
    \centering
\resizebox{\linewidth}{!}{
\begin{tabular}{lcccccccc}
    \toprule
    Method & CFM & Dual OT & $\mathcal{X}\times\mathcal{Y}$ & Universal & S8 & S9 & Pool. & Rank \\
    \midrule
    Source-Only & - & - & - & - & \result{54.98}{0.10} & \result{50.46}{0.10} & \result{52.72}{0.05} & 16 \\
    \midrule
    Discrete \xmark & - & - & \xmark & \xmark & \result{66.72}{0.44} & \result{58.51}{0.46} & \result{62.62}{0.41} & 8 \\
    Discrete \cmark & - & - & \cmark & \xmark & \result{66.67}{0.76} & \result{58.33}{0.97} & \result{62.50}{0.63} & 10 \\
    WGF \xmark & - & - & \xmark & \xmark & \result{66.90}{0.20} & \result{59.14}{0.66} & \result{63.02}{0.23} & 2 \\
    WGF \cmark & - & - & \cmark & \xmark & \result{66.15}{0.17} & \result{58.97}{0.20} & \result{62.56}{0.13} & 9 \\
    \midrule
    CW2B & - & - & \xmark & \xmark & \result{66.38}{0.27} & \result{59.03}{0.63} & \result{62.70}{0.35} & 4 \\
    WIN & - & - & \xmark & \xmark & \result{65.51}{1.52} & \result{54.40}{5.99} & \result{59.95}{3.69} & 15 \\
    NOT & - & - & \xmark & \xmark & \result{66.61}{0.78} & \result{58.85}{0.97} & \result{62.73}{0.10} & 3 \\
    U-NOT & - & - & \xmark & \xmark & \result{66.38}{0.80} & \result{58.91}{1.18} & \result{62.64}{0.81} & 6 \\
    TDSB & - & - & \xmark & \xmark & \result{66.67}{0.17} & \result{58.62}{0.20} & \result{62.64}{0.18} & 6 \\
    TSBM & - & - & \xmark & \xmark & \result{66.55}{0.36} & \result{58.45}{0.36} & \result{62.50}{0.31} & 10 \\
    NormFlow & - & - & \xmark & \xmark & \result{66.72}{0.10} & \result{58.16}{0.46} & \result{62.44}{0.22} & 12 \\
    GMM & - & - & \cmark & \xmark & \result{65.45}{0.52} & \result{58.45}{0.86} & \result{61.95}{0.52} & 14 \\
    NWB & - & - & \xmark & \xmark & \result{66.49}{0.30} & \result{58.33}{1.04} & \result{62.41}{0.54} & 13 \\
    NWBU & - & - & \xmark & \cmark & \result{66.15}{0.17} & \result{59.26}{1.60} & \result{62.70}{0.88} & 4 \\
    \midrule
    NeuralFP & \xmark & \cmark & \cmark & \cmark & \result{68.52}{0.30} & \result{61.34}{0.16} & \result{64.93}{0.19} & 1 \\
    BaryFM$_{\text{Emp}}$ & \cmark & \xmark & \cmark & \cmark & \result{68.00}{0.53} & \result{60.94}{0.30} & \result{64.47}{0.28} & 1 \\
    BaryFM$_{\text{unsup}}$ & \cmark & \cmark & \xmark & \cmark & \result{67.77}{0.20} & \result{62.67}{0.92} & \result{65.22}{0.53} & 1 \\
    BaryFM (no Universal) & \cmark & \cmark & \cmark & \xmark & \result{68.52}{0.20} & \result{61.05}{0.27} & \result{64.79}{0.18} & 1 \\
    BaryFM & \cmark & \cmark & \cmark & \cmark & \result{68.34}{0.40} & \result{60.71}{0.99} & \result{64.53}{0.65} & 1 \\
    \bottomrule
\end{tabular}
}
\caption{Per-domain BCI-IV-2a cross-subject results, along pooled performance, and rank.}
\label{tab:per-domain-bci-iv-2a-crosssub}
\end{table}

Overall, all methods improve over the baseline. This is natural, since the finetuning of the baseline does not have access to the target domain data, and the domain shift is large. Our \texttt{Baryfm} methods all achieve the best results in this benchmark, followed by \texttt{WGF} (unsupervised). The best results are achieved with \texttt{BaryFM}$_{\text{unsup}}$.

\newpage

\subsubsection{FACED}

\begin{figure}[ht]
    \centering
    \includegraphics[width=\linewidth]{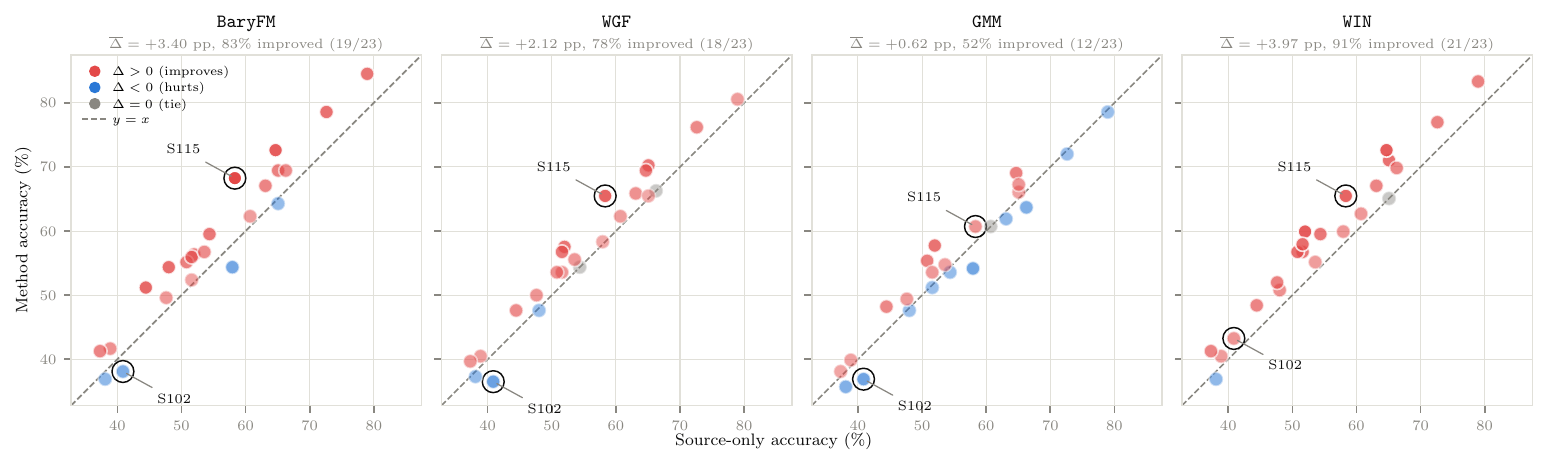}
    \caption{Source-only accuracy vs. method accuracy, for \texttt{BaryFM}$_{\text{Emp}}$, \texttt{BaryFM}, \texttt{WIN} and \texttt{CW2B}, per test subject. The bisecting line defines the no-improvement region. Points under this line (e.g., 119), in blue, show subjects with no improved performance. Subjects on top of that line (e.g., 115) show subjects with improved performance.}
    \label{fig:faced-perf}
\end{figure}

\noindent
\begin{minipage}[t]{0.42\linewidth}
\vspace{0pt}
This benchmark was proposed by~\cite{chen2023large}, and is composed of \glspl{eeg} with $V=32$ electrodes, at 250Hz, collected from 123 subjects, which makes it the benchmark with the most number of domains. Each \gls{eeg} corresponds to one of 9 classes: amusement, inspiration, joy,
tenderness, anger, fear, disgust, sadness and a neutral state in a balanced coverage of positive and negative emotions. We refer readers to~\cite{wang2025cbramod} for the detailed description on the data preprocessing and the backbone for this benchmark. Like these authors, we use subjects 1-80 for training, 81-100 for checkpoint selection, and 101-123 for testing. Table~\ref{tab:per-domain-faced} shows the per subject results alongside the pooled performance of each method.

As with other \gls{eeg} benchmarks (e.g. SEED-VIG and SHU-MI), there is a significant variation in the performance on each individual. In the case of FACED, as discussed in~\cite{chen2023large}, this variation is induced by differences in people's emotional experiences.

In this benchmark, most empirical approaches, except for unsupervised \texttt{WGF}~\citep{montesuma2025computing}, perform poorly. In contrast, neural methods tend to perform better, especially \texttt{WIN}~\citep{korotin2022wasserstein}. In comparison, our \texttt{BaryFM} is a close 2nd competitor (\result{58.25}{0.03} ours vs. \result{58.83}{0.43} theirs), and our empirical variant surpasses \texttt{WIN} (\result{59.27}{0.16} ours vs. \result{58.83}{0.43} theirs). Overall, solvers improve over the source-only baseline for most subjects, except for a few subjects (e.g., 103, 119), see Figure~\ref{fig:faced-perf}.
\end{minipage}\hfill
\begin{minipage}[t]{0.54\linewidth}
    \vspace{0pt}
    \rotatebox{90}{%
    \begin{minipage}{0.95\textheight}
        \centering
\resizebox{\linewidth}{!}{
\begin{tabular}{lccccccccccccccccccccccccccccc}
    \toprule
    Method & CFM & Dual OT & $\mathcal{X}\times\mathcal{Y}$ & Universal & S101 & S102 & S103 & S104 & S105 & S106 & S107 & S108 & S109 & S110 & S111 & S112 & S113 & S114 & S115 & S116 & S117 & S118 & S119 & S120 & S121 & S122 & S123 & Pool. & Rank \\
    \midrule
    Source-Only & - & - & - & - & \result{38.89}{0.69} & \result{40.87}{0.69} & \result{57.94}{0.69} & \result{51.98}{1.37} & \result{37.30}{0.69} & \result{65.08}{0.69} & \result{48.02}{0.69} & \result{66.27}{0.69} & \result{51.59}{0.69} & \result{60.71}{0.00} & \result{50.79}{1.37} & \result{51.59}{0.69} & \result{78.97}{0.69} & \result{63.10}{0.00} & \result{58.33}{0.00} & \result{72.62}{0.00} & \result{64.68}{0.69} & \result{44.44}{0.69} & \result{38.10}{0.00} & \result{54.37}{0.69} & \result{47.62}{0.00} & \result{65.08}{0.69} & \result{53.57}{0.00} & \result{54.87}{0.27} & 14 \\
    \midrule
    Discrete \xmark & - & - & \xmark & \xmark & \result{38.10}{1.19} & \result{36.51}{0.69} & \result{54.37}{3.00} & \result{51.59}{0.69} & \result{39.68}{0.69} & \result{66.67}{2.06} & \result{47.22}{3.00} & \result{65.48}{2.38} & \result{52.38}{3.15} & \result{57.94}{1.37} & \result{50.40}{1.37} & \result{53.57}{2.38} & \result{78.17}{1.37} & \result{59.52}{4.12} & \result{63.49}{0.69} & \result{75.40}{0.69} & \result{66.27}{0.69} & \result{46.43}{2.06} & \result{33.73}{3.83} & \result{54.37}{1.37} & \result{47.22}{1.37} & \result{61.11}{1.37} & \result{53.17}{1.82} & \result{54.47}{0.32} & 15 \\
    Discrete \cmark & - & - & \cmark & \xmark & \result{37.70}{2.48} & \result{38.10}{1.19} & \result{54.37}{2.48} & \result{50.40}{0.69} & \result{40.08}{2.48} & \result{67.46}{0.69} & \result{47.22}{3.64} & \result{66.27}{2.48} & \result{51.98}{2.48} & \result{57.54}{1.37} & \result{50.79}{3.00} & \result{55.16}{2.75} & \result{77.38}{1.19} & \result{59.92}{3.00} & \result{61.51}{1.82} & \result{73.02}{2.75} & \result{66.67}{1.19} & \result{46.03}{2.48} & \result{34.13}{2.48} & \result{53.17}{3.44} & \result{48.41}{0.69} & \result{60.71}{2.06} & \result{53.97}{0.69} & \result{54.43}{0.50} & 16 \\
    WGF \xmark & - & - & \xmark & \xmark & \result{42.06}{1.82} & \result{36.90}{0.00} & \result{59.92}{1.37} & \result{57.54}{1.37} & \result{40.48}{1.19} & \result{71.43}{1.19} & \result{51.98}{1.82} & \result{71.03}{1.37} & \result{53.97}{1.82} & \result{61.90}{0.00} & \result{55.95}{1.19} & \result{57.94}{1.82} & \result{80.56}{1.37} & \result{67.06}{1.37} & \result{64.68}{0.69} & \result{79.37}{1.37} & \result{71.03}{0.69} & \result{46.83}{0.69} & \result{38.49}{1.82} & \result{57.94}{2.75} & \result{48.41}{1.37} & \result{65.08}{0.69} & \result{54.76}{1.19} & \result{58.06}{0.24} & 4 \\
    WGF \cmark & - & - & \cmark & \xmark & \result{40.48}{0.00} & \result{36.51}{0.69} & \result{58.33}{1.19} & \result{57.54}{1.82} & \result{39.68}{1.82} & \result{70.24}{0.00} & \result{47.62}{1.19} & \result{66.27}{0.69} & \result{53.57}{1.19} & \result{62.30}{0.69} & \result{53.57}{2.06} & \result{56.75}{1.37} & \result{80.56}{0.69} & \result{65.87}{1.37} & \result{65.48}{1.19} & \result{76.19}{1.19} & \result{69.44}{0.69} & \result{47.62}{0.00} & \result{37.30}{1.82} & \result{54.37}{0.69} & \result{50.00}{1.19} & \result{65.48}{2.38} & \result{55.56}{0.69} & \result{56.99}{0.24} & 11 \\
    \midrule
    CW2B & - & - & \xmark & \xmark & \result{42.86}{0.00} & \result{38.10}{0.00} & \result{58.73}{1.82} & \result{59.13}{0.69} & \result{37.70}{0.69} & \result{69.05}{0.00} & \result{50.79}{0.69} & \result{69.84}{0.69} & \result{55.16}{0.69} & \result{62.70}{0.69} & \result{53.97}{0.69} & \result{57.54}{1.82} & \result{83.73}{0.69} & \result{66.67}{0.00} & \result{68.25}{0.69} & \result{79.76}{1.19} & \result{71.03}{0.69} & \result{47.62}{0.00} & \result{36.90}{1.19} & \result{55.95}{1.19} & \result{49.60}{0.69} & \result{65.08}{0.69} & \result{55.56}{0.69} & \result{58.07}{0.18} & 3 \\
    WIN & - & - & \xmark & \xmark & \result{40.48}{1.19} & \result{43.25}{2.48} & \result{59.92}{1.37} & \result{59.92}{0.69} & \result{41.27}{1.82} & \result{71.03}{0.69} & \result{50.79}{1.82} & \result{69.84}{0.69} & \result{56.75}{0.69} & \result{62.70}{2.48} & \result{56.75}{1.37} & \result{57.94}{0.69} & \result{83.33}{0.00} & \result{67.06}{1.82} & \result{65.48}{0.00} & \result{76.98}{0.69} & \result{72.62}{1.19} & \result{48.41}{1.37} & \result{36.90}{2.06} & \result{59.52}{0.00} & \result{51.98}{0.69} & \result{65.08}{0.69} & \result{55.16}{1.37} & \result{58.83}{0.43} & 1 \\
    NOT & - & - & \xmark & \xmark & \result{41.27}{0.69} & \result{37.70}{0.69} & \result{58.33}{0.00} & \result{59.92}{0.69} & \result{37.30}{0.69} & \result{70.24}{0.00} & \result{50.79}{0.69} & \result{68.65}{1.37} & \result{55.95}{0.00} & \result{61.51}{0.69} & \result{52.78}{1.37} & \result{58.33}{1.19} & \result{82.14}{0.00} & \result{66.27}{0.69} & \result{68.65}{0.69} & \result{78.57}{1.19} & \result{71.43}{1.19} & \result{46.83}{0.69} & \result{38.10}{0.00} & \result{55.95}{2.38} & \result{50.79}{0.69} & \result{65.48}{1.19} & \result{55.56}{0.69} & \result{57.94}{0.11} & 6 \\
    U-NOT & - & - & \xmark & \xmark & \result{42.06}{1.37} & \result{36.90}{0.00} & \result{56.35}{1.82} & \result{59.13}{0.69} & \result{37.70}{0.69} & \result{69.84}{0.69} & \result{51.19}{0.00} & \result{68.65}{1.82} & \result{55.95}{0.00} & \result{62.30}{0.69} & \result{53.97}{0.69} & \result{58.33}{1.19} & \result{82.54}{0.69} & \result{65.87}{0.69} & \result{67.86}{1.19} & \result{78.97}{0.69} & \result{71.43}{0.00} & \result{46.43}{0.00} & \result{37.70}{0.69} & \result{54.76}{2.38} & \result{50.79}{0.69} & \result{65.87}{1.37} & \result{55.95}{0.00} & \result{57.85}{0.26} & 7 \\
    TDSB & - & - & \xmark & \xmark & \result{40.08}{0.69} & \result{38.49}{3.00} & \result{55.56}{0.69} & \result{57.14}{0.00} & \result{39.68}{1.82} & \result{69.05}{1.19} & \result{51.98}{0.69} & \result{67.06}{0.69} & \result{54.37}{0.69} & \result{62.70}{0.69} & \result{53.97}{0.69} & \result{57.14}{0.00} & \result{82.14}{1.19} & \result{65.08}{1.37} & \result{63.49}{0.69} & \result{75.00}{1.19} & \result{71.43}{0.00} & \result{46.43}{0.00} & \result{36.11}{0.69} & \result{55.56}{0.69} & \result{49.60}{1.82} & \result{64.29}{0.00} & \result{55.56}{0.69} & \result{57.04}{0.09} & 10 \\
    TSBM & - & - & \xmark & \xmark & \result{40.48}{0.00} & \result{41.27}{2.48} & \result{51.98}{1.37} & \result{56.35}{1.82} & \result{37.30}{3.00} & \result{69.44}{2.48} & \result{48.41}{1.82} & \result{65.48}{1.19} & \result{51.59}{0.69} & \result{60.71}{2.38} & \result{51.98}{3.64} & \result{55.95}{0.00} & \result{78.97}{1.82} & \result{62.30}{4.18} & \result{64.29}{1.19} & \result{76.19}{3.15} & \result{70.63}{1.37} & \result{45.24}{2.38} & \result{36.11}{0.69} & \result{56.75}{1.82} & \result{48.81}{2.38} & \result{62.70}{2.75} & \result{54.76}{2.06} & \result{55.99}{0.65} & 12 \\
    NormFlow & - & - & \xmark & \xmark & \result{42.46}{0.69} & \result{36.90}{0.00} & \result{58.73}{0.69} & \result{59.13}{1.82} & \result{37.30}{1.37} & \result{70.63}{0.69} & \result{50.40}{0.69} & \result{68.65}{0.69} & \result{55.95}{0.00} & \result{63.10}{1.19} & \result{54.37}{0.69} & \result{57.94}{0.69} & \result{82.94}{1.37} & \result{66.27}{0.69} & \result{67.06}{0.69} & \result{77.78}{0.69} & \result{71.83}{0.69} & \result{47.62}{0.00} & \result{38.10}{0.00} & \result{54.76}{0.00} & \result{50.40}{0.69} & \result{65.87}{0.69} & \result{55.95}{1.19} & \result{58.01}{0.13} & 5 \\
    GMM & - & - & \cmark & \xmark & \result{39.88}{0.84} & \result{36.90}{1.68} & \result{54.17}{0.84} & \result{57.74}{2.53} & \result{38.10}{1.68} & \result{66.07}{2.53} & \result{47.62}{1.68} & \result{63.69}{2.53} & \result{51.19}{0.00} & \result{60.71}{0.00} & \result{55.36}{0.84} & \result{53.57}{1.68} & \result{78.57}{1.68} & \result{61.90}{0.00} & \result{60.71}{0.00} & \result{72.02}{0.84} & \result{69.05}{0.00} & \result{48.21}{0.84} & \result{35.71}{1.68} & \result{53.57}{1.68} & \result{49.40}{0.84} & \result{67.26}{0.84} & \result{54.76}{0.00} & \result{55.49}{0.07} & 13 \\
    NWB & - & - & \xmark & \xmark & \result{40.08}{0.69} & \result{36.90}{2.38} & \result{58.73}{1.37} & \result{57.54}{0.69} & \result{38.89}{1.82} & \result{69.05}{0.00} & \result{51.19}{0.00} & \result{67.06}{0.69} & \result{54.76}{0.00} & \result{63.49}{0.69} & \result{53.97}{0.69} & \result{57.14}{0.00} & \result{81.75}{0.69} & \result{67.46}{0.69} & \result{64.68}{1.37} & \result{77.78}{0.69} & \result{71.43}{0.00} & \result{46.03}{1.37} & \result{36.90}{0.00} & \result{54.37}{0.69} & \result{50.00}{0.00} & \result{65.08}{0.69} & \result{55.56}{0.69} & \result{57.38}{0.08} & 8 \\
    NWBU & - & - & \xmark & \cmark & \result{40.48}{2.06} & \result{36.11}{1.37} & \result{58.33}{1.19} & \result{55.95}{1.19} & \result{39.29}{1.19} & \result{68.65}{1.82} & \result{50.40}{2.75} & \result{65.87}{0.69} & \result{55.56}{1.37} & \result{62.30}{1.82} & \result{53.97}{0.69} & \result{57.94}{0.69} & \result{81.75}{0.69} & \result{65.48}{0.00} & \result{65.48}{0.00} & \result{75.79}{0.69} & \result{70.63}{0.69} & \result{46.83}{1.82} & \result{36.51}{0.69} & \result{55.56}{2.75} & \result{49.21}{1.37} & \result{64.68}{0.69} & \result{55.56}{0.69} & \result{57.06}{0.21} & 9 \\
    \midrule
    NeuralFP & \xmark & \cmark & \cmark & \cmark & \result{42.06}{0.69} & \result{35.71}{1.19} & \result{51.19}{1.19} & \result{55.56}{1.37} & \result{38.10}{0.00} & \result{68.65}{0.69} & \result{52.38}{0.00} & \result{68.25}{0.69} & \result{51.98}{1.37} & \result{61.90}{0.00} & \result{53.57}{0.00} & \result{56.35}{0.69} & \result{84.92}{1.37} & \result{64.29}{0.00} & \result{64.68}{0.69} & \result{76.98}{1.37} & \result{69.84}{0.69} & \result{47.62}{1.19} & \result{37.30}{0.69} & \result{57.94}{1.82} & \result{48.81}{1.19} & \result{63.49}{1.37} & \result{54.37}{0.69} & \result{56.78}{0.10} & 11 \\
    BaryFM$_{\text{Emp}}$ & \cmark & \xmark & \cmark & \cmark & \result{42.86}{0.00} & \result{40.08}{0.69} & \result{54.76}{0.00} & \result{58.33}{0.00} & \result{40.48}{0.00} & \result{70.63}{0.69} & \result{53.97}{0.69} & \result{68.65}{0.69} & \result{55.56}{0.69} & \result{63.49}{1.37} & \result{56.35}{0.69} & \result{57.14}{0.00} & \result{86.51}{0.69} & \result{66.67}{0.00} & \result{68.25}{0.69} & \result{80.56}{0.69} & \result{72.62}{0.00} & \result{53.17}{0.69} & \result{37.70}{0.69} & \result{61.11}{0.69} & \result{51.19}{0.00} & \result{63.89}{0.69} & \result{59.13}{0.69} & \result{59.27}{0.16} & 1 \\
    BaryFM$_{\text{unsup}}$ & \cmark & \cmark & \xmark & \cmark & \result{41.27}{0.69} & \result{37.30}{1.37} & \result{55.56}{0.69} & \result{56.75}{0.69} & \result{39.29}{1.19} & \result{69.84}{0.69} & \result{52.78}{1.82} & \result{68.65}{1.82} & \result{51.59}{0.69} & \result{61.90}{0.00} & \result{55.95}{1.19} & \result{56.35}{0.69} & \result{84.13}{0.69} & \result{66.67}{0.00} & \result{68.25}{2.48} & \result{79.37}{0.69} & \result{71.83}{1.37} & \result{50.40}{1.37} & \result{37.70}{0.69} & \result{59.13}{0.69} & \result{48.41}{0.69} & \result{64.29}{1.19} & \result{55.56}{0.69} & \result{57.95}{0.39} & 5 \\
    BaryFM (no Universal) & \cmark & \cmark & \cmark & \xmark & \result{42.06}{0.69} & \result{36.90}{1.19} & \result{55.56}{1.82} & \result{56.75}{0.69} & \result{39.68}{0.69} & \result{70.24}{0.00} & \result{53.17}{1.37} & \result{68.25}{1.82} & \result{51.98}{1.37} & \result{61.90}{0.00} & \result{55.16}{0.69} & \result{57.14}{0.00} & \result{84.52}{0.00} & \result{66.27}{1.37} & \result{65.48}{1.19} & \result{78.57}{1.19} & \result{71.43}{1.19} & \result{51.19}{0.00} & \result{36.90}{1.19} & \result{58.73}{0.69} & \result{49.21}{0.69} & \result{63.89}{0.69} & \result{54.76}{1.19} & \result{57.82}{0.23} & 7 \\
    BaryFM & \cmark & \cmark & \cmark & \cmark & \result{41.67}{0.00} & \result{38.49}{0.69} & \result{53.97}{0.69} & \result{57.14}{0.00} & \result{40.48}{1.19} & \result{69.84}{0.69} & \result{53.17}{0.69} & \result{69.84}{0.69} & \result{53.57}{0.00} & \result{61.90}{0.00} & \result{55.56}{0.69} & \result{56.35}{0.69} & \result{84.13}{0.69} & \result{66.67}{0.00} & \result{67.46}{0.69} & \result{79.37}{1.37} & \result{72.22}{0.69} & \result{51.19}{0.00} & \result{36.51}{0.69} & \result{59.52}{0.00} & \result{49.60}{0.69} & \result{64.29}{1.19} & \result{56.75}{1.37} & \result{58.25}{0.03} & 2 \\
    \bottomrule
\end{tabular}
}
        \captionof{table}{Detailed results on the FACED benchmark.}
        \label{tab:per-domain-faced}
    \end{minipage}%
}

\end{minipage}

\newpage

\subsubsection{SEED-VIG}

Introduced in~\cite{zheng2017multimodal}, SEED-VIG is a dataset for measuring vigilance estimation in a simulated driving task. Subjects drive a virtual system on a real car with a large screen while \gls{eeg} is recorded. Labels are continuous, PERCLOS index, that is, the proportion of time during which eyes are closed, measures via eye tracking glasses. Compared to the other tested benchmarks, SEED-VIG is the only regression downstream task, hence we report the Pearson correlation (in \%) between the ground-truth index and the predicted one. \glspl{eeg} in this benchmark are measured with $V=17$ electrodes at 200Hz, with a total of 20355 recordings across 21 subjects. We use subjects 1-13 for training, 14-17 for checkpoint selection, and the last 4 subjects (18, 19, 20, 21) for test. Our results are shown in Table~\ref{tab:per-domain-seedvig}.

\begin{table}[ht]
    \centering
\resizebox{\linewidth}{!}{
\begin{tabular}{lcccccccccc}
    \toprule
    Method & CFM & Dual OT & $\mathcal{X}\times\mathcal{Y}$ & Universal & S18 & S19 & S20 & S21 & Pool. & Rank \\
    \midrule
    Source-Only & - & - & - & - & \result{87.19}{0.56} & \result{7.13}{7.81} & \result{27.26}{9.24} & \result{70.33}{2.16} & \result{44.72}{2.15} & 9 \\
    \midrule
    Discrete \xmark & - & - & \xmark & \xmark & \result{82.93}{0.48} & \result{-4.43}{0.11} & \result{40.98}{0.68} & \result{64.49}{0.29} & \result{34.66}{0.50} & 16 \\
    Discrete \cmark & - & - & \cmark & \xmark & \result{82.39}{0.22} & \result{-8.97}{0.44} & \result{39.70}{0.81} & \result{64.94}{0.85} & \result{34.76}{0.60} & 15 \\
    WGF \xmark & - & - & \xmark & \xmark & \result{78.83}{0.56} & \result{13.64}{1.34} & \result{41.61}{1.42} & \result{70.10}{0.09} & \result{52.29}{0.62} & 3 \\
    WGF \cmark & - & - & \cmark & \xmark & \result{78.09}{1.52} & \result{12.28}{0.14} & \result{38.78}{2.26} & \result{70.89}{1.00} & \result{51.74}{0.44} & 4 \\
    \midrule
    CW2B & - & - & \xmark & \xmark & \result{83.97}{0.48} & \result{-20.12}{1.32} & \result{43.87}{0.50} & \result{72.42}{0.38} & \result{51.06}{0.74} & 5 \\
    WIN & - & - & \xmark & \xmark & \result{81.56}{1.16} & \result{-11.79}{9.07} & \result{44.73}{2.63} & \result{68.11}{3.49} & \result{49.78}{2.47} & 6 \\
    NOT & - & - & \xmark & \xmark & \result{79.62}{2.71} & \result{-15.07}{2.65} & \result{43.98}{5.34} & \result{65.58}{3.03} & \result{42.23}{2.74} & 11 \\
    U-NOT & - & - & \xmark & \xmark & \result{80.12}{0.23} & \result{-21.65}{0.04} & \result{41.10}{0.96} & \result{68.45}{0.49} & \result{43.98}{1.59} & 10 \\
    TDSB & - & - & \xmark & \xmark & \result{77.82}{0.28} & \result{-18.89}{0.54} & \result{40.27}{0.92} & \result{67.31}{0.54} & \result{49.58}{0.87} & 7 \\
    TSBM & - & - & \xmark & \xmark & \result{81.98}{0.59} & \result{-18.29}{0.39} & \result{42.45}{0.80} & \result{70.34}{0.54} & \result{55.03}{0.35} & 2 \\
    NormFlow & - & - & \xmark & \xmark & \result{78.32}{0.52} & \result{-19.47}{0.88} & \result{40.05}{0.25} & \result{66.79}{0.80} & \result{46.13}{1.83} & 8 \\
    GMM & - & - & \cmark & \xmark & \result{84.81}{4.45} & \result{-12.21}{5.66} & \result{55.62}{2.03} & \result{64.52}{2.55} & \result{41.25}{1.87} & 12 \\
    NWB & - & - & \xmark & \xmark & \result{80.51}{1.58} & \result{-17.46}{3.94} & \result{39.21}{9.60} & \result{64.62}{4.97} & \result{39.60}{2.25} & 13 \\
    NWBU & - & - & \xmark & \cmark & \result{80.24}{1.62} & \result{-16.12}{6.67} & \result{38.81}{0.63} & \result{60.60}{7.84} & \result{37.15}{2.92} & 14 \\
    \midrule
    NeuralFP & \xmark & \cmark & \cmark & \cmark & \result{83.42}{2.57} & \result{8.56}{3.26} & \result{38.60}{2.18} & \result{66.76}{2.34} & \result{47.65}{3.33} & 7 \\
    BaryFM$_{\text{Emp}}$ & \cmark & \xmark & \cmark & \cmark & \result{82.85}{0.46} & \result{13.60}{0.64} & \result{39.00}{0.63} & \result{72.26}{0.40} & \result{53.52}{0.49} & 2 \\
    BaryFM$_{\text{unsup}}$ & \cmark & \cmark & \xmark & \cmark & \result{80.44}{1.82} & \result{11.73}{1.52} & \result{46.26}{1.36} & \result{68.76}{3.35} & \result{52.15}{0.65} & 3 \\
    BaryFM (no Universal) & \cmark & \cmark & \cmark & \xmark & \result{83.14}{0.34} & \result{8.76}{0.40} & \result{38.53}{1.62} & \result{68.92}{1.19} & \result{49.99}{0.06} & 5 \\
    BaryFM & \cmark & \cmark & \cmark & \cmark & \result{87.02}{0.61} & \result{11.52}{2.30} & \result{45.94}{1.72} & \result{70.48}{0.96} & \result{57.42}{0.94} & 1 \\
    \bottomrule
\end{tabular}
}
\caption{Per-domain SEED-VIG results, along pooled performance, and rank.}
\label{tab:per-domain-seedvig}
\end{table}

In this benchmark, our \texttt{BaryFM} has a wider performance gap to the 2nd best, \texttt{TSBM} (\result{57.42}{0.94} ours vs. \result{55.03}{0.35}, theirs).

\newpage

\subsubsection{Mumtaz}

This dataset was publicly released by~\cite{mumtaz2016eeg}, and consists of a public record of resting-state \gls{eeg} for the diagnosis of major depressive disorder. Following the description of~\cite{wang2025cbramod}, the \glspl{eeg} in this benchmark consists of $V=19$ electrodes at 256Hz from 34 patients diagnosed with major depressive disorder, and 30 healthy control patients, with a total of 7143 samples, and 10 patients held out for test (5 healthy, and 5 patients with major depressive disorder).

Unlike other benchmarks, Mumtaz subjects contain a single class, which makes the Wasserstein geometry, in general, not well suited for coping with heterogeneity. For instance, computing the barycenter of a healthy patient and a patient with major depressive disorder would lead to fuzzy labels that, by design, do not respect the class structure in the embedding space. To fix that issue, we group source patients into 5 distinct cohorts. We apply the same methodology for each barycenter solver. Likewise, we group all target subjects into a single cohort. We show the grouping in Table~\ref{tab:mumtaz-cohorts}.

\begin{table}[ht]
    \centering
    \small
    \begin{tabular}{@{}lrrrll@{}}
        \toprule
        Domain & \#H & \#MDD & \# Samples & Healthy subjects & MDD subjects \\
        \midrule
        $p_1$ & 5 & 5 & 1135 & H\,21, 24, 25, 27, 28 & MDD\,15, 17, 18, 27, 28 \\
        $p_2$ & 4 & 5 &  990 & H\,13, 16, 19, 23     & MDD\,10, 11, 12, 22, 25 \\
        $p_3$ & 4 & 4 &  906 & H\,1, 17, 22, 26      & MDD\,1, 2, 16, 19 \\
        $p_4$ & 4 & 4 &  941 & H\,2, 12, 14, 15      & MDD\,13, 21, 23, 26 \\
        $p_5$ & 4 & 4 &  919 & H\,10, 11, 18, 20     & MDD\,14, 20, 24, 29 \\
        \midrule
        Sources (total) & 21 & 22 & 4891 & & \\
        Target (pooled) & 5 & 6 & 1211 & H\,5, 6, 7, 8, 9 & MDD\,4, 5, 6, 7, 9, 34 \\
        \bottomrule
    \end{tabular}
    \caption{Grouping of source domain subjects into different subjects.}
    \label{tab:mumtaz-cohorts}
\end{table}

Our results are shown in Table~\ref{tab:per-domain-mumtaz2016}. Overall, our \texttt{BaryFM} obtains the best performance in this benchmark, followed by unsupervised \texttt{WGF}~\citep{montesuma2025computing}, and \texttt{TSBM}~\citep{howard2025schrdinger}.

\begin{table}[ht]
    \centering
\begin{tabular}{lcccccc}
    \toprule
    Method & CFM & Dual OT & $\mathcal{X}\times\mathcal{Y}$ & Universal & Pool. & Rank \\
    \midrule
    Source-Only & - & - & - & - & \result{91.25}{0.36} & 15 \\
    \midrule
    Discrete \xmark & - & - & \xmark & \xmark & \result{91.30}{0.13} & 13 \\
    Discrete \cmark & - & - & \cmark & \xmark & \result{91.30}{0.13} & 13 \\
    WGF \xmark & - & - & \xmark & \xmark & \result{92.79}{0.41} & 2 \\
    WGF \cmark & - & - & \cmark & \xmark & \result{91.77}{0.56} & 7 \\
    \midrule
    CW2B & - & - & \xmark & \xmark & \result{91.88}{0.19} & 4 \\
    WIN & - & - & \xmark & \xmark & \result{91.58}{0.08} & 11 \\
    NOT & - & - & \xmark & \xmark & \result{91.63}{0.13} & 10 \\
    U-NOT & - & - & \xmark & \xmark & \result{91.74}{0.22} & 8 \\
    TDSB & - & - & \xmark & \xmark & \result{91.74}{0.00} & 8 \\
    TSBM & - & - & \xmark & \xmark & \result{92.07}{0.22} & 3 \\
    NormFlow & - & - & \xmark & \xmark & \result{91.80}{0.05} & 5 \\
    GMM & - & - & \cmark & \xmark & \result{91.58}{0.08} & 11 \\
    NWB & - & - & \xmark & \xmark & \result{91.80}{0.17} & 5 \\
    NWBU & - & - & \xmark & \cmark & \result{91.14}{0.75} & 16 \\
    \midrule
    NeuralFP & \xmark & \cmark & \cmark & \cmark & \result{91.94}{0.17} & 3 \\
    BaryFM$_{\text{Emp}}$ & \cmark & \xmark & \cmark & \cmark & \result{93.12}{0.27} & 1 \\
    BaryFM$_{\text{unsup}}$ & \cmark & \cmark & \xmark & \cmark & \result{92.05}{0.48} & 3 \\
    BaryFM (no Universal) & \cmark & \cmark & \cmark & \xmark & \result{92.35}{0.99} & 2 \\
    BaryFM & \cmark & \cmark & \cmark & \cmark & \result{93.12}{0.59} & 1 \\
    \bottomrule
\end{tabular}
\caption{Mumtaz2016 results, along pooled performance, and rank.}
\label{tab:per-domain-mumtaz2016}
\end{table}

\newpage

\subsubsection{Physio}

\begin{figure}[ht]
    \centering
    \includegraphics[width=\linewidth]{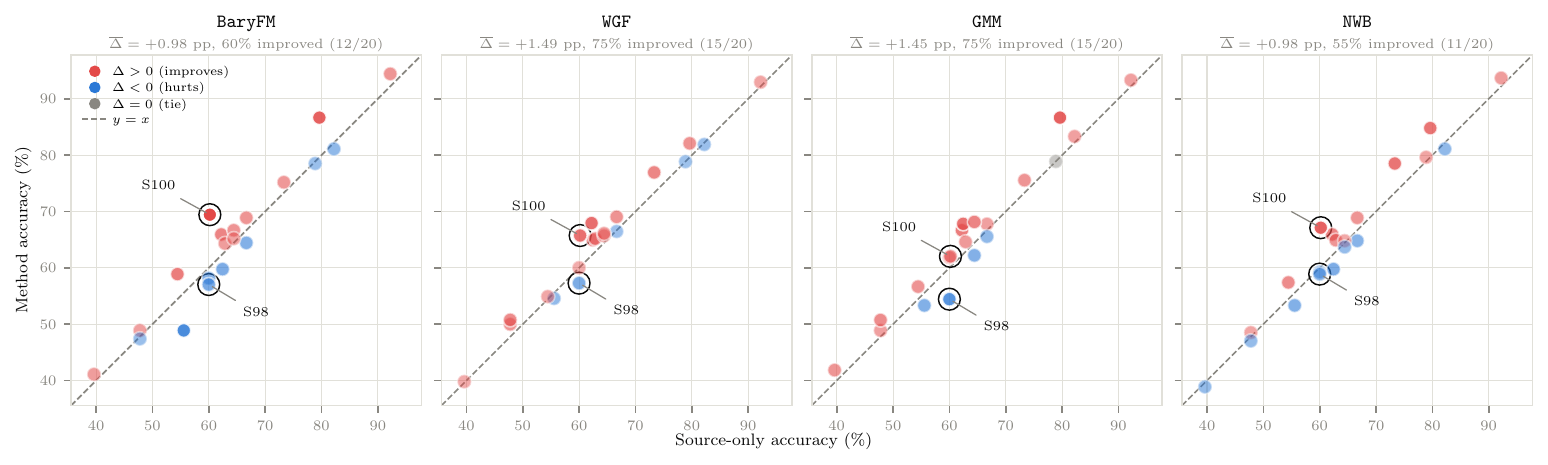}
    \caption{Source-only accuracy vs. method accuracy for \texttt{BaryFM} (5th), \texttt{WGF} (1st), \texttt{GMM} (1st) and \texttt{NWB} (4th). The bisecting line defines the no-improvement region. Points under this line (e.g., S98) denote subjects for which performance is worse than the source-only. Points above it denote subjects for which performance improved over that baseline.}
    \label{fig:physio-perf}
\end{figure}

\noindent
\begin{minipage}[t]{0.42\linewidth}
\vspace{0pt}
Similarly to BCI-CIV-2a, the Physio benchmark, proposed by~\cite{schalk2004bci2000}, is a motor imagery dataset. This dataset constains a collection of 9837 \glspl{eeg}, with $V=64$ electrodes, collected over 109 subjects. Subjects were asked to imagine the movement of the left fist, right fist, both fist and both feet, resulting in a 4-class classification problem. We refer readers to~\cite{wang2025cbramod} for further information on preprocessing.

In this benchmark, most methods have marginal improvement over the baseline (e.g., \texttt{WGF} best with \result{65.54}{0.23} vs. Source-only 16th with \result{64.10}{0.08}, where $\Delta = 1.44$). Our \texttt{BaryFM} achieves \result{65.00}{0.11}, behind a few methods. Similarly to FACED (c.f. Table~\ref{tab:per-domain-faced}), this benchmark also has large inter-subject variability. We provide in Figure~\ref{fig:physio-perf} a per-subject analysis on improvement over the source-only baseline. We see that methods improve performance on most subjects (e.g., 60\% of subjects for \texttt{BaryFM} and \texttt{NWB}, 75\% of subjects for \texttt{WGF} and \texttt{GMM}), but the overall performance improvement remains relatively narrow due to some subjects (e.g., Subject S98) having negative transfer. As mentioned, this seems to be a limitation of barycenter-base adaptation rather than the limitation of any specific method.
\end{minipage}\hfill
\begin{minipage}[t]{0.54\linewidth}
    \vspace{0pt}
    \rotatebox{90}{%
    \begin{minipage}{0.95\textheight}
        \centering
\resizebox{\linewidth}{!}{
\begin{tabular}{lcccccccccccccccccccccccccc}
    \toprule
    Method & CFM & Dual OT & $\mathcal{X}\times\mathcal{Y}$ & Universal & S100 & S101 & S102 & S103 & S104 & S105 & S106 & S107 & S108 & S109 & S90 & S91 & S92 & S93 & S94 & S95 & S96 & S97 & S98 & S99 & Pool. & Rank \\
    \midrule
    Source-Only & - & - & - & - & \result{60.19}{0.80} & \result{92.22}{0.00} & \result{62.22}{0.00} & \result{78.89}{0.00} & \result{62.45}{0.66} & \result{79.63}{0.64} & \result{55.56}{0.00} & \result{47.78}{0.00} & \result{66.67}{0.00} & \result{47.78}{0.00} & \result{73.33}{0.00} & \result{66.67}{0.00} & \result{62.87}{0.51} & \result{54.44}{0.00} & \result{64.44}{0.00} & \result{60.00}{0.00} & \result{82.22}{0.00} & \result{64.44}{0.00} & \result{60.00}{0.00} & \result{39.63}{0.64} & \result{64.10}{0.08} & 16 \\
    \midrule
    Discrete \xmark & - & - & \xmark & \xmark & \result{63.43}{0.80} & \result{92.59}{0.64} & \result{67.41}{1.70} & \result{77.78}{0.00} & \result{64.75}{0.66} & \result{82.96}{0.64} & \result{54.81}{0.64} & \result{48.52}{0.64} & \result{68.52}{0.64} & \result{49.63}{3.21} & \result{77.04}{1.28} & \result{64.81}{0.64} & \result{65.20}{1.34} & \result{53.70}{0.64} & \result{65.56}{0.00} & \result{60.00}{0.00} & \result{82.59}{0.64} & \result{64.07}{1.70} & \result{56.30}{1.70} & \result{38.89}{0.00} & \result{64.95}{0.29} & 10 \\
    Discrete \cmark & - & - & \cmark & \xmark & \result{64.35}{1.60} & \result{92.59}{0.64} & \result{67.78}{2.22} & \result{78.15}{0.64} & \result{65.13}{0.66} & \result{83.33}{0.00} & \result{54.81}{0.64} & \result{48.15}{1.28} & \result{68.52}{0.64} & \result{49.63}{1.70} & \result{75.93}{1.70} & \result{65.56}{1.92} & \result{65.20}{1.34} & \result{53.33}{0.00} & \result{65.56}{0.00} & \result{60.37}{0.64} & \result{81.48}{1.70} & \result{64.81}{2.31} & \result{55.56}{1.92} & \result{38.89}{0.00} & \result{64.97}{0.27} & 7 \\
    WGF \xmark & - & - & \xmark & \xmark & \result{64.35}{2.12} & \result{93.66}{0.00} & \result{63.60}{0.67} & \result{78.10}{1.31} & \result{61.01}{1.19} & \result{83.51}{1.09} & \result{52.44}{0.63} & \result{46.46}{1.89} & \result{68.40}{0.66} & \result{49.64}{1.26} & \result{78.08}{3.42} & \result{64.56}{1.06} & \result{66.36}{0.51} & \result{57.07}{0.64} & \result{65.30}{0.63} & \result{59.26}{1.71} & \result{82.58}{1.06} & \result{66.77}{1.94} & \result{58.06}{1.24} & \result{43.13}{1.96} & \result{65.12}{0.40} & 3 \\
    WGF \cmark & - & - & \cmark & \xmark & \result{65.74}{1.60} & \result{92.96}{0.60} & \result{67.95}{0.63} & \result{78.85}{1.14} & \result{64.94}{2.35} & \result{82.10}{0.69} & \result{54.58}{0.66} & \result{49.99}{1.66} & \result{69.05}{0.66} & \result{50.77}{1.66} & \result{76.94}{1.26} & \result{66.47}{0.66} & \result{65.19}{0.00} & \result{54.91}{0.63} & \result{65.71}{0.00} & \result{60.03}{1.14} & \result{81.90}{0.58} & \result{66.11}{1.67} & \result{57.29}{1.14} & \result{39.80}{0.00} & \result{65.54}{0.23} & 1 \\
    \midrule
    CW2B & - & - & \xmark & \xmark & \result{67.59}{0.80} & \result{94.44}{0.00} & \result{65.19}{0.64} & \result{79.63}{0.64} & \result{59.77}{0.00} & \result{85.19}{0.64} & \result{53.33}{1.11} & \result{48.52}{0.64} & \result{68.89}{0.00} & \result{45.56}{0.00} & \result{77.41}{0.64} & \result{64.81}{0.64} & \result{65.20}{0.51} & \result{56.67}{0.00} & \result{65.56}{0.00} & \result{60.37}{0.64} & \result{81.11}{1.11} & \result{62.96}{0.64} & \result{58.52}{0.64} & \result{38.89}{0.00} & \result{64.97}{0.14} & 7 \\
    WIN & - & - & \xmark & \xmark & \result{67.59}{0.80} & \result{94.44}{0.00} & \result{64.81}{0.64} & \result{78.89}{0.00} & \result{58.62}{0.00} & \result{83.70}{1.28} & \result{54.07}{1.28} & \result{48.52}{1.70} & \result{66.67}{0.00} & \result{47.78}{1.92} & \result{77.41}{0.64} & \result{62.96}{0.64} & \result{64.62}{1.01} & \result{57.04}{0.64} & \result{65.56}{0.00} & \result{60.00}{0.00} & \result{79.26}{0.64} & \result{62.96}{2.31} & \result{58.52}{0.64} & \result{38.89}{0.00} & \result{64.60}{0.03} & 15 \\
    NOT & - & - & \xmark & \xmark & \result{67.59}{0.80} & \result{93.33}{0.00} & \result{65.56}{1.11} & \result{78.89}{1.11} & \result{59.77}{0.00} & \result{86.30}{0.64} & \result{52.96}{0.64} & \result{48.52}{0.64} & \result{68.89}{0.00} & \result{46.67}{0.00} & \result{77.41}{0.64} & \result{64.44}{0.00} & \result{65.20}{0.51} & \result{56.30}{0.64} & \result{65.19}{0.64} & \result{60.00}{0.00} & \result{80.74}{0.64} & \result{64.07}{0.64} & \result{58.89}{0.00} & \result{38.89}{0.00} & \result{64.97}{0.28} & 7 \\
    U-NOT & - & - & \xmark & \xmark & \result{67.13}{0.80} & \result{94.07}{0.64} & \result{64.44}{1.11} & \result{79.63}{0.64} & \result{59.77}{0.00} & \result{85.56}{0.00} & \result{52.96}{0.64} & \result{48.52}{1.28} & \result{68.15}{0.64} & \result{46.30}{0.64} & \result{77.78}{0.00} & \result{64.44}{0.00} & \result{65.20}{0.51} & \result{56.30}{0.64} & \result{65.19}{0.64} & \result{60.00}{0.00} & \result{81.48}{0.64} & \result{62.59}{0.64} & \result{58.52}{0.64} & \result{38.89}{0.00} & \result{64.84}{0.17} & 12 \\
    TDSB & - & - & \xmark & \xmark & \result{66.67}{0.00} & \result{94.44}{0.00} & \result{65.19}{0.64} & \result{78.89}{0.00} & \result{59.77}{0.00} & \result{85.19}{0.64} & \result{52.59}{0.64} & \result{48.15}{0.64} & \result{67.41}{0.64} & \result{46.67}{0.00} & \result{77.78}{1.11} & \result{65.19}{1.28} & \result{65.50}{0.51} & \result{56.67}{0.00} & \result{65.56}{0.00} & \result{59.63}{0.64} & \result{81.85}{0.64} & \result{62.59}{0.64} & \result{58.89}{0.00} & \result{38.89}{0.00} & \result{64.87}{0.08} & 11 \\
    TSBM & - & - & \xmark & \xmark & \result{65.74}{1.60} & \result{94.44}{1.11} & \result{66.67}{1.11} & \result{81.48}{0.64} & \result{58.62}{1.99} & \result{81.85}{1.70} & \result{53.70}{1.28} & \result{48.89}{1.11} & \result{66.67}{1.11} & \result{47.78}{0.00} & \result{77.78}{1.11} & \result{63.33}{1.11} & \result{64.91}{0.00} & \result{56.30}{1.70} & \result{65.56}{1.11} & \result{59.26}{1.28} & \result{80.37}{0.64} & \result{65.19}{1.70} & \result{57.78}{0.00} & \result{39.26}{0.64} & \result{64.78}{0.34} & 13 \\
    NormFlow & - & - & \xmark & \xmark & \result{68.06}{0.00} & \result{94.07}{0.64} & \result{64.44}{1.11} & \result{80.00}{0.00} & \result{59.77}{0.00} & \result{85.56}{0.00} & \result{52.96}{0.64} & \result{47.78}{0.00} & \result{68.15}{0.64} & \result{45.56}{0.00} & \result{77.41}{0.64} & \result{63.70}{0.64} & \result{64.91}{0.00} & \result{56.30}{0.64} & \result{65.56}{0.00} & \result{60.00}{0.00} & \result{80.74}{0.64} & \result{62.59}{0.64} & \result{58.89}{0.00} & \result{38.89}{0.00} & \result{64.74}{0.03} & 14 \\
    GMM & - & - & \cmark & \xmark & \result{62.04}{0.80} & \result{93.33}{0.00} & \result{66.67}{0.00} & \result{78.89}{0.00} & \result{67.82}{0.00} & \result{86.67}{1.11} & \result{53.33}{0.00} & \result{48.89}{0.00} & \result{67.78}{0.00} & \result{50.74}{0.64} & \result{75.56}{0.00} & \result{65.56}{0.00} & \result{64.62}{0.51} & \result{56.67}{0.00} & \result{68.15}{0.64} & \result{61.85}{0.64} & \result{83.33}{0.00} & \result{62.22}{0.00} & \result{54.44}{0.00} & \result{41.85}{0.64} & \result{65.54}{0.08} & 1 \\
    NWB & - & - & \xmark & \xmark & \result{67.13}{0.80} & \result{93.70}{0.64} & \result{65.93}{0.64} & \result{79.63}{0.64} & \result{59.77}{0.00} & \result{84.81}{0.64} & \result{53.33}{0.00} & \result{48.52}{0.64} & \result{68.89}{0.00} & \result{47.04}{0.64} & \result{78.52}{0.64} & \result{64.81}{0.64} & \result{64.91}{0.00} & \result{57.41}{0.64} & \result{64.81}{0.64} & \result{59.26}{0.64} & \result{81.11}{1.11} & \result{63.70}{1.28} & \result{58.89}{0.00} & \result{38.89}{0.00} & \result{65.04}{0.19} & 4 \\
    NWBU & - & - & \xmark & \cmark & \result{67.13}{0.80} & \result{94.07}{0.64} & \result{65.19}{0.64} & \result{79.26}{1.28} & \result{59.77}{0.00} & \result{85.19}{0.64} & \result{53.33}{0.00} & \result{48.15}{0.64} & \result{68.52}{0.64} & \result{47.41}{0.64} & \result{78.15}{0.64} & \result{65.19}{0.64} & \result{65.50}{0.51} & \result{56.67}{1.11} & \result{65.19}{0.64} & \result{59.63}{0.64} & \result{81.11}{0.00} & \result{63.70}{1.28} & \result{58.15}{0.64} & \result{38.89}{0.00} & \result{65.00}{0.06} & 6 \\
    \midrule
    NeuralFP & \xmark & \cmark & \cmark & \cmark & \result{68.06}{0.00} & \result{96.44}{0.49} & \result{66.60}{0.00} & \result{78.85}{0.00} & \result{60.66}{1.05} & \result{86.86}{0.91} & \result{49.85}{0.00} & \result{46.90}{0.49} & \result{68.41}{0.51} & \result{48.90}{2.39} & \result{72.79}{1.07} & \result{64.79}{0.51} & \result{65.40}{0.81} & \result{57.79}{0.89} & \result{66.75}{0.00} & \result{57.38}{1.07} & \result{81.67}{0.56} & \result{66.70}{0.00} & \result{56.63}{0.00} & \result{42.02}{0.00} & \result{65.15}{0.09} & 3 \\
    BaryFM$_{\text{Emp}}$ & \cmark & \xmark & \cmark & \cmark & \result{70.83}{1.39} & \result{95.19}{0.64} & \result{65.19}{0.64} & \result{80.74}{0.64} & \result{59.00}{0.66} & \result{86.67}{0.00} & \result{48.52}{0.64} & \result{49.26}{0.64} & \result{67.78}{0.00} & \result{48.89}{1.92} & \result{77.41}{0.64} & \result{65.56}{0.00} & \result{64.04}{0.00} & \result{56.67}{0.00} & \result{66.30}{0.64} & \result{60.00}{0.00} & \result{78.15}{0.64} & \result{64.07}{0.64} & \result{57.78}{0.00} & \result{38.89}{0.00} & \result{64.98}{0.21} & 6 \\
    BaryFM$_{\text{unsup}}$ & \cmark & \cmark & \xmark & \cmark & \result{69.91}{0.80} & \result{94.81}{0.64} & \result{65.93}{1.28} & \result{78.15}{0.64} & \result{59.39}{1.33} & \result{86.67}{0.00} & \result{49.26}{0.64} & \result{48.89}{0.00} & \result{68.89}{0.00} & \result{47.04}{0.64} & \result{75.19}{0.64} & \result{65.19}{0.64} & \result{64.62}{0.51} & \result{58.89}{0.00} & \result{66.67}{0.00} & \result{58.15}{0.64} & \result{81.11}{0.00} & \result{64.44}{0.00} & \result{56.67}{0.00} & \result{41.11}{0.00} & \result{65.00}{0.17} & 5 \\
    BaryFM (no Universal) & \cmark & \cmark & \cmark & \xmark & \result{68.52}{1.60} & \result{94.81}{0.64} & \result{66.30}{0.64} & \result{78.15}{0.64} & \result{59.39}{0.66} & \result{86.67}{0.00} & \result{48.52}{0.64} & \result{48.89}{0.00} & \result{68.89}{0.00} & \result{47.41}{0.64} & \result{74.44}{0.00} & \result{64.44}{1.11} & \result{65.50}{0.51} & \result{58.52}{0.64} & \result{66.30}{0.64} & \result{57.78}{0.00} & \result{81.11}{0.00} & \result{65.19}{0.64} & \result{56.67}{0.00} & \result{40.74}{0.64} & \result{64.89}{0.17} & 10 \\
    BaryFM & \cmark & \cmark & \cmark & \cmark & \result{69.44}{0.00} & \result{94.44}{0.00} & \result{65.93}{0.64} & \result{78.52}{0.64} & \result{59.77}{0.00} & \result{86.67}{0.00} & \result{48.89}{0.00} & \result{48.89}{0.00} & \result{68.89}{0.00} & \result{47.41}{0.64} & \result{75.19}{1.28} & \result{64.44}{1.11} & \result{64.33}{0.51} & \result{58.89}{0.00} & \result{66.67}{0.00} & \result{58.15}{0.64} & \result{81.11}{0.00} & \result{65.19}{0.64} & \result{57.04}{0.64} & \result{41.11}{0.00} & \result{65.00}{0.11} & 5 \\
    \bottomrule
\end{tabular}
}
        \captionof{table}{Detailed results on the Physio benchmark.}
        \label{tab:per-domain-physio}
    \end{minipage}%
}

\end{minipage}

\newpage

\subsubsection{SHU-MI}

SHU-MI is a benchmark proposed by~\cite{ma2022large} concerned with motor imagery (like Physio and BCI-CIV-2a), with 2 classes, where subjects were asked to imagine left and right hand grasping. The \glspl{eeg} in this benchmark were collected with $V=32$ electrodes, and 25 subjects over 5 separate sessions, with 100 trials per session. There are, in total, 11988 samples. Subjects are divided into $1,\cdots,15$ used for training, $16,\cdots,20$ for checkpoint selection, and $21,\cdots,25$ for test. We refer readers to~\cite{wang2025cbramod} for further details on pre-processing. We report our results in Table~\ref{tab:per-domain-shumi}.

\begin{table}[ht]
    \centering
\resizebox{\linewidth}{!}{
\begin{tabular}{lccccccccccc}
    \toprule
    Method & CFM & Dual OT & $\mathcal{X}\times\mathcal{Y}$ & Universal & S21 & S22 & S23 & S24 & S25 & Pool. & Rank \\
    \midrule
    Source-Only & - & - & - & - & \result{76.97}{0.21} & \result{67.50}{0.42} & \result{67.97}{1.46} & \result{52.86}{1.43} & \result{46.46}{1.19} & \result{62.68}{0.41} & 14 \\
    \midrule
    Discrete \xmark & - & - & \xmark & \xmark & \result{57.95}{6.86} & \result{61.53}{3.34} & \result{61.52}{7.94} & \result{48.36}{3.81} & \result{49.06}{2.61} & \result{55.89}{3.26} & 15 \\
    Discrete \cmark & - & - & \cmark & \xmark & \result{53.32}{4.50} & \result{59.10}{4.90} & \result{58.57}{10.66} & \result{49.05}{3.50} & \result{48.27}{1.63} & \result{53.80}{4.03} & 16 \\
    WGF \xmark & - & - & \xmark & \xmark & \result{76.69}{1.04} & \result{68.12}{2.29} & \result{67.35}{0.63} & \result{53.47}{0.95} & \result{47.19}{1.15} & \result{62.87}{0.90} & 13 \\
    WGF \cmark & - & - & \cmark & \xmark & \result{78.08}{0.24} & \result{68.75}{0.62} & \result{69.14}{0.00} & \result{53.70}{0.95} & \result{47.19}{0.57} & \result{63.70}{0.11} & 1 \\
    \midrule
    CW2B & - & - & \xmark & \xmark & \result{76.90}{0.73} & \result{67.43}{0.24} & \result{69.07}{1.32} & \result{53.32}{0.79} & \result{46.90}{0.98} & \result{63.05}{0.20} & 7 \\
    WIN & - & - & \xmark & \xmark & \result{76.90}{0.24} & \result{67.01}{0.12} & \result{68.52}{0.21} & \result{53.85}{1.15} & \result{46.97}{1.32} & \result{62.96}{0.32} & 10 \\
    NOT & - & - & \xmark & \xmark & \result{76.35}{0.21} & \result{67.71}{1.65} & \result{69.07}{0.86} & \result{53.55}{1.21} & \result{46.68}{0.12} & \result{62.99}{0.52} & 9 \\
    U-NOT & - & - & \xmark & \xmark & \result{76.63}{0.48} & \result{67.57}{1.22} & \result{69.27}{0.83} & \result{52.71}{1.06} & \result{46.75}{0.37} & \result{62.92}{0.21} & 12 \\
    TDSB & - & - & \xmark & \xmark & \result{76.14}{0.21} & \result{67.92}{0.55} & \result{68.66}{1.34} & \result{53.62}{0.87} & \result{47.84}{0.37} & \result{63.14}{0.21} & 4 \\
    TSBM & - & - & \xmark & \xmark & \result{76.69}{0.84} & \result{67.36}{1.42} & \result{69.14}{1.25} & \result{53.09}{1.27} & \result{47.40}{1.21} & \result{63.06}{0.70} & 6 \\
    NormFlow & - & - & \xmark & \xmark & \result{76.35}{0.83} & \result{67.64}{0.98} & \result{68.79}{0.43} & \result{53.32}{0.83} & \result{47.04}{1.44} & \result{62.95}{0.26} & 11 \\
    GMM & - & - & \cmark & \xmark & \result{78.01}{1.30} & \result{68.54}{0.00} & \result{69.41}{0.52} & \result{53.55}{0.40} & \result{46.25}{1.54} & \result{63.49}{0.46} & 3 \\
    NWB & - & - & \xmark & \xmark & \result{76.63}{1.22} & \result{67.99}{0.32} & \result{69.14}{0.82} & \result{53.39}{1.08} & \result{46.97}{0.94} & \result{63.14}{0.60} & 4 \\
    NWBU & - & - & \xmark & \cmark & \result{77.04}{0.84} & \result{67.29}{0.72} & \result{68.79}{0.12} & \result{53.39}{0.58} & \result{47.11}{0.54} & \result{63.05}{0.12} & 7 \\
    \midrule
    NeuralFP & \xmark & \cmark & \cmark & \cmark & \result{78.15}{0.84} & \result{68.54}{0.42} & \result{68.86}{0.43} & \result{54.31}{1.17} & \result{46.68}{0.45} & \result{63.63}{0.49} & 2 \\
    BaryFM$_{\text{Emp}}$ & \cmark & \xmark & \cmark & \cmark & \result{77.59}{0.36} & \result{68.33}{0.72} & \result{68.45}{1.34} & \result{54.08}{0.35} & \result{48.05}{1.32} & \result{63.61}{0.66} & 2 \\
    BaryFM$_{\text{unsup}}$ & \cmark & \cmark & \xmark & \cmark & \result{78.56}{1.14} & \result{66.67}{0.21} & \result{69.07}{1.32} & \result{54.54}{1.40} & \result{47.69}{0.54} & \result{63.61}{0.30} & 2 \\
    BaryFM (no Universal) & \cmark & \cmark & \cmark & \xmark & \result{78.84}{0.75} & \result{66.81}{0.12} & \result{69.55}{0.62} & \result{53.47}{0.48} & \result{47.04}{0.33} & \result{63.47}{0.28} & 3 \\
    BaryFM & \cmark & \cmark & \cmark & \cmark & \result{78.01}{0.21} & \result{68.96}{0.72} & \result{69.00}{0.12} & \result{53.39}{0.26} & \result{47.47}{1.09} & \result{63.70}{0.15} & 2 \\
    \bottomrule
\end{tabular}
}
\caption{Per-domain SHU-MI results, along pooled performance, and rank.}
\label{tab:per-domain-shumi}
\end{table}

In this benchmark, most empirical solvers fail (e.g., the discrete solver achieves \result{55.89}{3.26} vs. the source-only baseline at \result{62.68}{0.41}). One exception is \texttt{WGF}, who achieves the best performance of \result{63.70}{0.11}. Conversely, all neural solvers improve over the source-only baseline. Our \texttt{BaryFM} achieves 2nd place, being statistically indistinguishable from \texttt{WGF} (\result{63.70}{0.15} ours, \result{63.70}{0.11} theirs).

\newpage

\subsubsection{Tennessee Eastman Process}\label{sec:tep}

The Tennessee Eastman Process (TEP) is a large scale, widely used process simulator in the chemical engineering community, first introduced by~\cite{downs1993plant}. Here, we use its domain adaptation instantiation introduced by~\cite{montesuma2024benchmarking}, based on the simulations of~\cite{reinartz2021extended}. Essentially, it consists on multi-variate time series with $V=34$ variates of different physical quantities (e.g., pressure, temperature, concentration). For each simulation, there is one associated label. These labels describe the state of the system (one of 28 possible faults, plus a healthy state). The goal is \emph{fault diagnosis}, meaning, from the time-series, to determine the state of the system. This benchmark has 17289 samples, more or less uniformly distributed among 6 modes of operation. We refer readers to~\cite{montesuma2024benchmarking} for further information on this benchmark.

\begin{table}[ht]
    \centering
\resizebox{\linewidth}{!}{
\begin{tabular}{lcccccccccccc}
    \toprule
    Method & CFM & Dual OT & $\mathcal{X}\times\mathcal{Y}$ & Universal & Mode 1 & Mode 2 & Mode 3 & Mode 4 & Mode 5 & Mode 6 & Avg. & Rank \\
    \midrule
    Source-Only & - & - & - & - & \result{82.40}{0.07} & \result{67.03}{1.03} & \result{88.52}{0.07} & \result{78.30}{0.15} & \result{73.51}{0.13} & \result{85.34}{0.07} & \result{79.18}{0.18} & 16 \\
    \midrule
    Discrete \xmark & - & - & \xmark & \xmark & \result{89.66}{0.75} & \result{73.53}{1.35} & \result{84.96}{1.95} & \result{88.34}{0.75} & \result{84.91}{0.89} & \result{85.81}{2.34} & \result{84.54}{1.47} & 6 \\
    Discrete \cmark & - & - & \cmark & \xmark & \result{89.97}{1.25} & \result{73.11}{1.54} & \result{88.48}{1.76} & \result{88.52}{0.83} & \result{84.46}{0.84} & \result{86.85}{1.07} & \result{85.23}{1.26} & 5 \\
    WGF \xmark & - & - & \xmark & \xmark & \result{91.52}{1.32} & \result{73.78}{1.25} & \result{87.44}{1.17} & \result{89.39}{0.62} & \result{84.88}{1.76} & \result{86.09}{2.23} & \result{85.52}{1.48} & 3 \\
    WGF \cmark & - & - & \cmark & \xmark & \result{92.34}{1.05} & \result{76.10}{1.57} & \result{89.79}{0.90} & \result{89.32}{1.04} & \result{85.92}{1.36} & \result{87.75}{1.52} & \result{86.87}{1.27} & 2 \\
    \midrule
    CW2B & - & - & \xmark & \xmark & \result{88.76}{1.51} & \result{68.37}{0.85} & \result{85.34}{1.48} & \result{88.17}{1.00} & \result{81.44}{2.71} & \result{85.92}{0.95} & \result{83.00}{1.55} & 13 \\
    WIN & - & - & \xmark & \xmark & \result{85.10}{1.50} & \result{68.05}{1.59} & \result{89.27}{1.44} & \result{86.84}{2.81} & \result{79.29}{2.10} & \result{86.92}{1.91} & \result{82.58}{1.95} & 14 \\
    NOT & - & - & \xmark & \xmark & \result{90.28}{2.52} & \result{70.44}{2.95} & \result{85.27}{1.79} & \result{88.59}{1.65} & \result{83.07}{1.83} & \result{82.43}{1.45} & \result{83.35}{2.10} & 12 \\
    U-NOT & - & - & \xmark & \xmark & \result{89.07}{1.49} & \result{71.88}{1.95} & \result{85.34}{1.80} & \result{87.85}{0.83} & \result{83.90}{2.43} & \result{82.64}{2.50} & \result{83.45}{1.92} & 10 \\
    TDSB & - & - & \xmark & \xmark & \result{88.28}{2.12} & \result{70.19}{0.94} & \result{87.48}{1.71} & \result{87.75}{1.18} & \result{83.70}{1.43} & \result{85.16}{1.31} & \result{83.76}{1.50} & 9 \\
    TSBM & - & - & \xmark & \xmark & \result{90.41}{1.45} & \result{69.81}{0.86} & \result{88.48}{1.18} & \result{87.33}{0.75} & \result{84.01}{1.33} & \result{86.47}{1.64} & \result{84.42}{1.24} & 7 \\
    NormFlow & - & - & \xmark & \xmark & \result{88.62}{0.81} & \result{69.77}{1.80} & \result{87.89}{1.42} & \result{87.96}{1.13} & \result{82.86}{1.53} & \result{86.43}{0.83} & \result{83.92}{1.30} & 8 \\
    GMM & - & - & \cmark & \xmark & \result{89.33}{0.48} & \result{72.89}{1.49} & \result{88.81}{0.28} & \result{88.34}{0.72} & \result{85.53}{0.79} & \result{87.14}{0.65} & \result{85.34}{0.63} & 4 \\
    NWB & - & - & \xmark & \xmark & \result{87.62}{0.95} & \result{67.45}{1.34} & \result{88.48}{1.18} & \result{88.17}{0.86} & \result{80.89}{2.90} & \result{87.57}{1.77} & \result{83.36}{1.65} & 11 \\
    NWBU & - & - & \xmark & \cmark & \result{88.79}{0.49} & \result{58.21}{2.02} & \result{84.37}{1.66} & \result{85.79}{1.99} & \result{84.42}{1.35} & \result{81.84}{0.85} & \result{80.57}{1.50} & 15 \\
    \midrule
    NeuralFP & \xmark & \cmark & \cmark & \cmark & \result{92.38}{1.06} & \result{76.56}{1.79} & \result{90.10}{0.98} & \result{89.56}{1.02} & \result{85.92}{1.31} & \result{87.88}{1.56} & \result{87.07}{1.32} & 1 \\
    BaryFM$_{\text{Emp}}$ & \cmark & \xmark & \cmark & \cmark & \result{92.41}{1.03} & \result{76.91}{1.44} & \result{90.13}{0.92} & \result{89.56}{0.74} & \result{86.40}{1.32} & \result{88.12}{1.28} & \result{87.26}{1.15} & 1 \\
    BaryFM$_{\text{unsup}}$ & \cmark & \cmark & \xmark & \cmark & \result{92.45}{0.91} & \result{77.08}{1.67} & \result{90.31}{0.70} & \result{89.91}{0.66} & \result{87.10}{1.52} & \result{87.09}{1.31} & \result{87.32}{1.20} & 1 \\
    BaryFM (no Universal) & \cmark & \cmark & \cmark & \xmark & \result{92.34}{1.03} & \result{76.77}{1.48} & \result{89.82}{0.91} & \result{89.63}{0.87} & \result{86.19}{1.53} & \result{88.06}{1.33} & \result{87.14}{1.22} & 1 \\
    BaryFM & \cmark & \cmark & \cmark & \cmark & \result{92.69}{0.92} & \result{76.77}{2.14} & \result{89.96}{1.04} & \result{89.77}{0.82} & \result{86.82}{1.50} & \result{87.81}{1.48} & \result{87.30}{1.39} & 1 \\
    \bottomrule
\end{tabular}
}
\caption{Per-mode TEP results, along average over modes, and rank.}
\label{tab:per-domain-tep}
\end{table}

We adopt the reporting style of~\cite{montesuma2024benchmarking}, who pre-defined 5 folds on each domain. Therefore, we fix seed 42, and run each method using the embeddings of all source domain data, with an additional of 80\% of target domain points (folds $f_{1},\cdots,f_{4}$) with fold $f_{5} \in \{1, \cdots, 5\}$ kept for testing. The testing data is never seen neither through the training of the backbone, nor through the adaptation stage. We report the mean over folds, $\pm$ the standard deviation, in Table~\ref{tab:per-domain-tep}.

From Table~\ref{tab:per-domain-tep}, our \texttt{BaryFM} performs best in all variants, against \texttt{WGF} (2nd best). For instance, our best performing variant, \texttt{BaryFM}$_{\text{unsup}}$ achieves \result{87.32}{1.15} against \texttt{WGF}'s \result{86.87}{1.37}. Furthermore, all methods achieve a statistically significant improvement over the source-only baseline.

\newpage

\newpage
\section{Domain Generalization}\label{sec:additional-dg}

\subsection{Data Augmentation for Domain Generalization}

Our goal is training a model on measures from the Wasserstein simplex, $\mathcal{W}(p_{1:K})$. The underlying idea is that, despite the vertices $p_{1},\cdots,p_{K}$, the barycenters $\text{Bar}(\lambda, p_{1:K})$ will help enrich the source domain data, improving generalization. We therefore compose an augmented sample pool $\{ \{ \tilde{x}_{\ell,i}, \tilde{y}_{\ell,i} \}_{i=1}^{m} \}_{\ell=1}^{N}$, where $(\tilde{x}_{\ell, i}, \tilde{y}_{\ell, i}) \sim q_{\lambda_{\ell}}$ and $\lambda_{\ell} \sim \text{Dir}(\alpha)$. We then learn a model minimizing the risk,
\begin{align*}
    \theta^{\star} = \argmin{\theta \in \Theta}\dfrac{1-\rho}{n}\sum_{i=1}^{n}\ell(y_{k,i}, f_{\theta}(x_{k,i})) + \dfrac{\rho}{mN}\sum_{\ell=1}^{N}\sum_{i=1}^{m}\ell(\tilde{y}_{\ell,i}, f_{\theta}(\tilde{x}_{\ell,i})),
\end{align*}
where $f_{\theta}$ is a neural network, such as a ResNet 50~\citep{he2016deep}, $\theta$ are its weights, and $\rho \in [0, 1]$ is a importance factor. Note that this application requires 3 properties of barycenter solvers: (i) amortization over $z \sim \text{Bar}(\lambda, p_{1:K})$, (ii) \emph{universality}, i.e., amortization over $\lambda \in \Delta_{K}$ and (iii) joint modeling over $z = (x, y)$, all of which, \emph{to the best of our knowledge}, only our \texttt{BaryFM} suffices. We call this strategy BaryFM$_{\text{Random}}$. Alternatively, we build the augmented sampling pool with $N = K$ and $\lambda_{\ell} = e_\ell, \ell=1,\cdots,K$ being the vertices of the simplex. This roughly corresponds to sampling from the marginal measures.

As the main point of comparison, we adopt MixUp~\citep{zhang2017mixup} and Domain MixUp~\citep{wang2020heterogeneous} which correspond to interpolating $p_{1:K}$ linearly. First, MixUp corresponds to sampling $\{ (x_1, y_1), (x_2, y_2) \} \sim \sum_k (\nicefrac{n_k}{n})p_k \otimes \sum_k (\nicefrac{n_k}{n})p_k$, where $n_k$ is the number of samples from measure $k$, and $\lambda \sim \text{Beta}(\alpha, \alpha)$ then generating $\tilde{x} = (1-\lambda)x_1 + \lambda x_2$, $\tilde{y} = (1-\lambda)y_1 + \lambda y_2$. In contrast, Domain MixUp samples $(x_{1}, y_{1}),\cdots, (x_{K}, y_{K}) \sim p_{1}\otimes\cdots\otimes p_{K}$ then mixes using,
\begin{align}
    \tilde{x} = \sum_{k=1}^{K}\lambda_{k}x_{k}\text{, and }\tilde{y} = \sum_{k=1}^{K}\lambda_{k}y_{k}.\label{eq:domain-mixup}
\end{align}
In practice, for better capturing the semantics of complex objects, we perform interpolation in the latent space of an autoencoder using either version of BaryFM or MixUp. We use a \gls{vae}~\citep{kingma2013auto,rombach2022high} and \gls{pca} for EEG data.

\begin{figure}[ht]
    \centering
    \includegraphics[width=\linewidth]{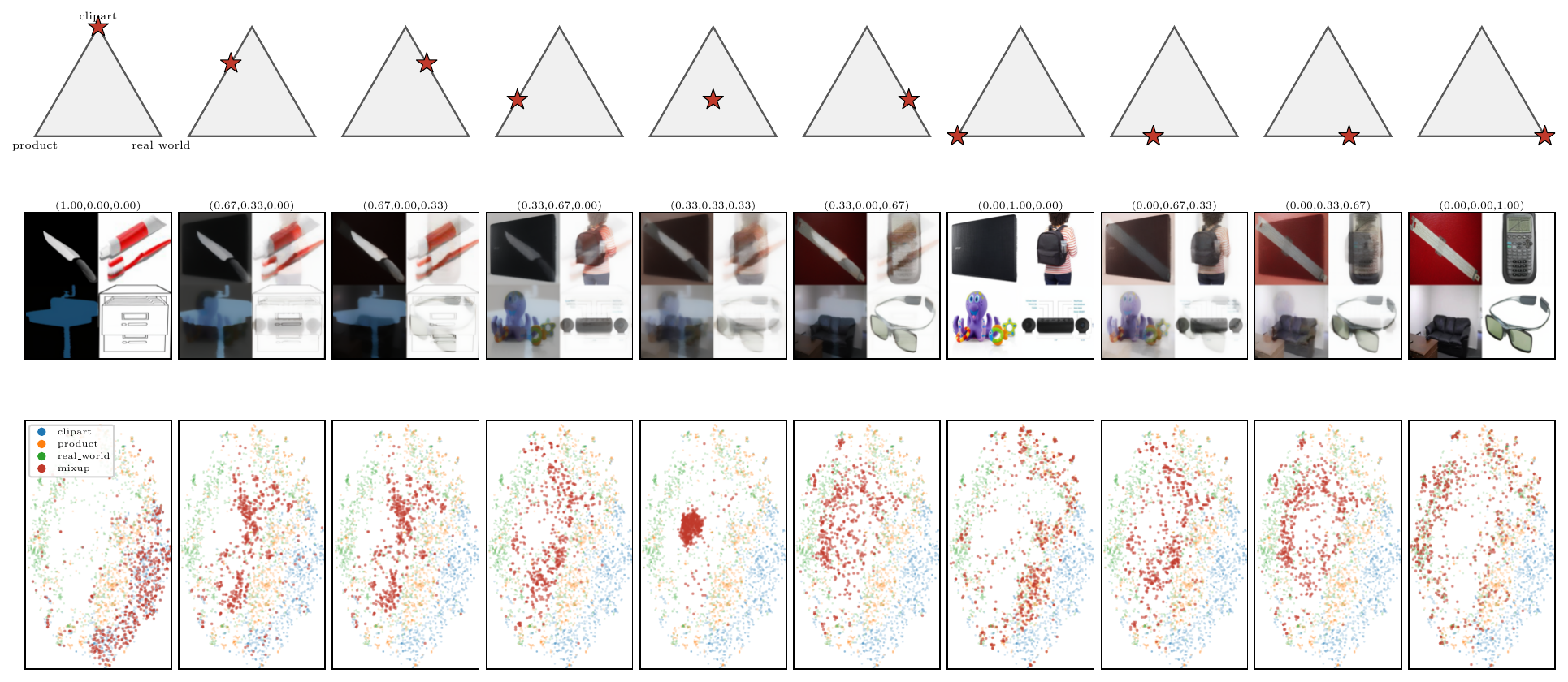}
    \caption{Visualization of the linear mixing of multiple domain data via Domain MixUp of~\cite{wang2020heterogeneous}. In the first row, we show the simplex, and the picked mixing weights $\lambda$. In the second row, we show the mixed samples obtained by equation~\ref{eq:domain-mixup}, and decoded back into image space. In the third row, we show the t-SNE visualization~\citep{maaten2008visualizing} of the \gls{vae} embeddings of the mixed samples.}
    \label{fig:domain-mixup-dg}
\end{figure}

\begin{figure}[ht]
    \centering
    \includegraphics[width=\linewidth]{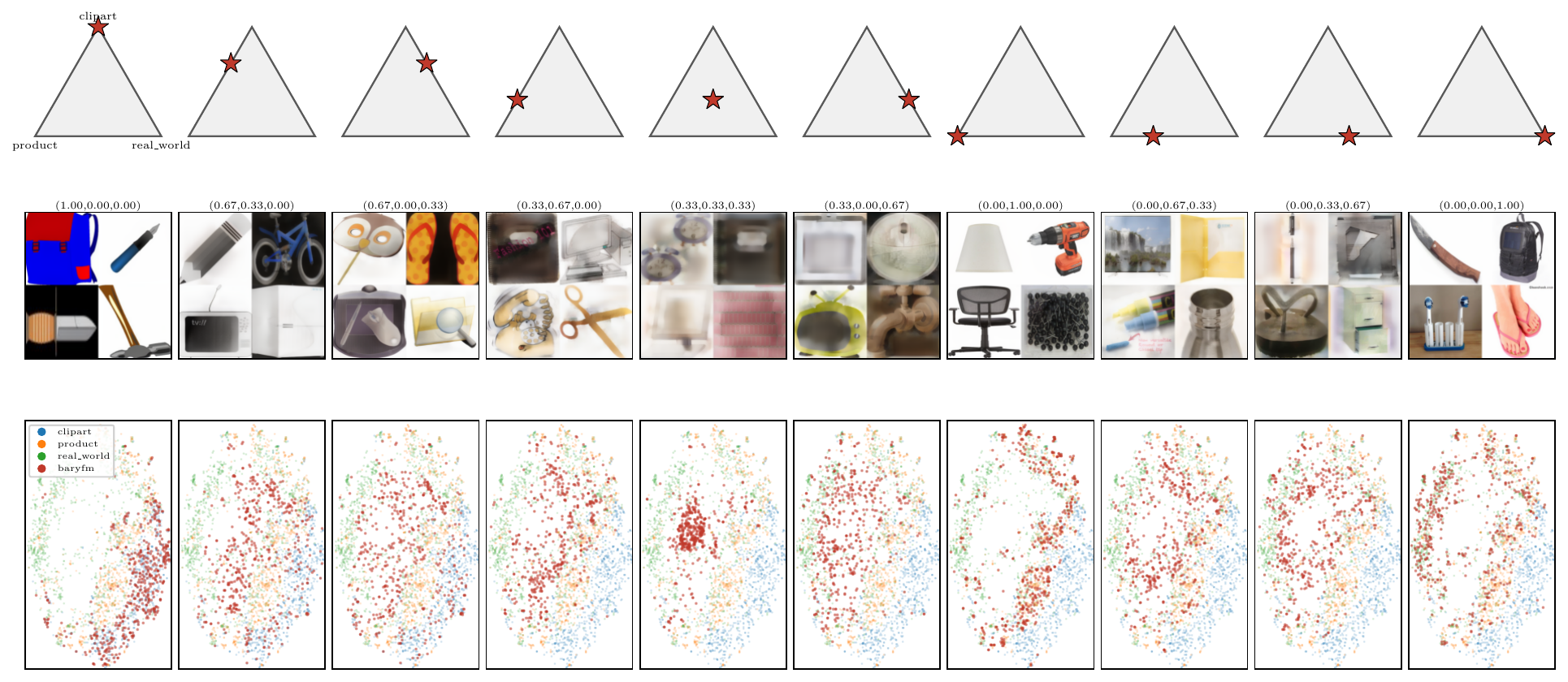}
    \caption{Visualization of nonlinear mixing of multiple domain data via \texttt{BaryFM} (ours). In the first row, we show the simplex, and picked mixing weights $\lambda$. In the second row, we show the decoded images of samples from $\text{Bar}(\lambda, p_{1:K})$. In the third row, we show a t-SNE visualization~\citep{maaten2008visualizing} of the \gls{vae} embeddings of the samples from the barycenter.}
    \label{fig:baryfm-dg}
\end{figure}

We show a comparison in the blending of domains in Figures~\ref{fig:domain-mixup-dg} (Domain MixUp) and~\ref{fig:baryfm-dg} (\texttt{BaryFM}, ours). As previously mentioned, these methods blend samples in the latent space of an autoencoder (a \gls{vae} for images, and \gls{pca} for \gls{eeg}). As is natural with MixUp~\citep{zhang2017mixup}, linearly blending samples induces a ghosting effect. While doing that in the latent space of a \gls{vae} alleviates that effect to an extent, these artifacts are still seen in the images (c.f., non-vertices of the simplex). From the t-SNE, we see that the generated samples span a different region of the latent space, indicating that, through Domain MixUp, one covers new regions, enriching the source data.

In comparison with Domain MixUp, our proposed \texttt{BaryFM} interpolated data exhibit fuzziness, rather than ghosting artifacts. Looking at the t-SNE embeddings in the third row of Figure~\ref{fig:baryfm-dg}, we can also see that the interpolated samples span a wider region of the latent space. For instance, for uniform weights, $\lambda = (0.33, 0.33, 0.33)$, our \texttt{BaryFM} concentrates less than linear mixing. These results help explaining the different performance of \texttt{BaryFM} and Domain MixUp, as well as the gain in performance our method has over Domain MixUp~\citep{wang2020heterogeneous} and MixUp~\citep{zhang2017mixup} on the Office Home dataset.

\subsection{Experimental Details}

As shown in Figure~\ref{fig:augmentation-dg}, we perform data augmentation to complement source domain data in the latent space of an autoencoder. For images, we use a \gls{vae}~\citep{kingma2013auto,rombach2022high}, and for \glspl{eeg}, we use a linear autoencoder (that is, \gls{pca}). The overall pipeline consists of pre-encoding the source domain data, then training a \texttt{BaryFM} model over the latent codes. The main advantage here consists of exploiting the semantic information encoded in such a latent space.

\begin{figure}[ht]
    \centering
    \begin{subfigure}{0.3\linewidth}
        \includegraphics[width=\linewidth]{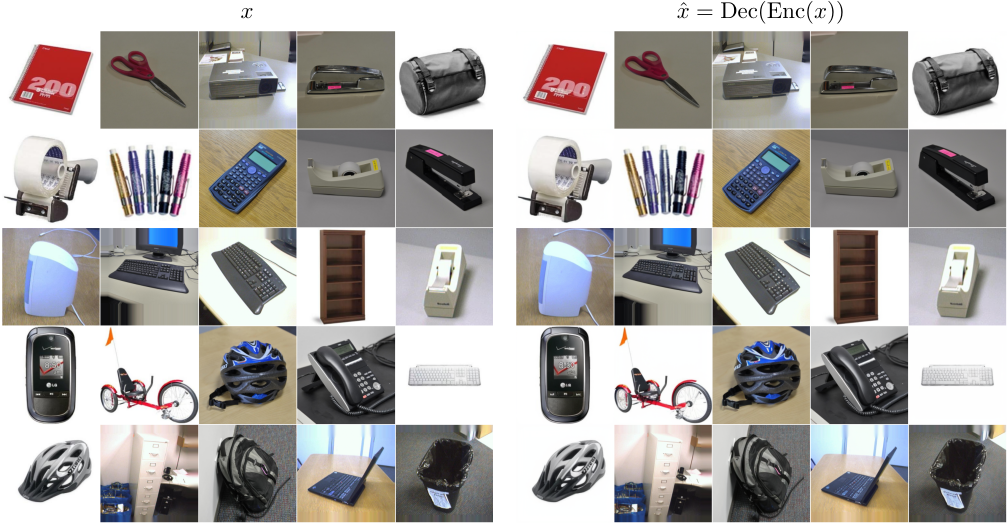}
        \caption{Office 31}
    \end{subfigure}
    \begin{subfigure}{0.3\linewidth}
        \includegraphics[width=\linewidth]{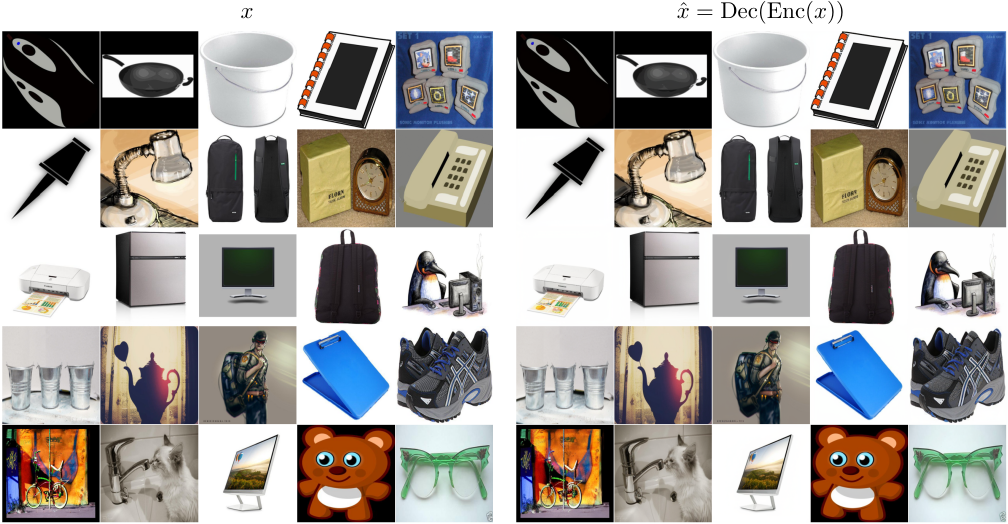}
        \caption{Office Home}
    \end{subfigure}
    \begin{subfigure}{0.3\linewidth}
        \includegraphics[width=\linewidth]{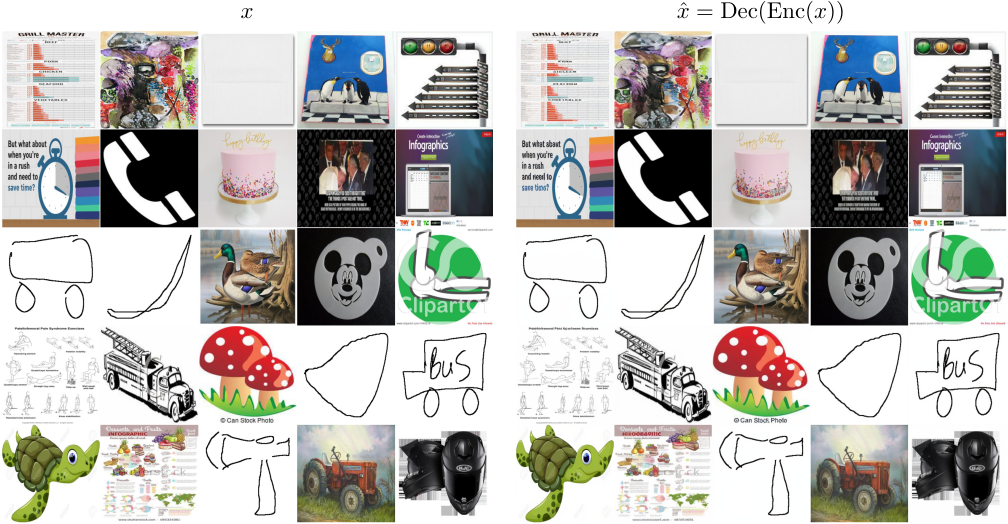}
        \caption{Domain Net}
    \end{subfigure}
    \caption{The effect of autoencoding images in the Office31, Office Home, and Domain Net, using the \gls{vae} of~\cite{rombach2022high}. On each sub-figure, we show in the left the original images in the dataset $x$, and the reconstructed images from the latent codes, $\hat{x} = \text{Dec}(\text{Enc}(x))$.}
    \label{fig:round-trip-VAE}
\end{figure}

\paragraph{Variational Autoencoder and \texttt{BaryFM} architecture.} We use the \gls{vae} of~\cite{rombach2022high} (\texttt{sd-vae-ft-mse}), which downsamples images by a factor of 8. In parallel, following standard practice~\citep{peng2019moment}, each image in the Office 31, Office Home, and DomainNet is resized to $3\times256\times256$. This means that the \gls{vae} produces latent codes of shape $4\times32\times 32$, that is, the dimensionality of inputs is $\text{dim}_{x} = 4096$. We show the effect of this round trip in Figure~\ref{fig:round-trip-VAE}.

\begin{figure}[ht]
    \centering
    \includegraphics[width=\linewidth]{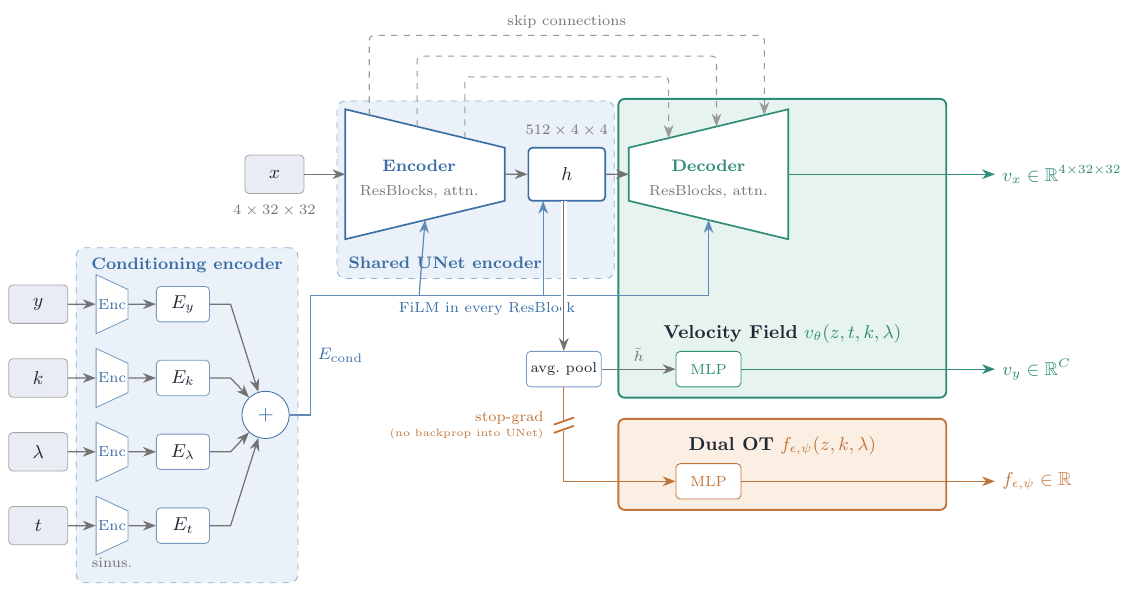}
    \caption{Architecture of our \texttt{BaryFM} used in the domain generalization experiments.}
    \label{fig:UNet Architecture}
\end{figure}

As a result of these choices, the latent codes in the image experiments in domain generalization have an underlying spatial structure. We therefore adapt the \gls{mlp} architecture we have used in domain adaptation, fairness and Bayesian posterior aggregation to take this structure into account. An overview of our architecture is shown in Figure~\ref{fig:UNet Architecture}. Briefly, $x$ is fed into the U-Net, which outputs the $x-$component $v_{x}$ of $v_{\theta}$. We use the bottleneck of the U-Net, after an average pooling operation, to produce the label component of the velocity field, $v_{y} \in \mathbb{R}^{C}$, and the dual \gls{ot} head $f_{\epsilon,\psi}(z,k,\lambda)$. The conditioning variables $y, k, \lambda$ and $t$ are embedded using 2-layer \glspl{mlp} ($y \in \Delta_{C}$ and $\lambda \in \Delta_{K}$), a lookup table $k \in \{1,\cdots,K\}$, and sinusoidal positional encoding ($t \in [0, 1]$). These produce embedding vectors $E_{t}, E_{k}, E_{\lambda}, E_{y}$ which are summed to produce a conditional embedding $E_{\text{cond}} \in \mathbb{R}^{\text{dim}_{\text{cond}}}$, with $\text{dim}_{\text{cond}} = 256$. The embedding $E_{\text{cond}}$ modulates the residual blocks in the UNet through a feature-wise affine transformation~\citep{perez2018film}. Furthermore, following~\cite{ho2020denoising}, the residual blocks at resolutions $16\times16$, $8\times8$, $4\times4$, and the bottleneck, are followed by multi-head and self-attention~\citep{vaswani2017attention} over spatial positions of the feature maps. We summarize this architecture in Table~\ref{tab:arch-dg}.

\begin{table}[ht]
    \centering
        \begin{tabular}{lcccccc}
            \toprule
            Dataset & Latent shape & $K$ & $C$ & Channels & Attention res. & \# Params\\
            \midrule
            Office 31   & $4\times32\times32$ & 2 & 31  & $(128, 256, 512, 512)$ & $\{16, 8, 4\}$ & 117.23M\\
            Office-Home & $4\times32\times32$ & 3 & 65  & $(128, 256, 512, 512)$ & $\{16, 8, 4\}$ & 117.25M\\
            DomainNet   & $4\times32\times32$ & 5 & 345 & $(128, 256, 512, 512)$ & $\{16, 8, 4\}$ & 117.39M\\
            \bottomrule
        \end{tabular}
    \caption{Architecture of the UNet used for generating the augmented pools in domain generalization. $K$ is the number of source domains in each leave-one-domain-out fold, and $C$ the number of classes. All networks use 2 residual blocks per level, 4 attention heads, and a conditioning dimension $\text{dim}_{c} = 256$. The last column gives the total number of trainable parameters, including the label and dual \gls{ot} heads.}
    \label{tab:arch-dg}
\end{table}

\begin{figure}[ht]
    \centering
    \begin{subfigure}{0.24\linewidth}
        \includegraphics[width=\linewidth]{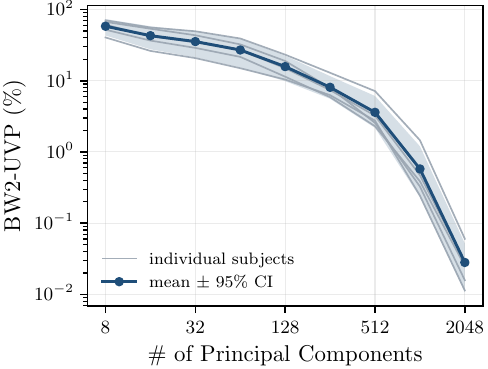}
        \caption{}
    \end{subfigure}
    \begin{subfigure}{0.24\linewidth}
        \includegraphics[width=\linewidth]{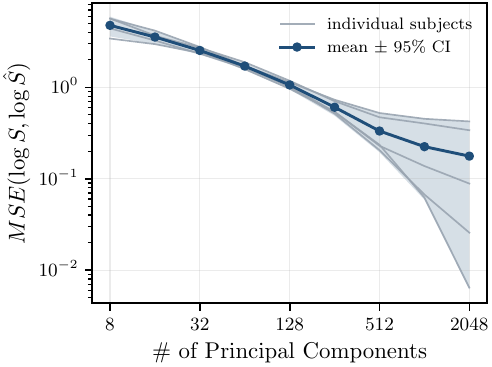}
        \caption{}
    \end{subfigure}
    \begin{subfigure}{0.24\linewidth}
        \includegraphics[width=\linewidth]{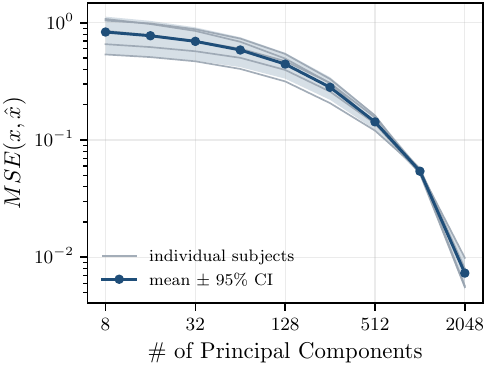}
        \caption{}
    \end{subfigure}
    \begin{subfigure}{0.24\linewidth}
        \includegraphics[width=\linewidth]{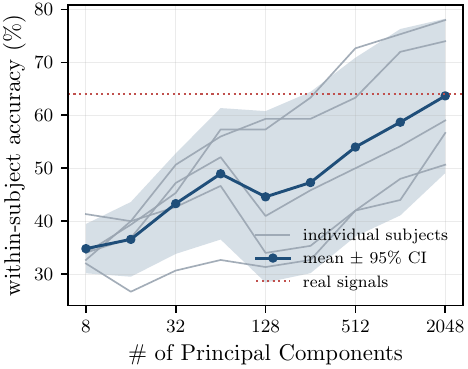}
        \caption{}
    \end{subfigure}\\
    \begin{subfigure}{0.24\linewidth}
        \includegraphics[width=\linewidth]{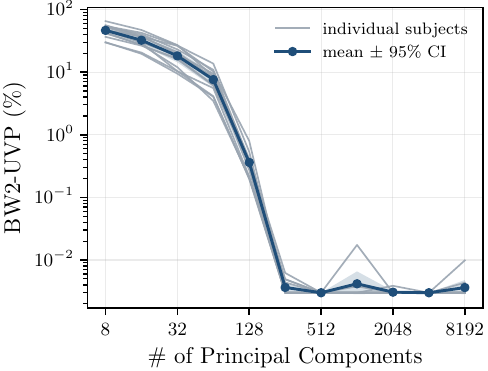}
        \caption{}
    \end{subfigure}
    \begin{subfigure}{0.24\linewidth}
        \includegraphics[width=\linewidth]{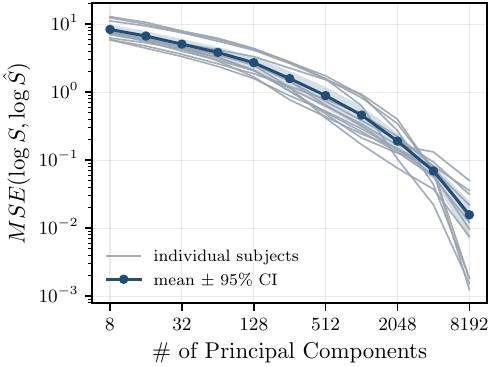}
        \caption{}
    \end{subfigure}
    \begin{subfigure}{0.24\linewidth}
        \includegraphics[width=\linewidth]{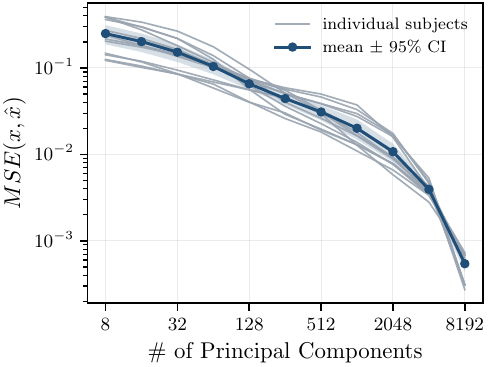}
        \caption{}
    \end{subfigure}
    \begin{subfigure}{0.24\linewidth}
        \includegraphics[width=\linewidth]{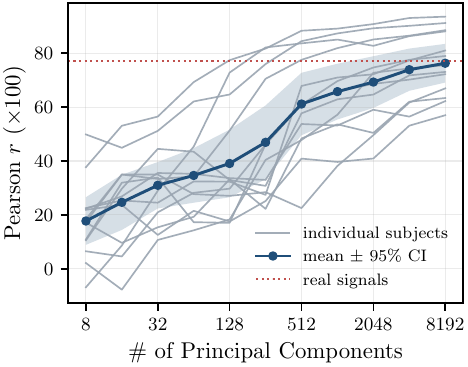}
        \caption{}
    \end{subfigure}
    \caption{Channel-wise BW2-UVP\%, log-spectral distance, MSE and downstream performance as a function of number of principal components for BCI-IV-2a (a -- d) and SEED-VIG (e -- h).}
    \label{fig:pca-fit-eeg}
\end{figure}

\paragraph{Principal Component Analysis.} For BCI-CIV-2a and SEED-VIG, we fit a \gls{pca} model to the standardized \glspl{eeg} in the source subjects of each benchmark. In these benchmarks, input data is $x \in \mathbb{R}^{V\times T}$, where $V$ is the number of electrodes and $T$ is the number of timesteps. We flatten these \glspl{eeg} into $x \in \mathbb{R}^{VT}$ before learning \gls{pca}. This makes a raw data space of $22\times800 = 17600$ dimensions for BCI-CIV-2a, and $17\times 1600 = 27200$ dimensions for SEED-VIG. In this context, we measure several metrics to select a reasonable dimensionality for the latent space. In the following we denote $\hat{x} = \text{Dec}(\text{Enc}(x))$,
\begin{enumerate}
    \item variate-wise BW2-UVP\% between $x$ and $\hat{x}$. This metric reduces each \gls{eeg} to a vector $\mu \in \mathbb{R}^{V}$ of variate-wise means, and compute the BW2-UVP\% metric over it. It measures how much the latent space distorts the distributional information of the variates.
    \item Mean Squared Error (MSE) of log spectral features, denoted $S$ for the Welch spectral density of $x$ (resp. $\hat{S}$ for $\hat{x}$).
    \item MSE between $x$ and $\hat{x}$.
    \item Downstream performance. For BCI-IV-2a, we report the per-subject, intra-subject classification accuracy. For SEED-VIG, we report the per-subject, intra-subject Pearson correlation in \%. This metric probes how well the latent space preserves the task structure (e.g., class separation for classification).
\end{enumerate}
We show a comparison of these metrics as a function of number of principal components in Figure~\ref{fig:pca-fit-eeg}. Overall, we can see that the metrics improve with more principal components, and, especially, around $\text{dim}=2048$ the downstream task performance matches the raw signals. We, therefore, use $\text{dim}=2048$ for BCI-CIV-2a, and $\text{dim}=4096$ for SEED-VIG.

\begin{table}[ht]
    \centering
\resizebox{\linewidth}{!}{
    \begin{tabular}{lccccc}
        \toprule
        Hyper-parameter & Office 31 & Office-Home & DomainNet & BCI-CIV-2a & SEED-VIG\\
        \midrule
        Latent space & \gls{vae}, $4\times32\times32$ & \gls{vae}, $4\times32\times32$ & \gls{vae}, $4\times32\times32$ & \gls{pca}, $2048$ & \gls{pca}, $4096$\\
        Architecture & UNet & UNet & UNet & MLP & MLP\\
        \midrule
        Optimizer & Adam & Adam & Adam & AdamW & AdamW\\
        Learning rate & $10^{-4}$ & $10^{-4}$ & $10^{-4}$ & $10^{-4}$ & $10^{-4}$\\
        Weight decay & -- & -- & -- & 0.05 & 0.05\\
        Learning rate schedule & Cosine, $\text{LR}_{\text{min}}=10^{-6}$ & Cosine, $\text{LR}_{\text{min}}=10^{-6}$ & Cosine, $\text{LR}_{\text{min}}=10^{-6}$ & Cosine, $\text{LR}_{\text{min}}=10^{-6}$ & Cosine, $\text{LR}_{\text{min}}=10^{-6}$\\
        Training iterations & $30\,000$ & $30\,000$ & $30\,000$ & $50\,000$ & $50\,000$\\
        Batch size & $128$ & $128$ & $128$ & $512$ & $256$\\
        Gradient clipping & None & None & None & $1.0$ & $1.0$\\
        EMA decay & $0.9999$ & $0.9999$ & $0.9999$ & $0.9999$ & $0.9999$\\
        Entropic regularization & $10^{-2}$ & $10^{-2}$ & $10^{-2}$ & $10^{-3}$ & $10^{-3}$\\
        Dual steps per iteration & $1$ & $1$ & $1$ & $1$ & $1$\\
        Fixed-point iterations & $1$ & $1$ & $1$ & $1$ & $1$\\
        \midrule
        ODE solver & Euler & Euler & Euler & Midpoint & Midpoint\\
        $n_{\text{gen}}$ & 100,000 & 100,000 & 100,000 & 5,000 & 26,550\\
        \gls{ode} steps $T$ & $100$ & $100$ & $100$ & $100$ & $100$\\
        \midrule
        Seeds (DG loop) & \texttt{0--4} & \texttt{0--4} & \texttt{0--2} & \texttt{0--9} & \texttt{0--9}\\
        \bottomrule
    \end{tabular}
}
    \caption{Training and inference hyper-parameters used in our domain generalization experiments.}
    \label{tab:hps-dg}
\end{table}

\paragraph{Hyper-Parameters.} We summarize the overall training and inference hyper-parameters in Table~\ref{tab:hps-dg} which are similar to the hyper-parameters reported for domain adaptation (Tables~\ref{tab:train-hps-da} and~\ref{tab:inference-hps-da}).

\subsection{Detailed Results}

Similarly do domain adaptation, we detail our obtained results per domain for each benchmark. Note that results are different from domain adaptation (and sometimes, inferior) due to different factors. First, the \gls{wbt} pipeline runs over embeddings, and can therefore be seen as linear probing. In contrast, our domain generalization results fine-tune the complete network from the synthetic generated data. Second, in domain generalization one does not have access to the target domain data, which makes the task itself harder than domain adaptation.

\begin{table}[ht]
    \centering
\begin{tabular}{lccccc}
    \toprule
    Method & Amazon & dSLR & Webcam & Avg. & Rank \\
    \midrule
    ERM (ResNet 50) & \result{60.39}{0.74} & \result{98.40}{0.55} & \result{94.34}{1.33} & \result{84.38}{0.53} & 5 \\
    MixUp & \result{60.74}{1.08} & \result{99.00}{0.00} & \result{97.23}{1.14} & \result{85.66}{0.69} & 2 \\
    Domain MixUp & \result{58.72}{1.49} & \result{99.20}{0.84} & \result{96.98}{1.03} & \result{84.97}{0.49} & 3 \\
    \midrule
    BaryFM$_{\text{Vertex}}$ & \result{60.04}{1.08} & \result{98.80}{0.45} & \result{95.85}{1.14} & \result{84.89}{0.58} & 4 \\
    BaryFM$_{\text{Random}}$ & \result{62.20}{1.31} & \result{99.40}{0.89} & \result{96.23}{1.18} & \result{85.94}{0.66} & 1 \\
    \bottomrule
\end{tabular}
\caption{Per domain Office 31 domain generalization results. Results denote classification accuracy (in \%) over 5 seeds.}
\label{tab:dg-per-domain-office31}
\end{table}

On Office 31~\citep{saenko2010adapting}, we report our results in Table~\ref{tab:dg-per-domain-office31}. The performance of methods on dSLR and Webcam is comparable to domain adaptation (c.f. Table~\ref{tab:per-domain-office31}). This is intuitive, since for each of these targets, there is a closely resembling domain in the source data (e.g., Webcam in source, for dSLR as target). The biggest difference comes from Amazon (c.f., DA achieves around 70\% accuracy, while DG achieves 62.20), as this is the most different and challenging domain. In comparison, \texttt{BaryFM}$_{\text{Random}}$ achieves the best rank, with \texttt{MixUp} being a close second best. We note the gap between \texttt{Domain MixUp} and our proposed $\texttt{BaryFM}_{\text{Random}}$ of $0.97$ (\result{85.94}{0.66} ours vs. \result{84.97}{0.69} theirs), higher than seed noise.

\begin{table}[ht]
    \centering
\begin{tabular}{lcccccc}
    \toprule
    Method & Art & Clipart & Product & Real World & Avg. & Rank \\
    \midrule
    ERM (ResNet 50) & \result{65.68}{1.19} & \result{53.91}{1.10} & \result{79.37}{1.16} & \result{80.00}{0.69} & \result{69.74}{0.57} & 4 \\
    MixUp & \result{65.56}{0.94} & \result{57.54}{1.43} & \result{78.99}{0.69} & \result{79.64}{0.55} & \result{70.43}{0.34} & 3 \\
    Domain MixUp & \result{63.12}{2.06} & \result{55.25}{1.83} & \result{78.30}{1.16} & \result{78.78}{0.75} & \result{68.86}{0.82} & 5 \\
    \midrule
    BaryFM$_{\text{Vertex}}$ & \result{66.43}{1.56} & \result{55.74}{1.00} & \result{80.15}{0.79} & \result{79.87}{0.82} & \result{70.55}{0.47} & 2 \\
    BaryFM$_{\text{Random}}$ & \result{67.18}{1.34} & \result{57.49}{1.11} & \result{80.17}{0.92} & \result{79.82}{0.38} & \result{71.17}{0.76} & 1 \\
    \bottomrule
\end{tabular}
\caption{Per domain Office Home domain generalization results. denote classification accuracy (in \%) over 5 seeds.}
\label{tab:dg-per-domain-office-home}
\end{table}

On Office Home~\citep{venkateswara2017deep}, we report our results in Table~\ref{tab:dg-per-domain-office-home}. Here, except for Domain MixUp, all methods are able to improve the \gls{erm} baseline over the seed standard deviation, with \texttt{BaryFM}$_{\text{Random}}$ being the best method. We provide an example of the different interpolations acquired by our method in the Office Home dataset in Figures~\ref{fig:domain-mixup-dg} and~\ref{fig:baryfm-dg}. Note that, while Domain MixUp contains some degree of a ghosting effect, and the samples concentrate in the embedding space as the mixing weights go away from the vertices of the simplex. Our \texttt{BaryFM}, on the other hand, becomes more blurred, and its samples are more distributed in the latent space.

\begin{table}[ht]
    \centering
\resizebox{\linewidth}{!}{
\begin{tabular}{lcccccccc}
    \toprule
    Method & Clipart & Infograph & Painting & Quickdraw & Real & Sketch & Avg. & Rank \\
    \midrule
    ERM (ResNet 101) & \result{69.97}{0.33} & \result{28.34}{0.94} & \result{58.34}{1.20} & \result{13.61}{0.07} & \result{70.13}{0.34} & \result{57.05}{0.71} & \result{49.57}{0.24} & 4 \\
    MixUp & \result{69.71}{0.21} & \result{28.24}{0.45} & \result{58.99}{0.66} & \result{13.50}{0.59} & \result{69.59}{0.60} & \result{57.19}{0.34} & \result{49.53}{0.11} & 5 \\
    Domain MixUp & \result{71.00}{0.27} & \result{28.91}{0.54} & \result{59.53}{1.01} & \result{14.97}{0.30} & \result{70.93}{0.60} & \result{58.43}{0.40} & \result{50.63}{0.20} & 1 \\
    \midrule
    BaryFM$_{\text{Vertex}}$ & \result{69.72}{0.39} & \result{28.25}{0.81} & \result{58.79}{0.82} & \result{13.43}{0.09} & \result{70.57}{0.68} & \result{57.40}{1.18} & \result{49.69}{0.34} & 3 \\
    BaryFM$_{\text{Random}}$ & \result{70.57}{0.18} & \result{28.68}{0.47} & \result{59.83}{0.84} & \result{13.83}{0.56} & \result{70.25}{0.72} & \result{57.99}{0.47} & \result{50.19}{0.21} & 2 \\
    \bottomrule
\end{tabular}
}
\caption{Per domain DomainNet domain generalization results. Results denote classification accuracy (in \%) over 3 seeds.}
\label{tab:dg-per-domain-domain-net}
\end{table}

Next, we analyze DomainNet~\citep{peng2019moment}, which is the more challenging and large scale image benchmark. Overall, improvements in this benchmark are smaller (biggest difference being Domain MixUp vs. ERM, with $\delta = 1.06$). This is normal, due the hardness of the task. This dataset contains 345 classes, under severe domain shift (e.g., Quickdraw). Overall, Domain MixUp of~\cite{wang2020heterogeneous} proves very effective, being the best method over all domains. Our \texttt{BaryFM}$_{\text{Random}}$ comes in second. Nonetheless, we remark that both methods improve over the \gls{erm}, which proves the effectiveness of the augmentation strategy by mixing domain characteristics.

\begin{table}[ht]
    \centering
\begin{tabular}{lcccc}
    \toprule
    Method & S8 & S9 & Pool. & Rank \\
    \midrule
    ERM (CBraMod) & \result{50.06}{4.84} & \result{51.20}{2.78} & \result{50.63}{3.45} & 5 \\
    MixUp & \result{51.04}{3.30} & \result{53.70}{3.36} & \result{52.37}{3.11} & 4 \\
    Domain MixUp & \result{50.89}{3.02} & \result{54.51}{2.90} & \result{52.70}{2.78} & 3 \\
    \midrule
    BaryFM$_{\text{Vertex}}$ & \result{53.75}{4.06} & \result{54.62}{3.76} & \result{54.18}{3.48} & 1 \\
    BaryFM$_{\text{Random}}$ & \result{52.71}{3.70} & \result{54.76}{3.91} & \result{53.73}{3.43} & 2 \\
    \bottomrule
\end{tabular}
\caption{Per domain BCI-IV-2a domain generalization results. denote classification accuracy (in \%) over 5 seeds.}
\label{tab:dg-per-domain-bci-iv-2a}
\end{table}

Now, we analyze the generalization results on BCI-CIV-2a~\citep{brunner2008bci}, which are run in the same setting as~\cite{wang2025cbramod}. We use subjects 1 -- 5 as sources, 6 -- 7 for checkpoint selection, and 8 -- 9 for testing. In general, \gls{eeg} datasets have significantly wider confidence interval margins than image benchmarks, due the intrinsic difficulty of \gls{eeg} decoding. Overall, our interpolations improve over the \gls{erm} and MixUp variations. For instance, our \texttt{BaryFM}$_{\text{Vertex}}$ is the best performing method with an improvement of $3.54\%$ over ERM. The second best performing method is $\texttt{BaryFM}_{\text{Random}}$, with an improvement of $3.09$ over ERM. The larger improvement over ERM goes in the same direction as our domain adaptation results (c.f. Table~\ref{tab:per-domain-bci-iv-2a-crosssub}).

\begin{table}[ht]
    \centering
\begin{tabular}{lcccccc}
    \toprule
    Method & S14 & S15 & S16 & S17 & Pool. & Rank \\
    \midrule
    ERM (CBraMod) & \result{55.57}{4.05} & \result{33.58}{5.75} & \result{80.43}{3.16} & \result{62.29}{7.41} & \result{58.30}{2.44} & 5 \\
    MixUp & \result{53.46}{2.91} & \result{39.40}{4.83} & \result{80.79}{2.19} & \result{59.13}{6.94} & \result{61.11}{2.57} & 2 \\
    Domain MixUp & \result{57.46}{2.61} & \result{40.36}{5.49} & \result{82.77}{2.28} & \result{68.12}{6.10} & \result{61.04}{3.15} & 3 \\
    \midrule
    BaryFM$_{\text{Vertex}}$ & \result{53.62}{2.43} & \result{47.04}{4.94} & \result{79.04}{2.09} & \result{59.13}{5.07} & \result{59.25}{5.63} & 4 \\
    BaryFM$_{\text{Random}}$ & \result{54.93}{2.84} & \result{43.18}{5.95} & \result{80.68}{2.33} & \result{63.45}{6.31} & \result{62.08}{2.38} & 1 \\
    \bottomrule
\end{tabular}
\caption{Per domain SEED-VIG domain generalization results. Results denote Pearson correlation (in \%) over 5 seeds.}
\label{tab:dg-per-domain-seedvig}
\end{table}

Finally, we test augmentations on SEED-VIG~\citep{zheng2017multimodal}, where we use subjects 1 -- 13 as source subjects, 14 -- 17 as test subjects, and the rest for checkpoint selection. This split is different from the results used in domain adaptation (c.f. Table~\ref{tab:per-domain-seedvig}). Overall, our \texttt{BaryFM}$_{\text{Random}}$ has an improvement of 3.78\% in Pearson correlation, above MixUp variants (2.81 and 2.74 for MixUp and Domain MixUp, respectively).

\paragraph{Conclusion.} Overall, our \texttt{BaryFM}$_{\text{Random}}$, which consists of drawing samples from randomly sampled barycenters in $\mathcal{W}(p_{1:K})$ is the best augmentation method in 3 (Office 31, Office Home and SEED-VIG) our of 5 datasets, being more consistent than \texttt{BaryFM}$_{\text{Vertex}}$, which augments by sampling the vertices of the simplex. This demonstrates that there is a true advantage in mixing domain characteristics.

\end{document}